\documentclass[]{academic_template}
\IfFileExists{latexml.sty}{\usepackage{latexml}}{%
  \ifdefined\iflatexml\else\newif\iflatexml\fi
}
\providecommand{\printprojectlinks}{}
\iflatexml
  \usepackage{authblk}
  \usepackage{hyperref}
  \makeatletter
  \newcount\vxl@nextlabel
  \def\vxl@getlabel#1{\vxl@labelhead#1\vxl@end}
  \def\vxl@labelhead#1#2\vxl@end{%
    \ifcsname vxl@label@\detokenize{#1#2}\endcsname\else
      \global\advance\vxl@nextlabel by 1
      \expandafter\xdef\csname vxl@label@\detokenize{#1#2}\endcsname{vxl\the\vxl@nextlabel}%
    \fi
    \edef\vxl@resolved{\csname vxl@label@\detokenize{#1#2}\endcsname}%
  }
  \def\vxl@stop{\vxl@stop}
  \def\vxl@parselabel#1,{%
    \def\vxl@item{#1}%
    \ifx\vxl@item\vxl@stop
      \let\vxl@next\relax
    \else
      \if\relax\detokenize{#1}\relax\else
        \vxl@getlabel{#1}%
        \ifx\vxl@converted\@empty
          \edef\vxl@converted{\vxl@resolved}%
        \else
          \edef\vxl@converted{\vxl@converted,\vxl@resolved}%
        \fi
      \fi
      \let\vxl@next\vxl@parselabel
    \fi
    \vxl@next
  }
  \def\vxl@convertlabels#1{%
    \def\vxl@converted{}%
    \vxl@parselabel#1,\vxl@stop,
  }
  \let\vxl@nativeauthor\author
  \newcommand{\vxl@htmlauthor}[2][]{%
    \vxl@convertlabels{#1}%
    \expandafter\vxl@nativeauthor\expandafter[\vxl@converted]{#2}%
  }
  \let\author\vxl@htmlauthor
  \renewcommand{\affiliation}[2][]{%
    \vxl@convertlabels{#1}%
    \ifx\vxl@converted\@empty
      \affil{#2}%
    \else
      \expandafter\affil\expandafter[\vxl@converted]{#2}%
    \fi
  }

  \def\vxl@htmlinfo{}
  \providecommand{\contribution}{}
  \renewcommand{\contribution}[2][]{%
    \if\relax\detokenize{#1}\relax
      \g@addto@macro\vxl@htmlinfo{\par\noindent\emph{#2}\\\par}%
    \else
      \vxl@convertlabels{#1}%
      \expandafter\affil\expandafter[\vxl@converted]{\emph{#2}}%
    \fi
  }
  \providecommand{\checkdata}{}
  \renewcommand{\checkdata}[2][]{%
    \g@addto@macro\vxl@htmlinfo{\par\noindent\textbf{#1:} #2\\\par}%
  }

  \date{}
  \renewcommand{\printprojectlinks}{%
    \vxl@htmlinfo
    \global\let\vxl@htmlinfo\@empty
  }

  \providecommand{\setthemecolor}[1]{}
  \providecommand{\setsectionfont}[1]{}
  \providecommand{\setsubsectionfont}[1]{}
  \providecommand{\setsectionlinecolor}[1]{}
  \providecommand{\setsectionlinethickness}[1]{}
  \providecommand{\settitlefont}[1]{}
  \providecommand{\setabstractfont}[1]{}
  \providecommand{\setabstractframecolor}[1]{}
  \providecommand{\setabstractbgcolor}[1]{}
  \providecommand{\setlogoheight}[1]{}
  \providecommand{\setlogospacing}[1]{}
  \providecommand{\setlogotolineshift}[1]{}
  \providecommand{\settoprulethickness}[1]{}
  \providecommand{\setbottomrulethickness}[1]{}
  \providecommand{\settitlerulethickness}[1]{}

  \providecommand{\sectionformatnoline}{}
  \providecommand{\sectionformatwithline}{}

  \providecommand{\abstractboxon}{}

  \providecommand{\setsjtublue}{}
  \providecommand{\themeblue}[1]{\textbf{#1}}
  
  \providecommand{\beginappendix}{%
    \appendix
    \section*{Appendix}
    \setcounter{section}{1}%
    \renewcommand{\thesubsection}{\Alph{section}.\arabic{subsection}}%
  }
  \makeatother
\fi

\usepackage[toc,page,header]{appendix}
\usepackage{amssymb}
\usepackage{amsmath}
\usepackage{makecell}
\usepackage{algorithm}
\usepackage{algpseudocode}
\usepackage{graphicx}
\usepackage{listings}
\usepackage{url}
\usepackage{xspace}
\usepackage{booktabs}
\usepackage{longtable}
\usepackage{wrapfig}
\usepackage{colortbl}
\usepackage{xcolor}
\usepackage{subcaption}
\usepackage{microtype}
\usepackage{placeins}
\usepackage{etoolbox}
\usepackage{multirow}
\usepackage{nicematrix}
\usepackage{bm}
\usepackage[most]{tcolorbox}
\usepackage{hyperref}
\usepackage{cleveref}

\newcommand{\tablestyle}[2]{\setlength{\tabcolsep}{#1}\renewcommand{\arraystretch}{#2}\centering\footnotesize}

\newcommand{\benchname}{\textsc{V-ICAL}\xspace}

\newcommand{\PaperAbstract}{%
While In-Context Learning (ICL) enables models to adapt from exemplars without parameter updates, multimodal ICL remains largely underexplored, particularly regarding video demonstrations in interactive environments. For multimodal agents, learning from videos presents unique challenges: they must translate in-context demonstrations into executable policies, ground these policies in novel visual states, and iteratively refine actions based on environmental feedback. We introduce \benchname, a novel benchmark designed to evaluate video-based ICL for multimodal agents. Comprising 342 interactive tasks across 37 environments, \benchname utilizes human-curated demonstration videos as task-specific behavioral exemplars, evaluating agents through sustained interaction from a target initialization. The benchmark seamlessly connects in-context knowledge induction with core agentic capabilities, including state grounding, temporal memory, planning, and adaptation in dynamic environments. Extensive evaluations across 19 state-of-the-art multimodal agents reveal significant limitations: the best-performing model, Seed-2.1-Pro, achieves a score of only 54.4/100, while other leading models (e.g., Gemini-3.1-Pro, Gemini-3.6-Flash, GPT-5.6-sol) fail to surpass 50, far below the human baseline of 83.6. Controlled comparisons further demonstrate that current agents struggle to reliably translate video exemplars into effective policies, failing to yield consistent performance gains. Ultimately, \benchname exposes a critical gap in the ICL capabilities of multimodal agents, underscoring an urgent need for future research.%
}

\setlogoheight{14mm}
\setlogospacing{5mm}
\setsjtublue
\settitlerulethickness{3pt}
\abstractboxon
\setabstractframecolor{gray}
\setabstractbgcolor{gray!10}
\setlogotolineshift{5mm}
\settoprulethickness{2.5pt}
\setbottomrulethickness{1.5pt}

\title{V-ICAL Bench: Evaluating Video In-Context Learning for Multimodal Agents in Interactive Environments}

\author[2,*]{Ziqian Fan}
\author[1,*]{Shibo Xu}
\author[2,*]{Junjie Li}
\author[1,*,\ddagger]{Xiangyu Zhao}
\author[3]{Shengyuan Ding}
\author[4]{Yifan Yang}
\author[5]{Zhenjie Yang}
\author[6]{Haodong Duan}
\author[7]{Yue Zhou}
\author[1]{Zhihang Zhong}
\author[1, \dagger]{Xue Yang}

\affiliation[1]{Shanghai Jiao Tong University}
\affiliation[2]{South China University of Technology}
\affiliation[3]{Fudan University}
\affiliation[4]{Microsoft Research Asia}
\affiliation[5]{The University of Hong Kong}
\affiliation[6]{The Chinese University of Hong Kong}
\affiliation[7]{East China Normal University}

\contribution[*]{Core Contributors}
\contribution[\ddagger]{Project Lead}
\contribution[\dagger]{Corresponding Author}

\abstract{\PaperAbstract}
\date{\today}
\checkdata[Code]{\url{https://github.com/VisionXLab/V-ICAL}}
\checkdata[Dataset]{\url{https://huggingface.co/datasets/VisionXLab/V-ICAL}}
\checkdata[Homepage]{\url{https://visionxlab.github.io/V-ICAL-Homepage}}

\renewcommand{\topfraction}{0.9}
\renewcommand{\bottomfraction}{0.8}
\renewcommand{\textfraction}{0.07}
\renewcommand{\floatpagefraction}{0.85}
\newcommand{\PaperBibStyle}{plainnat}
\newcommand{\PaperAppendixStart}{\beginappendix}

\begin{document}
\maketitle
\printprojectlinks
\section{Introduction}
\label{sec:intro}

\begin{figure*}[t]
    \centering
    \includegraphics[width=0.8\textwidth]{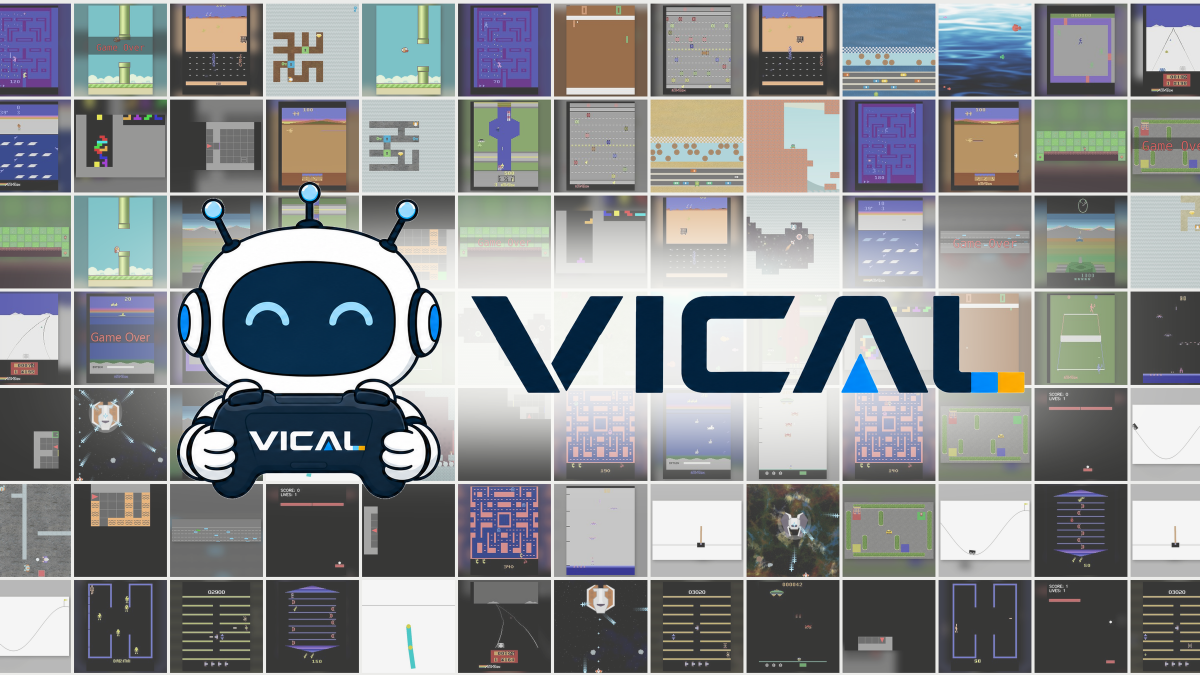}
    \par\vspace{0.4em}
    \includegraphics[width=0.8\textwidth]{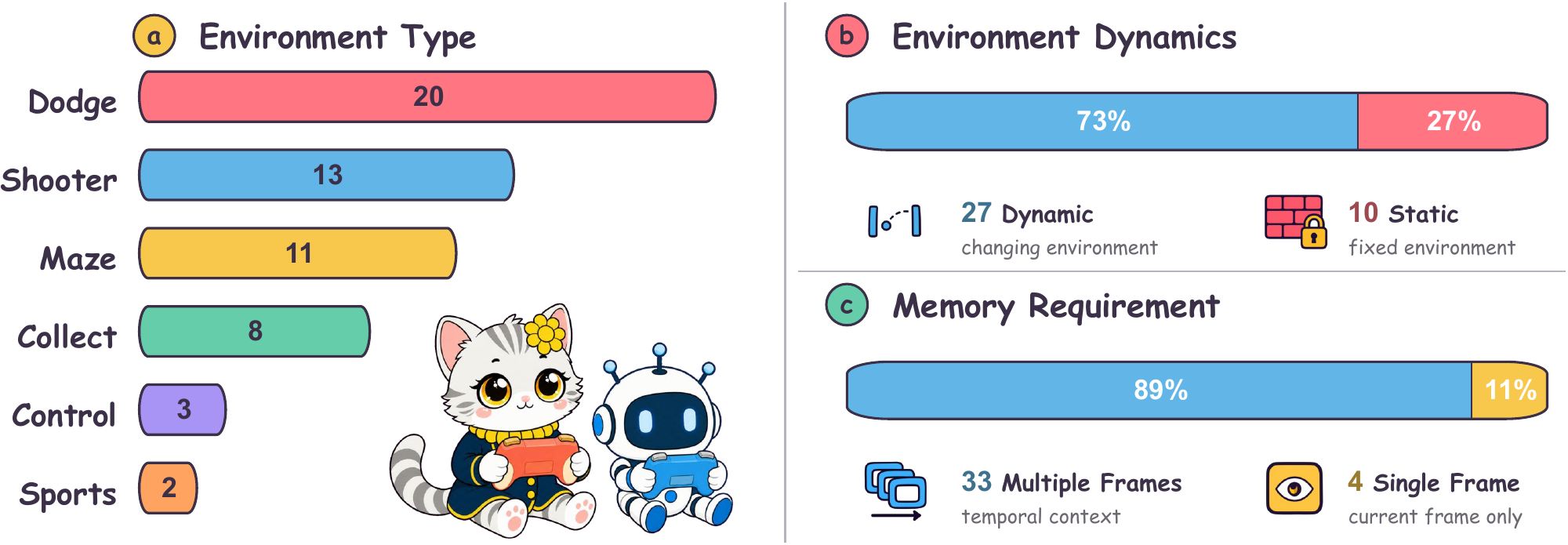}
    \caption{\benchname\ comprises 342 evaluation tasks across 37 environments, covering diverse environment types, environment dynamics, and memory requirements. As each environment may belong to multiple environment-type categories, the category counts do not sum to 37.}
    \label{fig:main}
\end{figure*}

In-context learning (ICL) allows models to adapt from demonstrations without parameter updates~\citep{agarwal2024many,jiang2024many,zong2025vl}.
For multimodal agents, videos provide a natural medium for conveying procedures, strategies, and interactive behaviors.
However, the evaluation of multimodal ICL remains largely confined to static, offline paradigms, where models passively predict answers based on visual inputs rather than actively engaging with or receiving feedback from an environment~\citep{kim2025videoicl,dong2026demo}.
This discrepancy leaves a critical void in our understanding of whether ICL can genuinely support sustained agency.

When learning from video demonstrations, multimodal agents must extract strategies from in-context videos and deploy them in closed perception–decision–action loops. This requires grounding demonstrated behaviors in novel states, executing valid actions, and adapting based on feedback~\citep{li2026mm,yang2026skillopt,liu2026rise}. However, existing benchmarks assess only isolated aspects: standard video tasks test semantic understanding~\citep{fu2025video,fu2026video,liu2026moment}, and interactive benchmarks rely on rigid traces or text-only tutorials~\citep{ruoss2024lmact,jang2025videowebarena,zhang2026gameverse}. We ask: \textit{Can multimodal agents induce policies from in-context video demonstrations and execute them through sustained interaction in dynamic environments?}

To systematically assess the capability of multimodal agents in in-context learning, we present \benchname\ (\textit{\textbf{V}ideo \textbf{I}n-\textbf{C}ontext \textbf{A}gentic \textbf{L}earning}), a comprehensive benchmark of 342 evaluation tasks spanning 37 environments and seven environment families, tailored for video ICL in interactive settings. 
Each task supplies human-curated demonstration videos as the only in-context behavioral exemplars, paired with a target initial state, an action space and minimal text specifying non-visual constraints; the agent must infer the task from the video and execute a multi-turn trajectory from pixels with a task identifier, without parameter updates or further human guidance. 
To ensure a robust evaluation, the task pool covers static and dynamic environments and decisions requiring either single-frame spatial perception or long-horizon temporal memory. Performance is measured by pass-rate and normalized-score metrics, capturing both basic ability evaluation and partial progress.

Evaluating 19 multimodal agents exposes a large gap between comprehending a demonstration and using it for sustained agency.
Even the best model, Seed-2.1-Pro, scores only 54.4/100. Other formidable models perform even worse, including Gemini-3.6-Flash (49.5), Gemini-3.1-Pro (48.0), and GPT-5.6-sol (46.7), all far below the human baseline of 83.6; the best open-weight model, DeepSeek-V4.1-Flash, reaches only 34.8.
When the environment evolves on its own or earlier observations must be retained, the average score of the four strongest models drops by 42.3\% and 32.6\%, respectively, versus 20.2\% and 17.4\% for humans. See Table~\ref{tab:main-results}.
A trajectory audit based on a four-stage decision model further indicates that demonstrated strategies transfer far less reliably than demonstrated rules, and that most failures arise in grounding live visual states and in planning.
Finally, controlled ablations show that videos improve three of four models under basic textual rules (by up to 7.1 points), but the gains are inconsistent and largely vanish once detailed rules are given.

Our contributions are threefold:

\textbf{Benchmark.} We formulate video ICL for multimodal agents as closed-loop sequential decision making and build V-ICAL, 342 tasks across 37 environments with human-curated demonstrations and calibrated pass and score targets.

\textbf{Evaluation.} We evaluate 19 proprietary and open-weight agents against human and random baselines, exposing consistent weaknesses in temporal memory and in dynamic environments.

\textbf{Diagnosis.} We propose a trajectory-auditing protocol, validated against blind human labels (95.0\% agreement), that locates where the decision loop breaks, and we ablate video availability and rule specificity to quantify what demonstrations add beyond model priors, which shows that video gains are model-dependent.

\section{Related Work}
\label{sec:related}

\textbf{In-context learning for multimodal models.}
In text-only settings, many-shot prompts yield substantial gains~\citep{agarwal2024many}. For vision-language models, \citet{jiang2024many} report approximately log-linear scaling with nearly 2{,}000 examples across 14 image datasets, while VL-ICL Bench broadens evaluation to perception, reasoning, and generation~\citep{zong2025vl}. These studies use static image--text demonstrations.

\textbf{Video ICL and demonstration-conditioned agents.}
Video-MME evaluates video comprehension through question answering rather than adaptation~\citep{fu2025video,fu2026video}. VideoICL and Demo-ICL-Bench measure procedural transfer from video demonstrations to prediction tasks~\citep{kim2025videoicl,dong2026demo}, but remain offline settings. LMAct tests interactive policies with up to 512 full episodes, finding diminishing returns~\citep{ruoss2024lmact}; our setting instead uses one to four compact clips as task specifications and tests sustained closed-loop execution across 37 heterogeneous environments. VideoWebArena evaluates tutorial use in web tasks but finds agents might perform worse with tutorials~\citep{jang2025videowebarena}.

\textbf{Benchmark in interactive environments.}
Several benchmarks evaluate agents across interactive  environments~\citep{paglieri2025balrog,hu2026lmgame,zhao2026mm,waytowich2024atari,ouyang2026gameworld}. GameVerse uses post-failure reflection via a separate model~\citep{zhang2026gameverse}. Unlike prior work relying on reflection or offline prediction, we treat video ICL as an agentic capability: the agent induces task rules from demonstrations and deploys them through closed-loop interaction.
\section{Benchmark Design}
\label{sec:benchmark}

Humans acquire skills from video demonstrations and transfer them to novel situations. We study whether multimodal agents can achieve similar capabilities through video in-context learning. 
Unlike passive video understanding, this requires inferring executable strategies from demonstrations and deploying them through sustained closed-loop interaction. Accordingly, \benchname integrates a closed-loop agent protocol, diverse environments and initial states, and complementary evaluation metrics.

\subsection{Environment Setting}
\label{sec:envpool}

From more than 100 candidates, we select 37 environments that pass six screening criteria (Appendix~\ref{app:environment-selection}), chiefly visual legibility, interaction stability, timely outcome feedback, and policy discriminability, i.e., rule-informed play must clearly outperform random actions.
They span seven Gym-ecosystem families: Toy Text (e.g., Taxi), MiniGrid (e.g., LavaCrossing), Classic Control (e.g., CartPole), ALE Atari (e.g., Seaquest), Procgen (e.g., Heist), HighwayEnv, and third-party games (e.g., Tetris, FlappyBird).

We expand the 37 base tasks into 342 tasks with two mechanisms (Appendix~\ref{app:task-pool-expansion}).
\emph{Parameterized variants} (139) change random seeds or initial frame skips to vary layouts and levels, and \emph{human-initialized variants} (166) replay a recorded human action sequence to reach a targeted start state, such as a key decision point, a typical trap, or a resource-constrained situation. For example, the Seaquest start state replays three Down actions after a 230-frame warm-up, placing the submarine underwater with a threat approaching from the left (Figure~\ref{fig:appendix-human-initialization}).

\subsection{Task Formalization}
\label{sec:task}

\begin{figure*}[t]
    \centering
    \includegraphics[width=\textwidth]{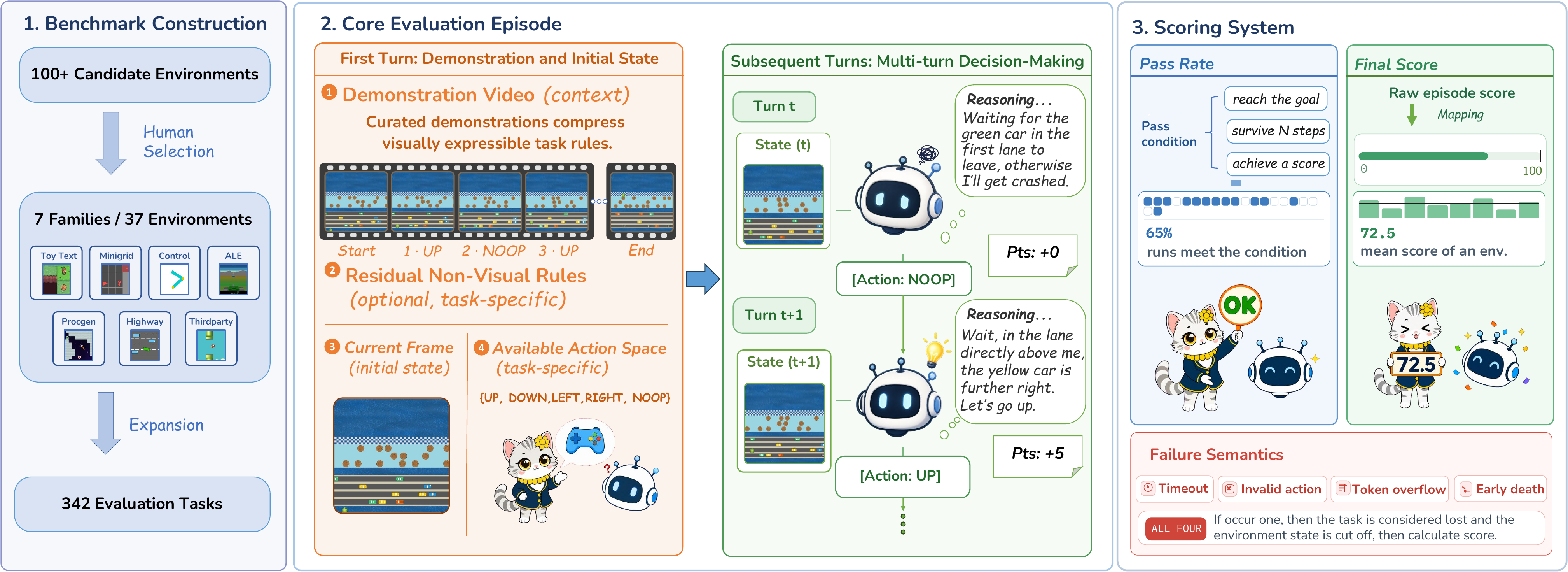}
    \caption{\textbf{Overview of \benchname.} The benchmark contains 342 evaluation tasks spanning 37 environments and seven environment families. Each episode supplies an optional in-context video demonstration package, then evaluates a multimodal agent through multi-turn visual decision-making in a closed perception--action loop. Performance is summarized by pass rate and final score with explicit failure semantics.}
    \label{fig:overview}
\end{figure*}

We formulate video ICL as a multi-turn sequential decision-making problem. The agent receives demonstration videos as task exemplars, executes actions, observes resulting visual states, and continues acting over a growing context within a perception--decision--action--feedback loop. Figure~\ref{fig:overview} gives an overview.

\paragraph{Demonstration construction.}
Human experts spend more than 200 hours in total calibrating initial states, interaction horizons, pass conditions, 
and score targets to ensure comparable difficulty across tasks. 
Each demonstration--prompt pair was independently validated and iteratively refined. We encode all demonstrable strategies in video and reserve text for residual information only, ensuring success primarily reflects video-based strategy inference rather than instruction following. 
The 342 tasks draw on 148 distinct source clips, with one to four clips per task. Each task's demonstration context spans 10 to 80 frames at 1 FPS (median 45), and every frame is captioned with the canonical name of the action that produced it.

\paragraph{First turn (in-context demonstration injection).}
The first turn establishes context through four inputs. 
(i) \textbf{Demonstration videos}: one or more human-curated clips that densely encode the environment's hidden rules, where each frame pairs a caption describing the preceding action with an image showing the resulting state, and the final frame marks a terminal status (success, failure, or voluntary stop). These clips serve as the task's in-context behavioral exemplars. 
(ii) \textbf{Supplementary textual rules}, which by default cover only non-visual constraints that cannot be inferred from the videos---hidden scoring mechanisms, latent variables, step limits, and delayed action effects---so that the videos remain the sole source of visually inferable rule knowledge.
(iii) The \textbf{current visual state} that the agent acts upon.
(iv) The \textbf{action space}, which is the set of admissible actions at each step. 

\paragraph{Subsequent turns (multi-turn decision-making).}
No additional demonstrations are provided in subsequent steps. The agent receives only the updated visual state resulting from its previous action and must determine the next move. By preserving the entire interaction history, which encompasses the initial demonstration package, past observations, and previous model outputs, the context window naturally forms a long-context evaluation scenario. At each turn, the model is required to output an action in the exact format of \texttt{[Action:<name>]}. 
Complete parsing and retry protocols are detailed in Appendix~\ref{app:task-protocol}.

\paragraph{Objective.}
This benchmark is deliberately designed to evaluate the multimodal agent holistically. While the agent inherently relies on the parametric knowledge of its foundational model, the demonstration videos serve as the sole task-specific exemplars provided during evaluation. To succeed, the agent must synthesize these visual exemplars with the task identifier, available action space and minimal text constraints to deduce the operational rules, goals, and environment dynamics. Crucially, it must transfer this induced knowledge to novel states without any parameter updates or post-failure corrections. This end-to-end evaluation paradigm rigorously tests the complete agent decision loop: from visual comprehension and rule induction to state grounding, planning, and closed-loop execution.

\subsection{Scoring System}
\label{sec:scoring}

We evaluate agent performance from two complementary perspectives: Pass Rate and Final Score.


\textbf{Pass rate} is the fraction of episodes satisfying a task-specific success
condition (e.g., reaching a spatial goal, achieving a target score, or surviving
for a specified duration). It reflects whether the agent has the fundamental ability to
complete a task, following the achievement-style scoring of prior agency
benchmarks~\citep{paglieri2025balrog, hafner2021benchmarking}.


\textbf{Final score} provides a finer-grained, $[0,100]$-normalized measure of partial progress. We adopt an environment's native progress score when available, and otherwise map raw metrics (rewards, survival time, event counts, milestones, or path efficiency) onto this scale through environment-specific rules. Human experts calibrate task-level reference targets; where achievable progress depends on the initial state, targets are set per task configuration. Scores are strictly bounded, with explicit adjustments for negative rewards, agent death, and incomplete progress at termination. For aggregation, we average task scores within each environment and then weight environments equally, synthesizing the arithmetic averaging of \citet{paglieri2025balrog} with the per-environment normalization of \citet{park2026orak}. The details are given in Appendix~\ref{app:score-computation}.

Additionally, we track four explicit failure modes during execution: timeouts, continuous invalid actions, token overflows, and early terminations. 
Formal definitions and detailed handling procedures for these modes are provided in Appendix~\ref{app:failure-details}.

\section{Experiments}
\label{sec:experiments}


\subsection{Experiment Setting}

To guarantee controlled and efficient evaluations, we standardized the selected environments. This was achieved by enforcing uniform parameters, such as maximum step limits and frame-skipping rates, alongside targeted modifications that include normalizing termination signals and masking redundant actions. Full implementation details are provided in Appendix~\ref{app:environment-details}. We uniformly sample each visual trajectory at 1 FPS. For video-native models, the frames are encoded into an MP4 and provided through \textit{video\_url} (or Gemini’s \textit{inline\_data}), whereas non-video-native models such as GPT receive the exact same frames as a temporally ordered image sequence. The visual content, frame count, ordering, and sampling rate are identical across settings; only the input representation differs, ensuring that all models receive identical visual evidence. All configurations hide the scalar reward, so agents must infer progress from visual state changes alone. A malformed or out-of-space action triggers up to three format retries.

\subsection{Main Results}

\begin{table*}[t]
\centering
\setlength{\tabcolsep}{2.5pt}
\renewcommand{\arraystretch}{1.4}
\caption{\textbf{Main results on \benchname.} 
 \textsc{Memory Req.}: memory requirement for action selection. Single means that the current frame is sufficient; Multiple means the agent must retain one or more earlier post-action observations and integrate them to choose a later action. \textsc{Env.\ Dynamics}: whether the environment changes independently of the agent (Static vs.\ Dynamic). \textbf{S}: normalized score (0--100); \textbf{P\%}: pass rate. }
\resizebox{\linewidth}{!}{
\begin{tabular}{l||rrrrrrrrrrrr||rrrr||rrrr||rr}
\hline\hline
\multirow{3}{*}{\textbf{Model}}
 & \multicolumn{12}{c||}{\cellcolor{blue!8}\textsc{Environment Type}}
 & \multicolumn{4}{c||}{\cellcolor{orange!8}\textsc{Memory Req.}}
 & \multicolumn{4}{c||}{\cellcolor{green!8}\textsc{Env.\ Dynamics}}
 & \multicolumn{2}{c}{\cellcolor{purple!8}\textsc{\textbf{Overall}}} \\
 & \multicolumn{2}{c}{\cellcolor{blue!8}Maze}
 & \multicolumn{2}{c}{\cellcolor{blue!8}Shooter}
 & \multicolumn{2}{c}{\cellcolor{blue!8}Control}
 & \multicolumn{2}{c}{\cellcolor{blue!8}Dodge}
 & \multicolumn{2}{c}{\cellcolor{blue!8}Collect}
 & \multicolumn{2}{c||}{\cellcolor{blue!8}Sports}
 & \multicolumn{2}{c}{\cellcolor{orange!8}Single}
 & \multicolumn{2}{c||}{\cellcolor{orange!8}Multiple}
 & \multicolumn{2}{c}{\cellcolor{green!8}Static}
 & \multicolumn{2}{c||}{\cellcolor{green!8}Dynamic}
 & \multicolumn{2}{c}{\cellcolor{purple!8}\textsc{\textbf{All}}} \\
 & \multicolumn{1}{c}{\scriptsize S} & \multicolumn{1}{c}{\scriptsize P\%}
 & \multicolumn{1}{c}{\scriptsize S} & \multicolumn{1}{c}{\scriptsize P\%}
 & \multicolumn{1}{c}{\scriptsize S} & \multicolumn{1}{c}{\scriptsize P\%}
 & \multicolumn{1}{c}{\scriptsize S} & \multicolumn{1}{c}{\scriptsize P\%}
 & \multicolumn{1}{c}{\scriptsize S} & \multicolumn{1}{c}{\scriptsize P\%}
 & \multicolumn{1}{c}{\scriptsize S} & \multicolumn{1}{c||}{\scriptsize P\%}
 & \multicolumn{1}{c}{\scriptsize S} & \multicolumn{1}{c}{\scriptsize P\%}
 & \multicolumn{1}{c}{\scriptsize S} & \multicolumn{1}{c||}{\scriptsize P\%}
 & \multicolumn{1}{c}{\scriptsize S} & \multicolumn{1}{c}{\scriptsize P\%}
 & \multicolumn{1}{c}{\scriptsize S} & \multicolumn{1}{c||}{\scriptsize P\%}
 & \multicolumn{1}{c}{\scriptsize S} & \multicolumn{1}{c}{\scriptsize P\%} \\
\hline
\rowcolor{gray!10}\multicolumn{23}{l}{\textit{Closed-source Models}} \\
\hline
Seed-2.1-Pro~\citeyearpar{seed21} & 64.4 & 70 & 43.3 & 49 & 88.4 & 73 & 47.1 & 51 & 52.3 & 60 & 54.3 & 80 & 68.9 & 70 & 52.7 & 57 & 74.3 & 70 & 47.0 & 54 & \textbf{54.4} & \textbf{58} \\
Gemini-3.6-Flash~\citeyearpar{gemini36flash} & 61.0 & 70 & 37.9 & 39 & 88.1 & 80 & 39.9 & 44 & 42.2 & 63 & 29.7 & 50 & 67.1 & 70 & 47.3 & 52 & 77.2 & 75 & 39.2 & 45 & \textbf{49.5} & \textbf{53} \\
Gemini-3.1-Pro~\citeyearpar{gemini31pro} & 49.1 & 54 & 44.9 & 46 & 65.5 & 53 & 42.0 & 48 & 40.3 & 45 & 34.3 & 70 & 61.5 & 70 & 46.4 & 51 & 67.7 & 69 & 40.8 & 47 & \textbf{48.0} & \textbf{53} \\
GPT-5.6-sol~\citeyearpar{gpt56systemcard} & 52.0 & 53 & 37.6 & 43 & 82.7 & 87 & 38.1 & 43 & 41.5 & 43 & 35.0 & 60 & 82.5 & 85 & 42.4 & 47 & 68.3 & 67 & 38.8 & 46 & \textbf{46.7} & \textbf{51} \\
Gemini-3-Flash~\citeyearpar{gemini3flash} & 36.9 & 37 & 37.6 & 39 & 64.5 & 67 & 34.7 & 36 & 31.1 & 40 & 34.3 & 60 & 52.2 & 50 & 38.1 & 41 & 54.7 & 53 & 34.0 & 38 & \textbf{39.6} & \textbf{42} \\
Seed-2.0-Lite~\citeyearpar{seed2026seed20modelcardintelligence} & 39.0 & 51 & 26.5 & 24 & 86.3 & 80 & 31.4 & 36 & 36.2 & 48 & 32.0 & 60 & 54.2 & 60 & 34.9 & 39 & 50.6 & 49 & 31.9 & 38 & \textbf{37.0} & \textbf{41} \\
\hline
\rowcolor{gray!10}\multicolumn{23}{l}{\textit{Open-weight Models}} \\
\hline
DeepSeek-V4.1-Flash~\citeyearpar{deepseekv41flash} & 34.1 & 36 & 28.7 & 26 & 71.6 & 67 & 32.4 & 33 & 26.0 & 30 & 32.3 & 50 & 46.1 & 45 & 33.5 & 34 & 51.0 & 50 & 28.8 & 30 & \textbf{34.8} & \textbf{35} \\

Kimi-K3~\citeyearpar{kimiteam2026kimik3openfrontier} & 32.3 & 40 & 26.4 & 34 & 70.1 & 73 & 27.2 & 34 & 25.4 & 30 & 26.3 & 40 & 51.8 & 55 & 30.2 & 37 & 46.7 & 46 & 27.3 & 36 & \textbf{32.5} & \textbf{39} \\
Qwen3.5-397B-A17B~\citeyearpar{qwen3.5} & 28.2 & 33 & 30.3 & 27 & 73.6 & 67 & 28.1 & 27 & 25.4 & 30 & 25.7 & 40 & 44.2 & 45 & 31.1 & 30 & 45.0 & 42 & 27.9 & 27 & \textbf{32.5} & \textbf{31} \\
DeepSeek-V4F-V~\citeyearpar{deepseek-4v} & 21.0 & 29 & 26.0 & 34 & 47.1 & 53 & 24.1 & 30 & 19.3 & 30 & 21.7 & 30 & 26.6 & 15 & 25.0 & 32 & 30.1 & 26 & 23.4 & 32 & \textbf{25.2} & \textbf{30} \\

gemma-4-31B-it~\citeyearpar{gemmateam2026gemma4} & 20.4 & 28 & 17.7 & 29 & 32.8 & 27 & 19.7 & 25 & 14.4 & 18 & 13.7 & 10 & 40.1 & 35 & 19.7 & 24 & 29.8 & 25 & 18.9 & 25 & \textbf{21.9} & \textbf{25} \\
MiMo-v2.5~\citeyearpar{mimov25} & 20.0 & 21 & 18.3 & 26 & 34.6 & 27 & 21.4 & 27 & 13.1 & 28 & 16.0 & 20 & 25.6 & 15 & 19.8 & 24 & 22.2 & 14 & 19.8 & 26 & \textbf{20.4} & \textbf{23} \\
Qwen2.5-VL-72B~\citeyearpar{qwen2.5-VL} & 14.3 & 16 & 16.3 & 35 & 20.1 & 7 & 20.4 & 31 & 11.5 & 25 & 10.3 & 0 & 19.0 & 10 & 17.5 & 25 & 16.9 & 8 & 17.9 & 29 & \textbf{17.7} & \textbf{23} \\
MiMo-VL-7B~\citeyearpar{coreteam2025mimovltechnicalreport} & 13.9 & 15 & 15.8 & 10 & 19.7 & 13 & 19.9 & 16 & 13.3 & 10 & 12.7 & 10 & 20.6 & 20 & 17.0 & 13 & 17.1 & 14 & 17.5 & 14 & \textbf{17.4} & \textbf{14} \\
Qwen3.6-27B~\citeyearpar{qwen3.6-27b} & 10.5 & 18 & 13.8 & 36 & 17.5 & 7 & 19.2 & 33 & 12.2 & 28 & 22.7 & 40 & 17.3 & 10 & 16.8 & 29 & 14.8 & 8 & 17.6 & 34 & \textbf{16.9} & \textbf{27} \\
GLM-4.5V~\citeyearpar{vteam2025glm45vglm41vthinkingversatilemultimodal} & 10.9 & 15 & 13.1 & 27 & 28.7 & 27 & 18.1 & 26 & 11.9 & 15 & 12.0 & 10 & 17.2 & 5 & 16.5 & 23 & 16.6 & 10 & 16.6 & 25 & \textbf{16.6} & \textbf{21} \\
Qwen3.5-27B~\citeyearpar{qwen3.5} & 13.3 & 15 & 12.4 & 35 & 22.4 & 13 & 17.8 & 32 & 12.2 & 20 & 20.0 & 40 & 18.5 & 0 & 16.2 & 27 & 14.2 & 4 & 17.3 & 32 & \textbf{16.4} & \textbf{24} \\
Qwen3-Omni-30B~\citeyearpar{xu2025qwen3omnitechnicalreport} & 9.7 & 13 & 5.8 & 20 & 19.5 & 13 & 15.6 & 27 & 5.3 & 18 & 16.7 & 30 & 19.6 & 10 & 12.8 & 22 & 16.9 & 10 & 12.3 & 24 & \textbf{13.6} & \textbf{20} \\
Qwen3-VL-235B~\citeyearpar{qwen3technicalreport} & 12.5 & 6 & 6.9 & 3 & 19.7 & 13 & 11.7 & 5 & 11.6 & 5 & 15.3 & 10 & 21.6 & 5 & 11.2 & 5 & 14.2 & 6 & 11.6 & 5 & \textbf{12.3} & \textbf{5} \\
\hline
\rowcolor{gray!10}\multicolumn{23}{l}{\textit{Reference Baselines}} \\
\hline
Random & 7.9 & 10 & 19.1 & 23 & 16.8 & 13 & 15.5 & 15 & 15.0 & 15 & 19.4 & 30 & 5.6 & 2 & 15.0 & 15 & 8.4 & 5 & 16.1 & 17 & \textbf{14.0} & \textbf{14} \\
Human & 92.1 & 96 & 76.4 & 85 & 94.6 & 89 & 82.8 & 90 & 91.3 & 99 & 69.8 & 97 & 99.0 & 100 & 81.8 & 90 & 98.1 & 99 & 78.3 & 88 & \textbf{83.6} & \textbf{91} \\
\hline\hline
\end{tabular}
}
\vspace{1mm}
\label{tab:main-results}
\end{table*}

\paragraph{Task Classification.}

The main evaluation results are presented in Table~\ref{tab:main-results}. Besides the \textbf{Environment Type}, we introduce two attributes to characterize task complexity: 
\textbf{Memory requirement} distinguishes whether tasks can be solved from the current frame alone (Single-frame, e.g., Taxi) or require integrating historical observations (Multi-frame, e.g., FlappyBird rhythm judgment, partially observable MiniGrid). \textbf{Environment dynamics} categorizes tasks by whether state transitions depend solely on agent actions (Static, e.g., Toy Text, MiniGrid) or evolve autonomously over time (Dynamic, e.g., ALE Atari, Procgen).

\paragraph{Overall Performance.}
As illustrated in Table~\ref{tab:main-results}, \textbf{Seed-2.1-Pro} achieves the best overall performance, securing a normalized score of \textbf{54.4} and a pass rate of \textbf{58\%}. Notably, even this leading score only marginally surpasses the halfway mark of the normalized scale. While it significantly outperforms the random baseline (14.0) by 40.4 points, it still falls 45.6 points short of the theoretical maximum. This substantial performance gap underscores the profound difficulty of end-to-end video ICL within closed-loop interactive environments.
Trailing the leading model are Gemini-3.6-Flash (49.5 score / 53\% pass rate), Gemini-3.1-Pro (48.0 / 53\%), and GPT-5.6-sol (46.7 / 51\%). Within the open-weight category, it is noted that DeepSeek-V4.1-Flash achieves the highest normalized score at 34.8 with a 35\% pass rate, whose gap with the closed-source model Seed-2.0-Lite has become quite small. Kimi-K3 and Qwen3.5-397B-A17B tie for second at 32.5. The best open-weight score remains 19.6 points below Seed-2.1-Pro. The random policy achieves a score of 14.0 with a 14\% pass rate, establishing a non-trivial baseline floor for the benchmark.

\paragraph{Limitation in Memory.}
All fully evaluated models exhibit a  consistent performance degradation on memory-intensive tasks. Among the four strongest models, Seed-2.1-Pro falls from 68.9 to 52.7, Gemini-3.6-Flash from 67.1 to 47.3, Gemini-3.1-Pro from 61.5 to 46.4, and notably, GPT-5.6-sol drops from 82.5 to 42.4. All 19 models score lower on memory-dependent (Multiple) tasks than on single-frame tasks, and the drop is pronounced for all six closed-source systems and the strongest open-weight models (12.6--40.1 points), whereas weaker models, already near the random floor, decline only marginally. It is also unlikely to be a score-scale artifact: the random policy shows the opposite ordering on both splits (Single 5.6 vs. Multiple 15.0; Static 8.4 vs. Dynamic 16.1).
Furthermore, as corroborated by the case audits in Section~\ref{sec:trajectory-audit}, this pervasive performance decline exposes a critical bottleneck in agentic visual memory. Specifically, current multimodal agents struggle to retain and synthesize the visual consequences of their own actions—a dynamic capability that traditional offline video QA benchmarks inherently fail to measure.

\paragraph{Vulnerability in Dynamic Environments.}
A similarly consistent weakness emerges in autonomously evolving environments. For instance, Seed-2.1-Pro's score drops from 74.3 in static settings to 47.0 in dynamic ones, while Gemini-3.6-Flash experiences a steep decline from 77.2 to 39.2. Because dynamic environments change independently of the agent's interventions, they demand robust continuous replanning and real-time state tracking. This vulnerability is universally observed across all six closed-source systems and the three strongest open-weight models. Consequently, this bottleneck highlights a profound gap between passive video comprehension and active agency: to achieve true autonomy, agents must possess the capability to process and react while the world continuously evolves around them.


\paragraph{Comparison with Human Performance.}
We establish a human baseline following \cite{paglieri2025balrog, park2026orak}: three participants per environment complete tasks under exactly the same conditions (including the in-context demonstrations), environments, and metrics as MLLM agents.
Results reveal a substantial capability gap: the strongest model (Seed-2.1-Pro) trails humans by 29.2 points (54.4 vs. 83.6) and 33 percentage points in pass rate (58\% vs. 91\%).
Interestingly, humans and models exhibit aligned performance degradation across task complexity—human scores drop from 99.0 to 81.8 (Single- to Multi-frame) and 98.1 to 78.3 (Static to Dynamic), mirroring model trends. This alignment suggests temporal memory and environment dynamics pose intrinsic difficulty rather than model-specific vulnerabilities. However, the persistent absolute gap underscores that current agents remain far less robust than humans in handling these fundamental challenges.

\begin{figure}[t]
    \centering
    \includegraphics[width=\textwidth]{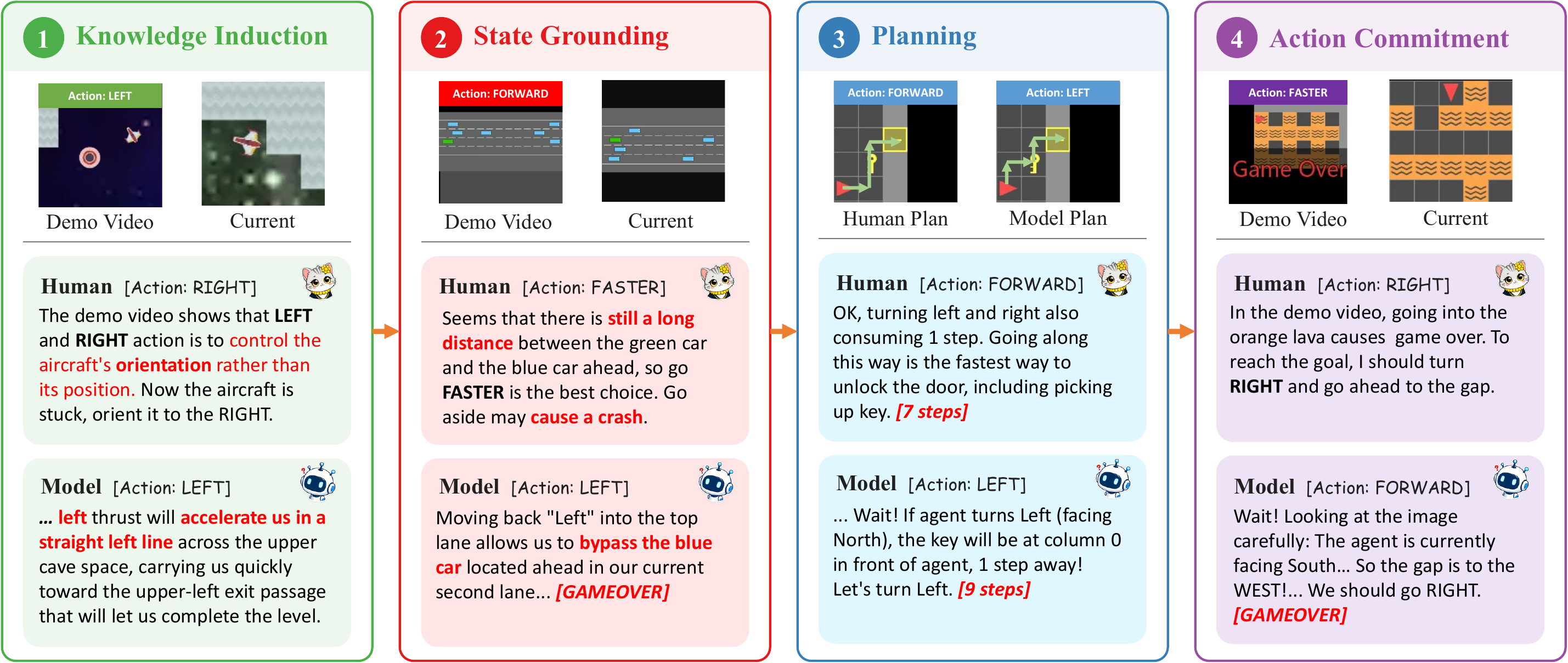}
    \caption{Representative failures across the four-stage decision-making process. Each
    panel contrasts the human decision with the model response at the
    corresponding deficient stage.}
    \label{fig:four-stage-pipeline}
\end{figure}

\subsection{Fine-Grained Auditing of Interaction Trajectories}
\label{sec:trajectory-audit}

The aggregate results above quantify overall benchmark performance; to obtain
finer-grained diagnostic profiles of model behavior during environment
interaction, we introduce a trajectory-auditing framework built around the
following staged model of the acting agent's decision-making process:
knowledge induction $\rightarrow$ state grounding $\rightarrow$ planning $\rightarrow$ action commitment.
\textbf{Knowledge induction} extracts a reusable task model from
the demonstration, including rules and strategies inferred from environment
dynamics that encompass both action-conditioned environmental responses and
reusable state--action--outcome regularities.
\textbf{State grounding} maps induced knowledge onto live visual evidence to estimate
relevant entities, relations, motion, and progress. \textbf{Planning} turns the
grounded state into an adaptive course of action through lookahead and revision. \textbf{Action commitment} converts
the selected step into the submitted action. The labels serve as rubric-guided
functional attributions from observable trajectory evidence rather than as
direct measurements of the model's internal computation.
Figure~\ref{fig:four-stage-pipeline} illustrates representative failures at each stage.

We analyze this process using two complementary criteria: \textbf{information
transfer evaluation} tracks demonstrated rules and strategies separately across
the complete decision-making chain and produces an end-to-end judgment for each
information type based on whether it is both induced and appropriately used,
while \textbf{sequential failure attribution} examines the chain stage by stage
and identifies the most upstream deficient stage that materially explains
unsuccessful or substantially suboptimal behavior.
Taken together, they provide two cross-cutting perspectives on each
trajectory. 
To apply these criteria, we develop an automated pipeline using a multimodal LLM auditor.
Trajectories from representative models are examined to gain insight into
their behavior.
Blind human verification of 333 stratified trajectories agrees with the automated labels on 95.0\% of decisions, with Cohen's $\kappa$ between 0.861 and 0.925. 
Appendix~\ref{app:diagnostic-setup} provides the complete protocol, prompts, and verification results.

\subsubsection{Information Transfer Evaluation}
\label{sec:learner-audit}

Information transfer is evaluated along two dimensions. \textbf{Rule transfer}
covers demonstrated environment mechanics and task requirements, while \textbf{strategy transfer} covers reusable
state--action--outcome regularities such as timing, sequencing, target selection,
risk management, and adaptation. Both are successful only when the relevant
information is induced and appropriately applied during environment interaction.
These judgments assess behavioral use of demonstrated information, not whether
the video was its exclusive source.
Table~\ref{tab:audit-learner-results} reports the results for both dimensions.

\begin{table}[t]
    \caption{Information transfer rates (\%), macro-averaged over 37 environments.}
    \label{tab:audit-learner-results}
    \tablestyle{4pt}{1.12}
    \begin{tabular}{
        @{}
        >{\raggedright\arraybackslash}p{0.22\linewidth}
        >{\centering\arraybackslash}p{0.07\linewidth}
        >{\centering\arraybackslash}p{0.08\linewidth}
        >{\centering\arraybackslash}p{0.09\linewidth}
        >{\centering\arraybackslash}p{0.09\linewidth}
        >{\centering\arraybackslash}p{0.09\linewidth}
        >{\centering\arraybackslash}p{0.09\linewidth}
        @{}
    }
        \toprule
        \multirow[c]{2}{*}[-0.75ex]{\textbf{Model}}
        & \multirow[c]{2}{*}[-0.75ex]{\textbf{Score}}
        & \multirow[c]{2}{*}[-0.75ex]{\makecell[c]{\textbf{Pass}\\\textbf{rate}}}
        & \multicolumn{2}{c}{\textbf{Rule transfer}}
        & \multicolumn{2}{c}{\textbf{Strategy transfer}} \\
        \cmidrule(lr){4-5}\cmidrule(lr){6-7}
        & & & \textbf{Succ.} & \textbf{Unsucc.}
        & \textbf{Succ.} & \textbf{Unsucc.} \\
        \midrule
        Seed-2.1-Pro          & 54.4 & 58 & 77.3 & 22.7 & \textbf{31.7} & \textbf{68.3} \\
        Gemini-3.6-Flash      & 49.5 & 53 & \textbf{84.3} & \textbf{15.7} & 29.0 & 71.0 \\
        Qwen3.5-397B-A17B     & 32.5 & 31 & 70.0 & 30.0 & 14.4 & 85.6 \\
        \bottomrule
    \end{tabular}
\end{table}

Strategy transfer lags well behind rule transfer across all three models. Seed-2.1-Pro and Gemini-3.6-Flash are broadly comparable, but Gemini leads in rule transfer and Seed in strategy transfer, which may contribute to Seed's higher benchmark score. Qwen3.5-397B-A17B trails on both.

\subsubsection{Sequential Failure Attribution}
\label{sec:player-audit}

In this audit, the auditor identifies the behavior contributing most to an
unsuccessful or substantially suboptimal outcome, then traces its
evidence-supported decision-making chain to locate the most upstream
consequentially deficient stage. A downstream stage is assigned only when
all preceding stages are adequately supported. Harmless or corrected errors
are ignored; \texttt{none} is assigned when no consequential deficiency is
found. Table~\ref{tab:audit-player-results} reports the results.

\begin{table}[t]
\caption{Sequential failure attribution (\%), macro-averaged over 37 environments.}
\label{tab:audit-player-results}

\centering
\small

\begin{tabular*}{\columnwidth}{
@{\extracolsep{\fill}}
l
c c
c c c c c
}
\toprule
\textbf{Model}
&
\textbf{Score}
&
\textbf{Pass}
&
\textbf{Knowledge}
&
\textbf{State}
&
\textbf{Planning}
&
\textbf{Commitment}
&
\textbf{None}
\\
&
&
\textbf{rate}
&
\textbf{induction}
&
\textbf{grounding}
&
&
&
\\
\midrule

Seed-2.1-Pro
&54.4&58&10.8&27.6&\textbf{21.5}&\textbf{0.0}&\textbf{40.1}
\\

Gemini-3.6-Flash
&49.5&53&\textbf{2.7}&\textbf{24.1}&40.9&1.8&30.6
\\

Qwen3.5-397B-A17B
&32.5&31&14.4&44.6&24.3&0.7&15.9
\\

\bottomrule
\end{tabular*}

\vskip -0.05in
\end{table}

Seed has the highest \texttt{none} rate, consistent with its stronger benchmark
performance, whereas Gemini receives the fewest knowledge-induction and
state-grounding attributions, indicating relatively stronger upstream processing.
Qwen has the highest attribution rates at these upstream stages and the lowest
\texttt{none} rate.
Across models, state grounding and planning dominate failure attributions,
while action commitment is rare.

\subsection{Ablation Study: Efficacy of In-Context Demonstrations and Rule Specificity}
We ablate video availability and textual rule specificity across four models (Table~\ref{tab:ablation-video-rules}). Detailed rules provide expert-written descriptions of visual context, task mechanics, and strategies.

As Table~\ref{tab:ablation-video-rules} shows, video benefits depend on how detailed the rules are. Under base rules, demonstration videos improve normalized scores for three of the four evaluated models, with substantial gains for Gemini-3.6-Flash (+4.7) and Seed-2.0-Lite (+7.1). Gemini-3-Flash is the only exception, showing a slight decline. These results indicate clear video in-context learning capabilities under basic textual instructions. Under detailed rules, video and no-video performance differs only marginally across all four models, suggesting diminished benefits from visual demonstrations when comprehensive task information is already available in text.

\begin{table}[h]
\centering
\caption{\textbf{Impact of demonstration video and rule specification on end-to-end agent performance.} \textit{Base + Video} is the baseline. Each cell reports normalized score (\textbf{S}) / pass rate (\textbf{P\%}).}
\label{tab:ablation-video-rules}
\footnotesize
\setlength{\tabcolsep}{6.5pt}
\renewcommand{\arraystretch}{1.16}
\begin{tabular}{@{}lcccc@{}}
\toprule
\multicolumn{1}{c}{\multirow[c]{2}{*}{\textbf{Model}}} &
\multicolumn{2}{c}{\textbf{Base Rule}} &
\multicolumn{2}{c}{\textbf{Detailed Rule}} \\
\cmidrule(lr){2-3}\cmidrule(l){4-5}
 & \textbf{Video} & \textbf{No Video}
 & \textbf{Video} & \textbf{No Video} \\
\midrule
Gemini-3.6-Flash & \textbf{49.5 / 53} & 44.8 / 48 & 48.4 / 51 & 46.8 / 50 \\
Gemini-3-Flash   & 39.6 / 42 & 41.2 / 46 & \textbf{44.1 / 50} & 43.6 / 47 \\
Seed-2.0-Lite    & 37.0 / 41 & 29.9 / 36 & 37.2 / 42 & \textbf{37.2 / 43} \\
Qwen3.5-397B-A17B & 32.5 / 31 & 30.7 / 37 & 33.9 / 36 & \textbf{34.2 / 39} \\
\bottomrule
\end{tabular}
\end{table}

Under the no-video condition, replacing base rules with detailed rules consistently improves performance across all four models, yielding gains of 2.0–7.3 points in normalized scores (Table~\ref{tab:ablation-video-rules}). This improvement demonstrates that detailed rules can partially compensate for the missing video by making task-relevant mechanics explicit. 

Importantly, models perform non-trivially even without videos or detailed rules, possibly owing to the rich prior knowledge of their foundation models. Thus, the no-video condition does not evaluate a “tabula rasa” agent; instead, it rigorously quantifies the marginal utility of in-context video demonstrations beyond pre-training priors. A closer look at the trajectories shows when this marginal utility is positive and when it is negative.

\textbf{Visual Demonstrations Can Impair Execution Feedback and State Grounding.}
In spatial navigation environments such as Procgen Heist and CustomMultiRoom, trajectory analysis reveals less reliable responses to execution feedback in video-conditioned agents, including misinterpreting unsuccessful actions as successful despite contrary observations. Agents without demonstrations more reliably recognize action outcomes and adjust subsequent behavior. These findings show that providing demonstrations does not necessarily improve closed-loop interaction and can impair responsiveness to feedback and grounding in the current state.

\textbf{The Dual Role of Demonstrations in Inferring Physical Dynamics.}
In CartPole, models without demonstrations can balance the pole using prior physical knowledge, whereas video-conditioned models can be destabilized by errors in estimating its orientation. In Acrobot, the less intuitive, underactuated dynamics pose a different challenge: models without demonstrations often misinterpret the physics or fail to infer motion quantities needed for control, while demonstrations provide useful spatiotemporal cues linking actions to system responses. These cases suggest that demonstrations help when they supplement insufficient prior knowledge, but can hurt when visual estimation errors outweigh the information gained.

\section{Conclusion}
\label{sec:conclusion}

We introduced \benchname, a benchmark for video in-context learning by multimodal agents, in which each task requires inducing rules and strategies from video exemplars and executing them through sustained closed-loop interaction. Across 342 tasks and 19 state-of-the-art multimodal agents, the strongest agent reaches only 54.4/100, and demonstrations yield no consistent aggregate gains. This exposes a gap between placing behavioral evidence in context and turning it into a usable policy. Two bottlenecks recur across model families: visual memory over the consequences of the agent's own actions, and adaptation in autonomously evolving environments. Our four-stage decision-loop taxonomy further separates failures of knowledge induction from downstream failures in state grounding, planning, and action commitment. Unlike offline ICL, which can end with a correct prediction, an agent must carry induced knowledge through self-generated experience and non-stationary worlds, and these capabilities remain largely unsolved. We view learning from watching as a cornerstone of general intelligence: humans acquire skills from a few demonstrations, and the web already stores such demonstrations at unmatched scale. By making this capability measurable and diagnosable, \benchname{} points to persistent visual memory and test-time adaptation as the path from agents that follow instructions to agents that learn on the fly.
\section*{AI Use Statement}

\textbf{Required disclosures.} We used generative AI to assist with method
implementation, dataset cleaning and reformatting, result interpretation,
and manuscript translation. Assistance with interpretation was limited to
summarizing trends in figures and tables and proposing candidate
explanations for the authors to evaluate. Qualitative analysis of agent
trajectories was performed by our LLM-based trajectory auditor, a
component of our method documented in Appendix~\ref{app:diagnostic-setup}.
AI was not used to generate synthetic data, develop theory or conceptual
frameworks, propose hypotheses, or design experiments; mathematical
claims and proofs are not applicable to this work.

\textbf{Recommended disclosures.} AI tools were also used for writing
code (including the released codebase and its documentation), figure
styling and layout, literature summarization, information search, and
readability editing. They were not used for ideation, identifying
literature or research gaps, choosing experimental parameters,
structuring or titling the paper, formatting references, or translating
dataset content.

\textbf{Verification.} All AI-assisted code was reviewed line by line by
at least three authors, tested with unit and end-to-end tests, and
cross-validated against independent results. Cleaned data were checked
in full against the raw logs, with automated checks of record counts and
formats. AI-proposed interpretations were verified against the original
trajectories and adopted only after discussion among the authors.
Literature summaries and search results were checked against their
sources, figure edits did not alter any plotted values, and translations
were cross-checked by multiple authors. The authors take full
responsibility for all content of this paper.

\section*{Ethics Statement}

This work involves human gameplay demonstrations, human baseline
evaluation, and human verification of automated trajectory annotations.
All game environments used in this work are open-source, and we use and
redistribute them in accordance with their respective licenses.
The human demonstration and baseline data will be publicly released
in de-identified form, with personally identifying information removed.
Our benchmark evaluates AI agents in simulated game environments and
involves no real-world deployment; we do not anticipate negative
societal impacts beyond those common to general research on autonomous
agents.

\section*{Reproducibility Statement}

Upon acceptance, we will publicly release the benchmark environments and
configuration files, the seeds and initialization settings used to
construct the task pool, the evaluation harness with all interaction
prompts, the scoring code, the trajectory-auditor prompts and
configuration, and documented scripts that reproduce the main results
step by step. We will also release the raw trajectories and model
responses underlying every reported result, so that all scores and
audit annotations can be recomputed without re-querying model APIs,
whose behavior may change or become unavailable over time. The human
demonstration and baseline data will be released in de-identified form,
as described in the Ethics Statement. Data will be released under the
CC BY 4.0 license and code under an open-source license.

Section~\ref{sec:scoring} describes the evaluation metrics.
Appendix~\ref{app:benchmark-construction} provides environment-selection
criteria, environment configurations, and demonstration-construction
procedures. Appendix~\ref{app:experimental-setup} and our repo details the evaluation
protocol, the exact model identifiers and versions queried,
provider-specific input representations, score computation, and
complete interaction prompts. Appendix~\ref{app:diagnostic-setup}
specifies the automated trajectory auditor, its inference configuration
and prompts, and the human-verification procedure.

\bibliographystyle{\PaperBibStyle}
\bibliography{references}

@article{agarwal2024many,
  title={Many-shot in-context learning},
  author={Agarwal, Rishabh and Singh, Avi and Zhang, Lei and Bohnet, Bernd and Rosias, Luis and Chan, Stephanie and Zhang, Biao and Anand, Ankesh and Abbas, Zaheer and Nova, Azade and others},
  journal={Advances in Neural Information Processing Systems},
  volume={37},
  pages={76930--76966},
  year={2024}
}

@article{jiang2024many,
  title={Many-shot in-context learning in multimodal foundation models},
  author={Jiang, Yixing and Irvin, Jeremy and Wang, Ji Hun and Chaudhry, Muhammad Ahmed and Chen, Jonathan H and Ng, Andrew Y},
  journal={arXiv preprint arXiv:2405.09798},
  year={2024}
}

@inproceedings{zong2025vl,
  title={Vl-icl bench: The devil in the details of multimodal in-context learning},
  author={Zong, Yongshuo and Bohdal, Ondrej and Hospedales, Timothy},
  booktitle={International Conference on Learning Representations},
  volume={2025},
  pages={100058--100100},
  year={2025}
}

@article{ruoss2024lmact,
  title={Lmact: A benchmark for in-context imitation learning with long multimodal demonstrations},
  author={Ruoss, Anian and Pardo, Fabio and Chan, Harris and Li, Bonnie and Mnih, Volodymyr and Genewein, Tim},
  journal={arXiv preprint arXiv:2412.01441},
  year={2024}
}

@inproceedings{jang2025videowebarena,
  title={Videowebarena: Evaluating long context multimodal agents with video understanding web tasks},
  author={Jang, Lawrence and Li, Yinheng and Zhao, Dan and Ding, Charles and Lin, Justin and Liang, Paul Pu and Bonatti, Rogerio and Koishida, Kazuhito},
  booktitle={International Conference on Learning Representations},
  volume={2025},
  pages={36934--36958},
  year={2025}
}

@article{zhang2026gameverse,
  title={GameVerse: Can Vision-Language Models Learn from Video-based Reflection?},
  author={Zhang, Kuan and Liu, Dongchen and Zhao, Qiyue and Hou, Jinkun and Zhang, Xinran and Xie, Qinlei and Liu, Miao and Li, Yiming},
  journal={arXiv preprint arXiv:2603.06656},
  year={2026}
}

@inproceedings{paglieri2025balrog,
  title={Balrog: Benchmarking agentic llm and vlm reasoning on games},
  author={Paglieri, Davide and Cupia{\l}, Bart{\l}omiej and Coward, Samuel and Piterbarg, Ulyana and Wo{\l}czyk, Maciej and Khan, Akbir and Pignatelli, Eduardo and Kuci{\'n}ski, {\L}ukasz and Pinto, Lerrel and Fergus, Rob and others},
  booktitle={International Conference on Learning Representations},
  volume={2025},
  pages={96666--96702},
  year={2025}
}

@article{hafner2021benchmarking,
  title={Benchmarking the spectrum of agent capabilities},
  author={Hafner, Danijar},
  journal={arXiv preprint arXiv:2109.06780},
  year={2021}
}

@inproceedings{liu2024agentbench,
  title={Agentbench: Evaluating llms as agents},
  author={Liu, Xiao and Yu, Hao and Zhang, Hanchen and Xu, Yifan and Lei, Xuanyu and Lai, Hanyu and Gu, Yu and Ding, Hangliang and Men, Kaiwen and Yang, Kejuan and others},
  booktitle={International Conference on Learning Representations},
  volume={2024},
  pages={52989--53046},
  year={2024}
}

@inproceedings{hu2026lmgame,
  title={lmgame-bench: How good are llms at playing games?},
  author={Hu, Lanxiang and Huo, Mingjia and Zhang, Yuxuan and Yu, Haoyang and Xing, Eric P and Stoica, Ion and Rosing, Tajana and Jin, Haojian and Zhang, Hao},
  booktitle={International Conference on Learning Representations},
  volume={2026},
  pages={81308--81356},
  year={2026}
}

@inproceedings{park2026orak,
  title={Orak: A foundational benchmark for training and evaluating llm agents on diverse video games},
  author={Park, Dongmin and Kim, Minkyu and Choi, Beongjun and Kim, Junhyuck and Lee, Keon and Lee, Jonghyun and Park, Inkyu and Lee, ByeongUk and Hwang, Jaeyoung and Ahn, Jaewoo and others},
  booktitle={International Conference on Learning Representations},
  volume={2026},
  pages={60684--60744},
  year={2026}
}

@article{fu2026video,
  title={Video-MME-v2: Towards the next stage in benchmarks for comprehensive video understanding},
  author={Fu, Chaoyou and Yuan, Haozhi and Dong, Yuhao and Zhang, Yi-Fan and Shen, Yunhang and Hu, Xiaoxing and Li, Xueying and Su, Jinsen and Long, Chengwu and Xie, Xiaoyao and others},
  journal={arXiv preprint arXiv:2604.05015},
  year={2026}
}

@article{dong2026demo,
  title={Demo-icl: In-context learning for procedural video knowledge acquisition},
  author={Dong, Yuhao and Tian, Shulin and Liu, Shuai and Ding, Shuangrui and Zang, Yuhang and Dong, Xiaoyi and Cao, Yuhang and Wang, Jiaqi and Liu, Ziwei},
  journal={arXiv preprint arXiv:2602.08439},
  year={2026}
}

@article{waytowich2024atari,
  title={Atari-gpt: Benchmarking multimodal large language models as low-level policies in atari games},
  author={Waytowich, Nicholas R and White, Devin and Sunbeam, MD and Goecks, Vinicius G},
  journal={arXiv preprint arXiv:2408.15950},
  year={2024}
}

@article{ouyang2026gameworld,
  title={GameWorld: Towards standardized and verifiable evaluation of multimodal game agents},
  author={Ouyang, Mingyu and Hu, Siyuan and Lin, Kevin Qinghong and Ng, Hwee Tou and Shou, Mike Zheng},
  journal={arXiv preprint arXiv:2604.07429},
  year={2026}
}

@inproceedings{fu2025video,
  title={Video-mme: The first-ever comprehensive evaluation benchmark of multi-modal llms in video analysis},
  author={Fu, Chaoyou and Dai, Yuhan and Luo, Yongdong and Li, Lei and Ren, Shuhuai and Zhang, Renrui and Wang, Zihan and Zhou, Chenyu and Shen, Yunhang and Zhang, Mengdan and others},
  booktitle={2025 IEEE/CVF Conference on Computer Vision and Pattern Recognition (CVPR)},
  pages={24108--24118},
  year={2025},
  organization={IEEE}
}

@inproceedings{kim2025videoicl,
  title={Videoicl: Confidence-based iterative in-context learning for out-of-distribution video understanding},
  author={Kim, Kangsan and Park, Geon and Lee, Youngwan and Yeo, Woongyeong and Hwang, Sung Ju},
  booktitle={2025 IEEE/CVF Conference on Computer Vision and Pattern Recognition (CVPR)},
  pages={3295--3305},
  year={2025},
  organization={IEEE}
}

@misc{coreteam2025mimovltechnicalreport,
      title={MiMo-VL Technical Report}, 
      author={LLM-Core-Team Xiaomi},
      year={2025},
      eprint={2506.03569},
      archivePrefix={arXiv},
      primaryClass={cs.CL},
      url={https://arxiv.org/abs/2506.03569}, 
}

@misc{qwen3.5,
    title  = {{Qwen3.5}: Towards Native Multimodal Agents},
    author = {{Qwen Team}},
    month  = {February},
    year   = {2026},
    url    = {https://qwen.ai/blog?id=qwen3.5}
}

@misc{qwen3.6-27b,
    title  = {{Qwen3.6-27B}: Flagship-Level Coding in a {27B} Dense Model},
    author = {{Qwen Team}},
    month  = {April},
    year   = {2026},
    url    = {https://qwen.ai/blog?id=qwen3.6-27b}
}

@misc{gemmateam2026gemma4,
      title={Gemma 4 Technical Report}, 
      author={Gemma Team},
      year={2026},
      eprint={2607.02770},
      archivePrefix={arXiv},
      primaryClass={cs.CL},
      url={https://arxiv.org/abs/2607.02770}, 
}

@misc{xu2025qwen3omnitechnicalreport,
      title={Qwen3-Omni Technical Report}, 
      author={Jin Xu and Zhifang Guo and Hangrui Hu and Yunfei Chu and Xiong Wang and Jinzheng He and Yuxuan Wang and Xian Shi and Ting He and Xinfa Zhu and Yuanjun Lv and Yongqi Wang and Dake Guo and He Wang and Linhan Ma and Pei Zhang and Xinyu Zhang and Hongkun Hao and Zishan Guo and Baosong Yang and Bin Zhang and Ziyang Ma and Xipin Wei and Shuai Bai and Keqin Chen and Xuejing Liu and Peng Wang and Mingkun Yang and Dayiheng Liu and Xingzhang Ren and Bo Zheng and Rui Men and Fan Zhou and Bowen Yu and Jianxin Yang and Le Yu and Jingren Zhou and Junyang Lin},
      year={2025},
      eprint={2509.17765},
      archivePrefix={arXiv},
      primaryClass={cs.CL},
      url={https://arxiv.org/abs/2509.17765}, 
}

@article{qwen2.5-VL,
  title={Qwen2. 5-VL technical report. CoRR abs/2502.13923 (2025)},
  author={Bai, Shuai and Chen, Keqin and Liu, Xuejing and Wang, Jialin and Ge, Wenbin and Song, Sibo and Dang, Kai and Wang, Peng and Wang, Shijie and Tang, Jun and others},
  journal={arXiv preprint arXiv:2502.13923},
  year={2025}
}

@misc{vteam2025glm45vglm41vthinkingversatilemultimodal,
      title={GLM-4.5V and GLM-4.1V-Thinking: Towards Versatile Multimodal Reasoning with Scalable Reinforcement Learning}, 
      author={V Team},
      year={2025},
      eprint={2507.01006},
      archivePrefix={arXiv},
      primaryClass={cs.CV},
      url={https://arxiv.org/abs/2507.01006}, 
}

@misc{kimiteam2026kimik3openfrontier,
      title={Kimi K3: Open Frontier Intelligence}, 
      author={Kimi Team},
      year={2026},
      eprint={2607.24653},
      archivePrefix={arXiv},
      primaryClass={cs.CL},
      url={https://arxiv.org/abs/2607.24653}, 
}

@misc{mimov25,
  author={XiaomiMiMo},
  title={MiMo-V2.5},
  year={2026},
  howpublished={\url{https://huggingface.co/collections/XiaomiMiMo/mimo-v25}},
}

@techreport{deepseekv41flash,
institution = {DeepSeek-AI}, 
author={DeepSeek-AI},
year={2026},
  title  = {DeepSeek-V4.1-Flash: Pushing the Limits of KV Cache Compression},
  url    = {https://huggingface.co/deepseek-ai/DeepSeek-V4.1-Flash/blob/main/DeepSeek_V41_Tech_Report.pdf}
}

@article{qwen3technicalreport,
  title={Qwen3-vl technical report},
  author={Bai, Shuai and Cai, Yuxuan and Chen, Ruizhe and Chen, Keqin and Chen, Xionghui and Cheng, Zesen and Deng, Lianghao and Ding, Wei and Gao, Chang and Ge, Chunjiang and others},
  journal={arXiv preprint arXiv:2511.21631},
  year={2025}
}

@techreport{seed21,
institution = {ByteDance Seed}, 
author={ByteDance Seed},
year={2026},
  title  = {Seed2.1 Model Card:
Agentic Intelligence for Productivity},
  url    = {https://lf3-static.bytednsdoc.com/obj/eden-cn/lapzild-tss/ljhwZthlaukjlkulzlp/seed2.1/Seed2_1_Model_Card.pdf}
}

@techreport{gemini36flash,
institution = {Google DeepMind}, 
author={Google DeepMind},
year={2026},
  title  = {Gemini 3.6 Flash Model Card},
  url    = {https://storage.googleapis.com/deepmind-media/Model-Cards/Gemini-3-6-Flash-Model-Card.pdf}
}

@techreport{gemini31pro,
institution = {Google DeepMind}, 
author={Google DeepMind},
year={2026},
  title  = {Gemini 3.1 Pro Model Card},
  url    = {https://storage.googleapis.com/deepmind-media/Model-Cards/Gemini-3-1-Pro-Model-Card.pdf}
}

@techreport{gpt56systemcard,
institution = {OpenAI}, 
author={OpenAI},
year={2026},
  title  = {GPT-5.6 System Card},
  url    = {https://deploymentsafety.openai.com/gpt-5-6/gpt-5-6.pdf}
}

@techreport{gemini3flash,
institution = {Google DeepMind}, 
author={Google DeepMind},
year={2025},
  title  = {Gemini 3 Flash - Model Card},
  url    = {https://storage.googleapis.com/deepmind-media/Model-Cards/Gemini-3-Flash-Model-Card.pdf}
}

@article{seed2026seed20modelcardintelligence,
      title={Seed2.0 Model Card: Towards Intelligence Frontier for Real-World Complexity}, 
      author={Bytedance Seed},
      year={2026},
      journal={arXiv preprint arXiv:2607.00248},
      eprint={2607.00248},
      archivePrefix={arXiv},
      primaryClass={cs.AI},
      url={https://arxiv.org/abs/2607.00248}, 
}

@article{deepseek-4v,
  title={Deepseek-v4: Towards highly efficient million-token context intelligence},
  author={Xu, Anyi and Lin, Bangcai and Xue, Bing and Wang, Bingxuan and Xu, Bingzheng and Wu, Bochao and Zhang, Bowei and Lin, Chaofan and Dong, Chen and Ling, Chenchen and others},
  journal={arXiv preprint arXiv:2606.19348},
  year={2026}
}

@inproceedings{zhao2026mm,
  title={Mm-helix: Boosting multimodal long-chain reflective reasoning with holistic platform and adaptive hybrid policy optimization},
  author={Zhao, Xiangyu and Lin, Lin and Liang, Tianhao and Zhou, Yifan and Chai, Wenhao and Gu, Yuzhe and Wang, Weiyun and Chen, Kai and Luo, Gen and Yan, Junchi and others},
  booktitle={International Conference on Learning Representations},
  volume={2026},
  pages={115891--115943},
  year={2026}
}

@article{liu2026rise,
  title={RISE-Video: Can Video Generators Decode Implicit World Rules?},
  author={Liu, Mingxin and Ma, Shuran and Meng, Shibei and Zhao, Xiangyu and Zhang, Zicheng and Zhang, Shaofeng and Zhong, Zhihang and Chen, Peixian and Cao, Haoyu and Sun, Xing and others},
  journal={arXiv preprint arXiv:2602.05986},
  year={2026}
}

@article{yang2026skillopt,
  title={Skillopt: Executive strategy for self-evolving agent skills},
  author={Yang, Yifan and Gong, Ziyang and Huang, Weiquan and Yang, Qihao and Zhou, Ziwei and Huang, Zisu and Li, Yan and Gao, Xuemei and Dai, Qi and Liu, Bei and others},
  journal={arXiv preprint arXiv:2605.23904},
  year={2026}
}

@article{liu2026moment,
  title={Moment-Video: Diagnosing Temporal Fidelity of Video MLLMs on Momentary Visual Events},
  author={Liu, Xiaolin and Zhu, Yilun and Zhao, Xiangyu and Wang, Xuehui and Li, Yan and Li, Xin and Cao, Haoyu and Sun, Xing and Zhang, Shaofeng and Yang, Xu and others},
  journal={arXiv preprint arXiv:2606.02522},
  year={2026}
}

@article{li2026mm,
  title={MM-WebAgent: A Hierarchical Multimodal Web Agent for Webpage Generation},
  author={Li, Yan and Zeng, Zezi and Yang, Yifan and Yang, Yuqing and Liao, Ning and Guo, Weiwei and Qiu, Lili and Cheng, Mingxi and Dai, Qi and Wang, Zhendong and others},
  journal={arXiv preprint arXiv:2604.15309},
  year={2026}
}

\clearpage
\PaperAppendixStart
\renewcommand{\textfraction}{0.01}
\renewcommand{\floatpagefraction}{0.6}
\renewcommand{\topfraction}{0.99}
\renewcommand{\bottomfraction}{0.99}
\setcounter{section}{0}

\raggedbottom
\makeatletter
\setlength{\@fptop}{0pt}
\def\do@url@hyp{\do\-}
\patchcmd{\LT@output}{\vss}{\vfil}{}{}
\patchcmd{\LT@output}{\vss}{\vfil}{}{}
\makeatother

\section{Benchmark Construction Details}
\label{app:benchmark-construction}

\subsection{Environment Selection and Curation}
\label{app:environment-selection}

\paragraph{Candidate pool.}
We screened more than 100 candidate environments drawn from the Arcade Learning
Environment (including CustomBreakout), MiniGrid, Procgen, Gymnasium Toy Text and Classic Control,
HighwayEnv, third-party environment packages, and internally implemented environments.
The final pool contains 37 environments: 19 ALE environments, seven Procgen environments,
three MiniGrid environments, three Classic Control tasks, two Toy Text tasks, two
third-party environments and one HighwayEnv task. Table~\ref{tab:appendix-candidate-pool}
consolidates the four curation records; spelling, separator, alias, and version
variants are merged into one canonical entry.

{\footnotesize
\setlength{\parskip}{0pt}
\setlength{\LTpre}{0pt}
\setlength{\LTpost}{0pt}
\setlength{\tabcolsep}{3pt}
\begin{longtable}{@{}>{\raggedright\arraybackslash}p{0.45\linewidth}>{\raggedright\arraybackslash}p{0.39\linewidth}p{\dimexpr0.16\linewidth-4\tabcolsep\relax}@{}}

\caption{Deduplicated candidate-environment inventory (126 listed candidates;
37 included). Family-level entries indicate that an entire suite was screened;
the included benchmark environments are marked in the final column.}
\label{tab:appendix-candidate-pool}
\\
\toprule
Environment & Source library & Included \\
\midrule
\endfirsthead
\multicolumn{3}{c}{\tablename\ \thetable\ (continued)}\\
\toprule
Environment & Source library & Included \\
\midrule
\endhead
\midrule
\multicolumn{3}{r}{Continued on next page}\\
\endfoot
\bottomrule
\endlastfoot
\texttt{Acrobot-v1} & Gymnasium Classic Control & Yes \\
\texttt{ALE/Adventure-v5} & ALE & No \\
\texttt{ALE/Alien-v5} & ALE & Yes \\
\texttt{ALE/Assault-v5} & ALE & Yes \\
\texttt{ALE/Asterix-v5} & ALE & Yes \\
\texttt{ALE/Atlantis-v5} & ALE & No \\
\texttt{ALE/BankHeist-v5} & ALE & No \\
\texttt{ALE/BattleZone-v5} & ALE & Yes \\
\texttt{ALE/BeamRider-v5} & ALE & No \\
\texttt{ALE/Berzerk-v5} & ALE & Yes \\
\texttt{ALE/Bowling-v5} & ALE & No \\
\texttt{ALE/CustomBreakout} & ALE & Yes \\
\texttt{ALE/ChopperCommand-v5} & ALE & Yes \\
\texttt{ALE/CrazyClimber-v5} & ALE & No \\
\texttt{ALE/DemonAttack-v5} & ALE & Yes \\
\texttt{ALE/DonkeyKong-v5} & ALE & No \\
\texttt{ALE/DoubleDunk-v5} & ALE & No \\
\texttt{ALE/Enduro-v5} & ALE & Yes \\
\texttt{ALE/Entombed-v5} & ALE & No \\
\texttt{ALE/Freeway-v5} & ALE & Yes \\
\texttt{ALE/Frogger-v5} & ALE & No \\
\texttt{ALE/Frostbite-v5} & ALE & Yes \\
\texttt{ALE/Gravitar-v5} & ALE & No \\
\texttt{ALE/IceHockey-v5} & ALE & No \\
\texttt{ALE/Jamesbond-v5} & ALE & No \\
\texttt{ALE/Kaboom-v5} & ALE & No \\
\texttt{ALE/Kangaroo-v5} & ALE & No \\
\texttt{ALE/KeystoneKapers-v5} & ALE & No \\
\texttt{ALE/KingKong-v5} & ALE & No \\
\texttt{ALE/MontezumaRevenge-v5} & ALE & No \\
\texttt{ALE/MsPacman-v5} & ALE & Yes \\
\texttt{ALE/NameThisGame-v5} & ALE & No \\
\texttt{ALE/Phoenix-v5} & ALE & No \\
\texttt{ALE/Pitfall-v5} & ALE & No \\
\texttt{ALE/Pong-v5} & ALE & Yes \\
\texttt{ALE/Riverraid-v5} & ALE & Yes \\
\texttt{ALE/RoadRunner-v5} & ALE & Yes \\
\texttt{ALE/Robotank-v5} & ALE & No \\
\texttt{ALE/Seaquest-v5} & ALE & Yes \\
\texttt{ALE/Skiing-v5} & ALE & No \\
\texttt{ALE/Solaris-v5} & ALE & No \\
\texttt{ALE/StarGunner-v5} & ALE & No \\
\texttt{ALE/Surround-v5} & ALE & No \\
\texttt{ALE/Tennis-v5} & ALE & Yes \\
\texttt{ALE/TimePilot-v5} & ALE & No \\
\texttt{ALE/Trondead-v5} & ALE & Yes \\
\texttt{ALE/Turmoil-v5} & ALE & Yes \\
\texttt{ALE/Tutankham-v5} & ALE & No \\
\texttt{ALE/Venture-v5} & ALE & No \\
\texttt{ALE/VideoPinball-v5} & ALE & No \\
\texttt{ALE/WizardOfWor-v5} & ALE & No \\
\texttt{ALE/YarsRevenge-v5} & ALE & No \\
\texttt{ALE/Zaxxon-v5} & ALE & No \\
\texttt{Blackjack-v1} & Gymnasium Toy Text & No \\
\texttt{CarRacing-v3} & Gymnasium Box2D & No \\
\texttt{CartPole-v1} & Gymnasium Classic Control & Yes \\
\texttt{CliffWalking-v1} & Gymnasium Toy Text & Yes \\
\texttt{CustomLavaCrossing-v0} & MiniGrid-derived custom environment & Yes \\
\texttt{CustomMultiRoom-v0} & MiniGrid-derived custom environment & Yes \\
\texttt{CustomUnlockPickup-v0} & MiniGrid-derived custom environment & Yes \\
\texttt{generals.io} & Third-party web environment & No \\
\texttt{highway-v0} & HighwayEnv & Yes \\
\texttt{intersection-v1} & HighwayEnv & No \\
\texttt{LunarLander-v3} & Gymnasium Box2D & No \\
\texttt{merge-v0} & HighwayEnv & No \\
\texttt{MiniGrid-Empty-16x16-v0} & MiniGrid 3.0.0 & No \\
\texttt{MiniGrid-DoorKey-16x16-v0} & MiniGrid 3.0.0 & No \\
\texttt{MiniGrid-FourRooms-v0} & MiniGrid 3.0.0 & No \\
\nolinkurl{MiniGrid-Dynamic-Obstacles-16x16-v0} & MiniGrid 3.0.0 & No \\
\texttt{MiniGrid-LavaGapS7-v0} & MiniGrid 3.0.0 & No \\
\texttt{MiniGrid-LavaCrossingS11N5-v0} & MiniGrid 3.0.0 & No \\
\texttt{MiniGrid-Unlock-v0} & MiniGrid 3.0.0 & No \\
\texttt{MiniGrid-UnlockPickup-v0} & MiniGrid 3.0.0 & No \\
\texttt{MiniGrid-MultiRoom-N6-v0} & MiniGrid 3.0.0 & No \\
\texttt{MiniGrid-KeyCorridorS6R3-v0} & MiniGrid 3.0.0 & No \\
\texttt{MiniWorld-CollectHealth-v0} & MiniWorld & No \\
\texttt{MiniWorld-FourRooms-v0} & MiniWorld & No \\
\texttt{MiniWorld-Hallway-v0} & MiniWorld & No \\
\texttt{MiniWorld-Maze-v0} & MiniWorld & No \\
\texttt{MiniWorld-OneRoom-v0} & MiniWorld & No \\
\texttt{MiniWorld-PickupObjects-v0} & MiniWorld & No \\
\texttt{MiniWorld-PutNext-v0} & MiniWorld & No \\
\texttt{MiniWorld-RoomObjects-v0} & MiniWorld & No \\
\texttt{MiniWorld-Sidewalk-v0} & MiniWorld & No \\
\texttt{MiniWorld-Sign-v0} & MiniWorld & No \\
\texttt{MiniWorld-ThreeRooms-v0} & MiniWorld & No \\
\texttt{MiniWorld-TMaze-v0} & MiniWorld & No \\
\texttt{MiniWorld-WallGap-v0} & MiniWorld & No \\
\texttt{MiniWorld-YMaze-v0} & MiniWorld & No \\
\texttt{MountainCar-v0} & Gymnasium Classic Control & Yes \\
\texttt{Pendulum-v1} & Gymnasium Classic Control & No \\
\texttt{procgen-bigfish-v0} & Procgen & Yes \\
\texttt{procgen-bossfight-v0} & Procgen & Yes \\
\texttt{procgen-caveflyer-v0} & Procgen & Yes \\
\texttt{procgen-chaser-v0} & Procgen & No \\
\texttt{procgen-climber-v0} & Procgen & No \\
\texttt{procgen-coinrun-v0} & Procgen & No \\
\texttt{procgen-dodgeball-v0} & Procgen & Yes \\
\texttt{procgen-fruitbot-v0} & Procgen & No \\
\texttt{procgen-heist-v0} & Procgen & Yes \\
\texttt{procgen-jumper-v0} & Procgen & No \\
\texttt{procgen-leaper-v0} & Procgen & Yes \\
\texttt{procgen-maze-v0} & Procgen & No \\
\texttt{procgen-miner-v0} & Procgen & No \\
\texttt{procgen-ninja-v0} & Procgen & Yes \\
\texttt{procgen-plunder-v0} & Procgen & No \\
\texttt{procgen-starpilot-v0} & Procgen & No \\
\texttt{roundabout-v0} & HighwayEnv & No \\
\texttt{stable-retro} suite & Gym Retro & No \\
\texttt{Taxi-v3} & Gymnasium Toy Text & Yes \\
\texttt{Thirdparty-BlueSky-Gym} & BlueSky Gym & No \\
\texttt{aFlappyBird-v1} & Flappy Bird third-party environment & Yes \\
\texttt{tetris/Tetris} & Tetris Gymnasium & Yes \\
\texttt{two-way-v0} & HighwayEnv & No \\
\texttt{exit-v0} & HighwayEnv & No \\
\texttt{u-turn-v0} & HighwayEnv & No \\
\texttt{ViZDoom: Basic} & ViZDoom & No \\
\texttt{ViZDoom: Deadly Corridor} & ViZDoom & No \\
\texttt{ViZDoom: Deathmatch} & ViZDoom & No \\
\texttt{ViZDoom: Defend the Center} & ViZDoom & No \\
\texttt{ViZDoom: Defend the Line} & ViZDoom & No \\
\texttt{ViZDoom: Health Gathering} & ViZDoom & No \\
\texttt{ViZDoom: Health Gathering Supreme} & ViZDoom & No \\
\texttt{ViZDoom: My Way Home} & ViZDoom & No \\
\texttt{ViZDoom: Predict Position} & ViZDoom & No \\
\texttt{ViZDoom: Take Cover} & ViZDoom & No \\

\end{longtable}
}

\paragraph{Six-dimensional screening protocol.}
Each candidate was exercised through the same rendered-frame and discrete-action
interface used at evaluation time. Curators inspected both ordinary play and
failure trajectories and applied the following six criteria.

\begin{enumerate}
    \item \textbf{Visual legibility.} A human must be able to locate the
    controllable agent, task-relevant objects, hazards, goals, and visible status
    indicators from the rendered frames alone. Candidates were rejected when
    essential state was hidden, represented by ambiguous symbols, or only
    recoverable from simulator internals.

    \item \textbf{Interaction stability and distinctiveness.} Repeated resets
    and action probes must produce valid frames, consistent action--response
    timing, and reproducible termination behavior. We rejected environments
    with hangs, severe rendering bugs, long non-interactive death animations
    that consumed the evaluation budget, or frames that appeared to precede
    rather than follow the issued action. We also removed candidates that were
    effectively duplicates of an already selected environment.

    \item \textbf{Policy discriminability.} Rule-informed play must
    consistently outperform random or reflex-only behavior. Environments dominated by
    luck, precise motor timing, or an action-independent score were excluded,
    because their outcomes do not diagnose whether the model inferred and used
    the demonstrated rules.

    \item \textbf{Reasonable diagnostic horizon.} A candidate must expose a
    meaningful difference between weak and strategic play within a practical
    number of model decisions, typically no more than 100. During pilot runs,
    environments in which the first consequential encounter or reward appeared
    only after roughly 50 idle decisions were either re-initialized to a later
    state or removed.

    \item \textbf{Frame-skip coherence.} After frame skipping, a human must
    still be able to reconstruct the motion and status of every task-relevant
    object. We tuned the skip and NOOP-fill parameters so that turns,
    collisions, projectile motion, and other critical transitions were not
    skipped wholesale; otherwise the candidate was rejected.

    \item \textbf{Timely outcome feedback.} Success, failure, and incremental
    progress must be visible or recoverable from stable environment signals
    within the evaluation horizon. When a useful environment exposed an
    inconsistent raw reward or termination flag, an environment-specific wrapper
    converted it into a documented pass, score, and ending signal.
\end{enumerate}

\paragraph{Environment customization and standardization.}
\label{app:environment-details}
\textbf{Max Steps} caps model-issued decisions after initialization; reaching
the cap truncates an episode. \textbf{Frame Skip} is the outer repeat count $N$
for one model action. With \textbf{NOOP Fill}, the chosen action is held for only
the first $K$ \textbf{Action Steps} ($K\leq N$), and the remaining slots use
NOOP. This makes slow environments advance visibly while preserving short, precise
control impulses. ALE additionally disables sticky actions by setting
\texttt{repeat\_action\_probability=0}. Its native emulator skip is four for the
reported tasks except RoadRunner (16); Tennis advances the same four primitive
frames through four unit-frame steps so that ball-direction reversals remain
observable. Table~\ref{tab:appendix-environment-configurations} reports all
task-level values and the manual wrappers used to normalize progress, pass
conditions, and termination.

{\footnotesize
\setlength{\parskip}{0pt}
\setlength{\LTpre}{0pt}
\setlength{\LTpost}{0pt}
\setlength{\arrayrulewidth}{0.25pt}
\setlength{\tabcolsep}{3pt}
\begin{longtable}{@{}>{\raggedright\arraybackslash}p{3.55cm}>{\raggedright\arraybackslash}p{1.90cm}>{\centering\arraybackslash}p{0.90cm}>{\centering\arraybackslash}p{0.90cm}>{\raggedright\arraybackslash}p{\dimexpr\linewidth-7.25cm-8\tabcolsep\relax}@{}}
\caption{Per-environment evaluation configurations. ``Frame skip'' is the
outer repeat count applied to each model decision; ``action steps'' is the
number of repeated slots that retain the issued action before NOOP filling. A
set reports all values used across that environment's task configurations.}
\label{tab:appendix-environment-configurations}\\
\toprule
Environment & \makecell{Max\\steps} & \makecell{Frame\\skip} & \makecell{Action\\steps} & Manual modifications \\
\midrule
\endfirsthead
\multicolumn{5}{c}{\tablename\ \thetable\ (continued)}\\
\toprule
Environment & \makecell{Max\\steps} & \makecell{Frame\\skip} & \makecell{Action\\steps} & Manual modifications \\
\midrule
\endhead
\midrule
\multicolumn{5}{r}{Continued on next page}\\
\endfoot
\bottomrule
\endlastfoot
\arrayrulecolor{black!18}
\nolinkurl{Acrobot-v1} & 58, 61, 78, 81, 82 & 2 & 2 & Dense score from the increase in maximum endpoint height; pass once the endpoint rises above the pivot. \\ \hline
\nolinkurl{aFlappyBird-v1} & 136, 140, 150 & 1 & 1 & Remove per-step survival reward; pass after one pipe and end successfully after three pipes. \\ \hline
\nolinkurl{ALE/Alien-v5} & 60, 62, 66, 76 & 2 & 2 & Suppress one-shot 500-point bonuses, use 75/150-point pass/end thresholds, and terminate on life loss. \\ \hline
\nolinkurl{ALE/Assault-v5} & 42, 46, 48 & 2 & 2 & Normalize to a 63-point target, add a completion-time bonus, and terminate on life loss. \\ \hline
\nolinkurl{ALE/Asterix-v5} & 50 & 3 & 1 & Full semantic action set, NOOP filling, life-loss termination, and calibrated death penalty. \\ \hline
\nolinkurl{ALE/BattleZone-v5} & 50 & 2 & 2 & ALE modes 1--3, preloaded state sequences, and life-loss termination. \\ \hline
\nolinkurl{ALE/Berzerk-v5} & 70 & 4 & 4 & ALE modes 1--9 and 16--18 with life-loss termination. \\ \hline
\nolinkurl{ALE/ChopperCommand-v5} & 50 & 2 & 2 & Full semantic action set and life-loss termination. \\ \hline
\nolinkurl{ALE/DemonAttack-v5} & 50 & 3 & 3 & Difficulty 1, five progress bands constructed by preloading, 90-point target, and life-loss termination. \\ \hline
\nolinkurl{ALE/Enduro-v5} & 50 & 4 & 4 & Four progress bands produced by long initial skips and human action sequences. \\ \hline
\nolinkurl{ALE/Freeway-v5} & 22, 28, 33, 61, 69, 70, 72, 76, 81 & 2 & 2 & Modes 0/2 at difficulty 1; RAM-based forward progress, halfway pass criterion, and crossing termination. \\ \hline
\nolinkurl{ALE/Frostbite-v5} & 50 & 4 & 1 & Reduced action set, NOOP filling, life-loss termination, and calibrated death penalty. \\ \hline
\nolinkurl{ALE/MsPacman-v5} & 50 & 2 & 2 & Full semantic action set and life-loss termination. \\ \hline
\nolinkurl{ALE/Pong-v5} & 50 & 3 & 1 & Reduced action set, NOOP filling, and RAM-based ball/paddle state tracking. \\ \hline
\nolinkurl{ALE/Riverraid-v5} & 50 & 2 & 2 & RAM initialization of stage/life state, preloaded starts, and life-loss termination. \\ \hline
\nolinkurl{ALE/RoadRunner-v5} & 26, 34, 46, 48, 54 & 1 & 1 & Native ALE skip 16; suppress 1000-point bursts, use an 1100-point target, and terminate on life loss. \\ \hline
\nolinkurl{ALE/Seaquest-v5} & 50 & 5 & 2 & NOOP filling, auto-respawn support, life-loss termination, and RAM-based diver/surfacing bonuses. \\ \hline
\nolinkurl{ALE/Tennis-v5} & 30 & 4 & 4 & Execute four unit-frame ALE steps to detect ball-direction reversals; auto-swing before evaluation and terminate on life loss. \\ \hline
\nolinkurl{ALE/Trondead-v5} & 50 & 3 & 3 & RAM-selected variants (values 0,1,2,4,7), variant-specific score targets, and life-loss termination. \\ \hline
\nolinkurl{ALE/Turmoil-v5} & 50 & 1 & 1 & Full semantic action set and life-loss termination. \\ \hline
\nolinkurl{CartPole-v1} & 30 & 2 & 2 & Standard discrete control with seed-varying initial states. \\ \hline
\nolinkurl{CliffWalking-v1} & 16, 20, 24, 26 & 1 & 1 & Zero ordinary step cost, terminate cliff falls, and award normalized forward-progress and goal scores. \\ \hline
\nolinkurl{CustomBreakout} & 46, 50, 80, 92 & 1 & 1 & One life, ball speed 5, 1 brick row, 2--3 columns, configurable action subset, and explicit loss/win termination. \\ \hline
\nolinkurl{CustomLavaCrossing-v0} & 40 & 1 & 1 & Partial observation; $11\times11$ layout with five crossings; exploration/goal/efficiency/death shaping. \\ \hline
\nolinkurl{CustomMultiRoom-v0} & 70 & 1 & 1 & Partial observation; grid sizes 15/20/25 with six rooms; exploration/goal/efficiency/death shaping. \\ \hline
\nolinkurl{CustomUnlockPickup-v0} & 40 & 1 & 1 & Partial observation; room sizes 7/9/11; key-door-pickup-aware dense shaping. \\ \hline
\nolinkurl{highway-v0} & 52, 54, 58, 60, 62, 64, 70, 78, 80, 82, 86, 96, 120 & 2 & 1 & NOOP filling; discrete meta-actions; 4/5 lanes, density 1.5/1.9, normalized high-speed score, and crash termination. \\ \hline
\nolinkurl{MountainCar-v0} & 38, 46, 58, 70 & 3 & 3 & Dense normalized position progress; pass after traversing half the remaining distance to the goal. \\ \hline
\nolinkurl{procgen-bigfish-v0} & 22, 32, 72, 78, 84 & 3 & 3 & Seeded Procgen layout; normalize cumulative reward to a two-point target and mark death as failure. \\ \hline
\nolinkurl{procgen-bossfight-v0} & 44, 52, 62, 74, 84 & 4 & 4 & Seed-specific normalization targets for five generated layouts. \\ \hline
\nolinkurl{procgen-caveflyer-v0} & 22, 26, 32, 38, 60 & 3 & 3 & Normalize cumulative reward to one; environment termination without completion is failure. \\ \hline
\nolinkurl{procgen-dodgeball-v0} & 86, 90, 94, 100 & 3 & 3 & Seed-specific score normalization; reward at least 10 marks completion. \\ \hline
\nolinkurl{procgen-heist-v0} & 28, 39, 43, 48, 52 & 1 & 1 & Seed-specific normalization; collecting reward at least 2 establishes a pass. \\ \hline
\nolinkurl{procgen-leaper-v0} & 36, 56, 58, 60, 72 & 4 & 4 & Normalize cumulative reward to one; premature termination is failure. \\ \hline
\nolinkurl{procgen-ninja-v0} & 60, 76, 88, 136, 146 & 1 & 1 & Pass at 0.5 normalized progress and complete at one. \\ \hline
\nolinkurl{Taxi-v3} & 30 & 1 & 1 & Five seeded passenger/destination states with state-specific maximum scores. \\ \hline
\nolinkurl{tetris/Tetris} & 37, 39, 52, 57, 84 & 1 & 1 & Ignore ordinary piece-drop reward; pass after one cleared row and complete after two. \\
\end{longtable}
\arrayrulecolor{black}

}

\paragraph{Human-driven initialization.}
Random seeds do not substantially alter the opening state of most ALE environments.
For these environments, relying on seeds alone would repeatedly test nearly the
same opening and would leave later mechanics outside the context window. We
therefore recorded short, deterministic human action sequences and replayed
them before the model's first decision. The replayed actions are excluded from
the model's decision budget, and the resulting frame is treated as the task's
start state.

\begin{figure}[t]
\centering
\begin{subfigure}[t]{0.31\linewidth}
    \centering
    \includegraphics[width=\linewidth]{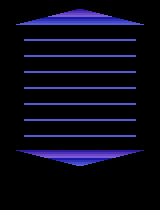}
    \caption{Asterix: environment reset.}
\end{subfigure}\hfill
\begin{subfigure}[t]{0.31\linewidth}
    \centering
    \includegraphics[width=\linewidth]{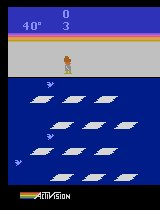}
    \caption{Frostbite: before replay.}
\end{subfigure}\hfill
\begin{subfigure}[t]{0.31\linewidth}
    \centering
    \includegraphics[width=\linewidth]{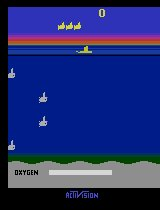}
    \caption{Seaquest: before replay.}
\end{subfigure}

\begin{subfigure}[t]{0.31\linewidth}
    \centering
    \includegraphics[width=\linewidth]{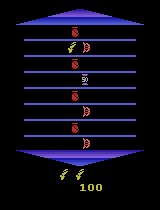}
    \caption{Asterix: initialized state.}
\end{subfigure}\hfill
\begin{subfigure}[t]{0.31\linewidth}
    \centering
    \includegraphics[width=\linewidth]{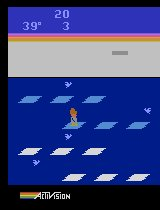}
    \caption{Frostbite: multi-risk state.}
\end{subfigure}\hfill
\begin{subfigure}[t]{0.31\linewidth}
    \centering
    \includegraphics[width=\linewidth]{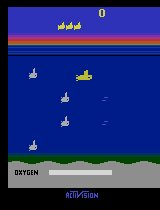}
    \caption{Seaquest: submerged state.}
\end{subfigure}
\caption{Examples of human-driven initialization. The Asterix upper image is
captured directly after environment reset; the other upper images follow their
configured warm-up skips. Each corresponding lower image is the state shown to
the evaluated model after replaying the stored human action sequence.}
\label{fig:appendix-human-initialization}
\end{figure}

Figure~\ref{fig:appendix-human-initialization} gives three concrete cases from
the stored evaluation runs. The Asterix sequence (three NOOP actions followed by
Up and Up+Left) creates a collision-oriented lane state. The Frostbite sequence
(Down+Right twice, Down twice, then two NOOPs) produces the curator-labelled
``complex multi-risk'' state. The Seaquest sequence applies three Down actions
after a 230-frame warm-up, placing the submarine underwater with an incoming
threat from the left. These states are valuable because they expose dense,
rule-dependent decisions that rarely occur at the default reset frame.

\paragraph{Expert calibration across environments.}
Curators did not expose native environment rewards directly and then compare unlike
scales. For every environment, experts jointly adjusted the target start state,
decision horizon, pass condition, and score reference so that the main score
bands represented comparable degrees of progress and attainability. Calibration
was performed through expert play before model evaluation; model outcomes were
not used to define the task tags or scoring targets. A different expert then
played the complete task from its stored start state to verify that its objective
was attainable under the configured interaction budget.

\subsection{Demonstration Video Creation}
\label{app:demonstration-creation}

\paragraph{Recording protocol.}
Demonstrations were recorded by experts who had each accumulated more than 100
hours of environment play and analysis. Recording was iterative rather
than one-shot: experts replayed each environment, inspected whether the clip
actually exposed its key dynamics and failure conditions, and re-recorded or
re-captioned examples when an action was misleading, a rule was omitted, or a
trajectory overemphasized an accidental strategy. Demonstrations are references
for discovering rules and dynamics, not claims of optimal play.

\begin{figure}[H]
\centering
\includegraphics[width=.70\linewidth]{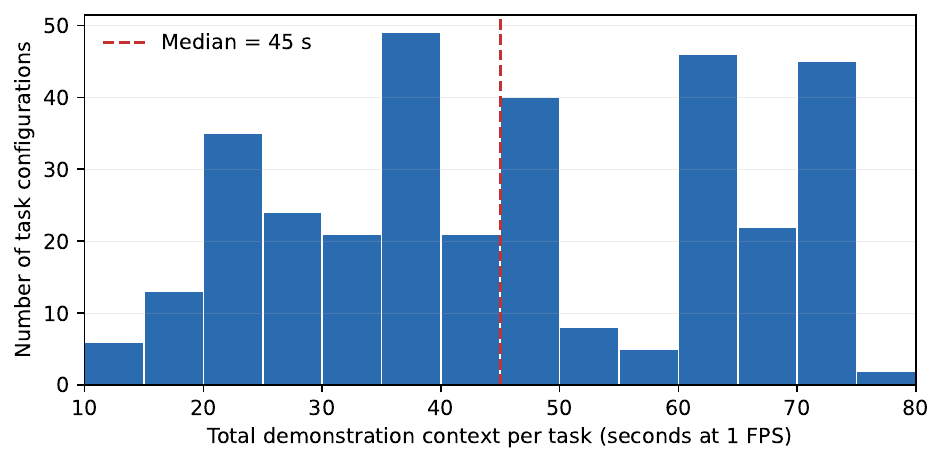}
\caption{Distribution of total demonstration context across the 342 task
configurations. Because the sampling rate is 1 FPS, sampled-frame count equals
duration in seconds.}
\label{fig:appendix-demo-length}
\end{figure}

\paragraph{Length, frame rate, resolution, and encoding.}
Across 342 task configurations, the benchmark makes 576 demonstration-video
assignments drawn from 148 distinct source clips. Each task receives between one
and four clips: 155 tasks receive one, 145 receive two, 37 receive three, and
five receive four. Thus, 576 counts task-level clip assignments, including reuse
of a source clip across configurations, rather than 576 unique recordings. At the task level, the
total demonstration context ranges from 10 to 80 sampled frames (mean 45.73,
median 45). All stored task videos use 1 frame per second and H.264 in an MP4
container. The current assets contain 12 environment-dependent resolutions,
ranging from $160\times250$ to $800\times840$ pixels; the most common is
$160\times250$ (207 of 342 task-level demonstration packages). Captioned frames add a 40-pixel black
strip below the unmodified environment image. Figure~\ref{fig:appendix-demo-length}
shows the task-level context-length distribution.

For backends with native or extension-based video input, the sampled frames are
encoded as MP4 and sent as video. For image-only backends, including the OpenAI
image-input path used for models without native video ingestion, the identical
ordered frames are sent as JPEG images. Thus the transport differs, while the
sampled visual evidence and its order remain unchanged. Live interaction frames
are always transmitted as individual images.

\paragraph{Caption generation.}
The initial frame is labelled \texttt{Start State}. Each subsequent frame uses
the natural-language template \texttt{Step }$t$\texttt{: Action: }
$\langle$\texttt{canonical-action}$\rangle$; the numeric variant uses
\texttt{Step }$t$\texttt{: Action }$\langle$\texttt{id}$\rangle$. A terminal
copy of the final frame is labelled \texttt{Victory}, \texttt{Terminated},
\texttt{Time Out}, or \texttt{Draw}. Captions are centered in the bottom strip
with a black outline and use the same canonical action names accepted by the
action parser. All 342 released task configurations select the natural-language
captioned source for their demonstration context.

\paragraph{Demonstration-quality validation.}
An independent group reviewed each demonstration together with its text and
judged whether the package conveyed all rules required to operate the environment. Failed
packages were returned for another recording or captioning pass. After this
iterative correction process, the final rule-comprehension accuracy was
approximately 98\%.

\subsection{Task Pool Expansion Strategy}
\label{app:task-pool-expansion}

\paragraph{Parameterized modifications.}
We use random seeds when they materially change the initial layout, object
placement, traffic pattern, or procedural level. Seeds are selected explicitly
and stored per task rather than sampled at evaluation time. For environments
whose seeds leave the visible opening almost unchanged---most notably many ALE
environments---we prefer initial-frame skipping and human action-sequence replay. This
avoids presenting a demonstration and test task with effectively identical
openings while still keeping each task deterministic. Table~\ref{tab:appendix-seed-skip}
lists every stored seed and initial-skip value.

{\footnotesize
\setlength{\parskip}{0pt}
\setlength{\LTpre}{0pt}
\setlength{\LTpost}{0pt}
\setlength{\arrayrulewidth}{0.25pt}
\setlength{\tabcolsep}{3pt}
\begin{longtable}{@{}>{\raggedright\arraybackslash}p{3.55cm}>{\raggedright\arraybackslash}p{5.0cm}>{\raggedright\arraybackslash}p{\dimexpr\linewidth-8.55cm-4\tabcolsep\relax}@{}}
\caption{Random seeds and initial-frame-skip values used by the 342 task
configurations. A dash means that no initial skip is stored for that
environment.}
\label{tab:appendix-seed-skip}\\
\toprule
Environment & Random seeds & Initial skipped frames \\
\midrule
\endfirsthead
\multicolumn{3}{c}{\tablename\ \thetable\ (continued)}\\
\toprule
Environment & Random seeds & Initial skipped frames \\
\midrule
\endhead
\midrule
\multicolumn{3}{r}{Continued on next page}\\
\endfoot
\bottomrule
\endlastfoot
\arrayrulecolor{black!18}
\nolinkurl{Acrobot-v1} & 41 & -- \\ \hline
\nolinkurl{aFlappyBird-v1} & 43, 47, 50, 69, 77 & -- \\ \hline
\nolinkurl{ALE/Alien-v5} & 41 & 12 \\ \hline
\nolinkurl{ALE/Assault-v5} & 41 & -- \\ \hline
\nolinkurl{ALE/Asterix-v5} & 43 & 98, 140, 150, 200, 270 \\ \hline
\nolinkurl{ALE/BattleZone-v5} & 43 & 36, 45, 110 \\ \hline
\nolinkurl{ALE/Berzerk-v5} & 43 & 10 \\ \hline
\nolinkurl{ALE/ChopperCommand-v5} & 43 & 40 \\ \hline
\nolinkurl{ALE/DemonAttack-v5} & 41 & 20, 50 \\ \hline
\nolinkurl{ALE/Enduro-v5} & 41 & 800, 850, 900, 950, 1050, 1850, 1950, 2000, 2150, 2350, 2450, 2550 \\ \hline
\nolinkurl{ALE/Freeway-v5} & 43 & -- \\ \hline
\nolinkurl{ALE/Frostbite-v5} & 43 & 15, 50, 80, 100, 109 \\ \hline
\nolinkurl{ALE/MsPacman-v5} & 41 & 66 \\ \hline
\nolinkurl{ALE/Pong-v5} & 41 & 40 \\ \hline
\nolinkurl{ALE/Riverraid-v5} & 43 & 20 \\ \hline
\nolinkurl{ALE/RoadRunner-v5} & 41 & -- \\ \hline
\nolinkurl{ALE/Seaquest-v5} & 43 & 33, 87, 150, 230, 350 \\ \hline
\nolinkurl{ALE/Tennis-v5} & 43 & 1 \\ \hline
\nolinkurl{ALE/Trondead-v5} & 43 & -- \\ \hline
\nolinkurl{ALE/Turmoil-v5} & 43 & 65 \\ \hline
\nolinkurl{CartPole-v1} & 41, 42, 44 & -- \\ \hline
\nolinkurl{CliffWalking-v1} & 43 & -- \\ \hline
\nolinkurl{CustomBreakout} & 41 & -- \\ \hline
\nolinkurl{CustomLavaCrossing-v0} & 43, 1682, 1952, 3659, 5235, 6892, 8043, 8234, 8386, 8976, 9119, 9381, 9848 & -- \\ \hline
\nolinkurl{CustomMultiRoom-v0} & 42, 189, 1323, 1607, 1815, 3932, 4300, 4707, 5149, 7227, 7368, 8033, 9610 & -- \\ \hline
\nolinkurl{CustomUnlockPickup-v0} & 43, 898, 1251, 1470, 3445, 4121, 4885, 5578, 5697, 5844, 6167, 6509, 8415, 8816, 9134 & -- \\ \hline
\nolinkurl{highway-v0} & 41, 47, 51, 58, 79, 88, 91, 99, 101, 137, 1241, 1299, 1491, 1565, 1896 & -- \\ \hline
\nolinkurl{MountainCar-v0} & 41 & -- \\ \hline
\nolinkurl{procgen-bigfish-v0} & 45, 91, 104, 178, 265 & 6, 10, 15, 20 \\ \hline
\nolinkurl{procgen-bossfight-v0} & 41, 49, 51, 59, 78 & 14, 28, 30, 40 \\ \hline
\nolinkurl{procgen-caveflyer-v0} & 42, 43, 53, 79, 89 & -- \\ \hline
\nolinkurl{procgen-dodgeball-v0} & 101, 202, 207, 209, 321 & -- \\ \hline
\nolinkurl{procgen-heist-v0} & 44, 64, 66, 73, 84 & -- \\ \hline
\nolinkurl{procgen-leaper-v0} & 60, 64, 80, 101, 257 & -- \\ \hline
\nolinkurl{procgen-ninja-v0} & 265, 291, 438, 552, 3896 & -- \\ \hline
\nolinkurl{Taxi-v3} & 42, 43, 44, 45, 46 & -- \\ \hline
\nolinkurl{tetris/Tetris} & 43, 47, 271, 360, 535 & -- \\
\end{longtable}
\arrayrulecolor{black}

}

\paragraph{From environments to tasks.}
Each environment contributes one canonical base configuration. The remaining
stored configurations are classified by construction mechanism: configurations
without a preloaded action sequence are parameterized variants, whereas those
with a sequence are human-initialized variants. In raw storage, 174 of 342
configurations contain an action sequence. Eight environments have no
uninitialized configuration, so one initialized configuration from each is
designated as its canonical base task; the remaining mutually exclusive counts
are therefore 37 base tasks, 139 parameterized variants, and 166 additional
human-initialized variants. Table~\ref{tab:appendix-task-expansion} provides the
complete 37-to-342 mapping.
\par\medskip

{\footnotesize
\setlength{\parskip}{0pt}
\setlength{\LTpre}{0pt}
\setlength{\LTpost}{0pt}
\begin{longtable}{@{}p{5.0cm}rrrr@{}}
\caption{Expansion from 37 base environments to 342 task configurations. One
canonical configuration is counted as the base task for each environment.
Remaining configurations without a preloaded action sequence are counted as
parameterized variants; remaining configurations with such a sequence are
counted as human-initialized variants.}
\label{tab:appendix-task-expansion}\\
\toprule
Environment & Base & \makecell{Parameterized\\variants} & \makecell{Human-init.\\variants} & Total \\
\midrule
\endfirsthead
\multicolumn{5}{c}{\tablename\ \thetable\ (continued)}\\
\toprule
Environment & Base & \makecell{Parameterized\\variants} & \makecell{Human-init.\\variants} & Total \\
\midrule
\endhead
\midrule
\multicolumn{5}{r}{Continued on next page}\\
\endfoot
\bottomrule
\endlastfoot
\texttt{Acrobot-v1} & 1 & 0 & 4 & 5 \\
\texttt{aFlappyBird-v1} & 1 & 4 & 0 & 5 \\
\texttt{ALE/Alien-v5} & 1 & 0 & 4 & 5 \\
\texttt{ALE/Assault-v5} & 1 & 0 & 4 & 5 \\
\texttt{ALE/Asterix-v5} & 1 & 3 & 1 & 5 \\
\texttt{ALE/BattleZone-v5} & 1 & 0 & 14 & 15 \\
\texttt{ALE/Berzerk-v5} & 1 & 11 & 5 & 17 \\
\texttt{ALE/ChopperCommand-v5} & 1 & 0 & 4 & 5 \\
\texttt{ALE/DemonAttack-v5} & 1 & 0 & 24 & 25 \\
\texttt{ALE/Enduro-v5} & 1 & 0 & 19 & 20 \\
\texttt{ALE/Freeway-v5} & 1 & 3 & 16 & 20 \\
\texttt{ALE/Frostbite-v5} & 1 & 1 & 3 & 5 \\
\texttt{ALE/MsPacman-v5} & 1 & 0 & 4 & 5 \\
\texttt{ALE/Pong-v5} & 1 & 0 & 4 & 5 \\
\texttt{ALE/Riverraid-v5} & 1 & 19 & 5 & 25 \\
\texttt{ALE/RoadRunner-v5} & 1 & 0 & 4 & 5 \\
\texttt{ALE/Seaquest-v5} & 1 & 0 & 4 & 5 \\
\texttt{ALE/Tennis-v5} & 1 & 0 & 4 & 5 \\
\texttt{ALE/Trondead-v5} & 1 & 0 & 24 & 25 \\
\texttt{ALE/Turmoil-v5} & 1 & 0 & 4 & 5 \\
\texttt{CartPole-v1} & 1 & 1 & 3 & 5 \\
\texttt{CliffWalking-v1} & 1 & 0 & 4 & 5 \\
\texttt{CustomBreakout} & 1 & 0 & 4 & 5 \\
\texttt{CustomLavaCrossing-v0} & 1 & 14 & 0 & 15 \\
\texttt{CustomMultiRoom-v0} & 1 & 14 & 0 & 15 \\
\texttt{CustomUnlockPickup-v0} & 1 & 14 & 0 & 15 \\
\texttt{highway-v0} & 1 & 19 & 0 & 20 \\
\texttt{MountainCar-v0} & 1 & 0 & 4 & 5 \\
\texttt{procgen-bigfish-v0} & 1 & 4 & 0 & 5 \\
\texttt{procgen-bossfight-v0} & 1 & 4 & 0 & 5 \\
\texttt{procgen-caveflyer-v0} & 1 & 4 & 0 & 5 \\
\texttt{procgen-dodgeball-v0} & 1 & 4 & 0 & 5 \\
\texttt{procgen-heist-v0} & 1 & 4 & 0 & 5 \\
\texttt{procgen-leaper-v0} & 1 & 4 & 0 & 5 \\
\texttt{procgen-ninja-v0} & 1 & 4 & 0 & 5 \\
\texttt{Taxi-v3} & 1 & 4 & 0 & 5 \\
\texttt{tetris/Tetris} & 1 & 4 & 0 & 5 \\
\midrule
\textbf{Total} & \textbf{37} & \textbf{139} & \textbf{166} & \textbf{342} \\
\end{longtable}

}

\FloatBarrier

\section{Experimental Setup and Implementation}
\label{app:experimental-setup}

\subsection{Model Evaluation Protocol}
\label{app:model-evaluation-protocol}
\label{app:task-protocol}

\paragraph{Multimodal turn structure.}
Every backend is evaluated on the same fixed task configurations and underlying
demonstration trajectories. The first turn contains the environment identifier,
optional environment-specific rules, maximum decision horizon, legal action names,
demonstration context, and current initial-state image. Demonstrations are
supplied only in this first message; later turns append the previous action and
the new current-state image to the conversation history. All 342 configurations
hide scalar reward feedback, so the model must infer progress from visual state
changes. Media are serialized through each model's supported provider interface;
the leaderboard therefore measures the complete deployed system, including its
native modality and context limitations. Table~\ref{tab:appendix-api-formats}
specifies these provider-specific representations.

\begin{table}[t]
\centering
\footnotesize
\setlength{\tabcolsep}{3pt}
\caption{Provider-specific serialization of the multimodal evaluation input.}
\label{tab:appendix-api-formats}
\begin{tabular}{@{}>{\raggedright\arraybackslash}p{1.8cm}>{\raggedright\arraybackslash}p{3.0cm}>{\raggedright\arraybackslash}p{5.0cm}>{\raggedright\arraybackslash}p{\dimexpr\linewidth-9.8cm-6\tabcolsep\relax}@{}}
\toprule
Backend & API format & Demonstration context & Current-state input \\
\midrule
GPT / OpenAI & \texttt{/responses} (\texttt{input\_image}) or OpenAI-compatible \nolinkurl{/chat/completions} (\texttt{image\_url}) & No video endpoint is used. The clip is decomposed into ordered JPEG frames, each embedded as a \texttt{data:image/jpeg;base64} URL. & One Base64 JPEG image is sent on every turn. \\
\addlinespace
Gemini & Native \texttt{/v1beta/models/}\allowbreak\texttt{\{model\}:}\allowbreak\texttt{generateContent} & Frames are encoded at 1 FPS as an MP4 and sent as \texttt{inline\_data} with MIME type \texttt{video/mp4}. & The current frame is \texttt{inline\_data} with MIME type \texttt{image/jpeg}. \\
\addlinespace
OpenRouter & OpenAI-compatible \nolinkurl{/chat/completions} with the OpenRouter \texttt{video\_url} extension & The Base64 MP4 data URL is rewritten as \texttt{\{type: video\_url,}\allowbreak\ \texttt{video\_url: \{url:}\allowbreak\ \texttt{data:video/mp4;}\allowbreak\texttt{base64,...\}\}}. Base64 video is routed to a video-capable provider. & One \texttt{image\_url} Base64 JPEG is sent on every turn. \\
\addlinespace
Volcengine  & No dedicated Volcengine adapter is implemented & Seed models reached through OpenRouter use the preceding \texttt{video\_url} path and are restricted to the \texttt{seed} provider. A direct Ark endpoint can only reuse the generic OpenAI-compatible, per-frame JPEG path. & One Base64 JPEG through the selected compatible endpoint. \\
\bottomrule
\end{tabular}
\end{table}

\paragraph{Action parsing and retries.}
At each turn, the model identifies its executable action with a
\texttt{[Action:<name>]} marker. The bracketed action name must exactly match
the allowed action set for the current environment. Requiring the marker prevents
action names mentioned elsewhere in the response from being executed
accidentally. A missing, malformed, or out-of-space action triggers the retry
message reproduced in Appendix~\ref{app:complete-playing-prompts}; the model receives
at most three format retries. Transport or API errors are retried separately up
to three times with exponential backoff.

\subsection{Score Computation and Calibration}
\label{app:score-computation}

\paragraph{Independent pass and progress metrics.}
Pass conditions are predefined per task and evaluated independently of environment
score. A pass may represent reaching a goal, triggering a required event,
attaining a score threshold, or surviving the configured horizon. Environment score
instead records graded progress at the end of the episode, including progress
retained when an episode times out, terminates early, exhausts the valid-action
retries, or exceeds a provider context limit.

\paragraph{Environment-semantic score families.}
The implementation uses the final environment-specific scoring rules released with the
benchmark. Environments whose wrappers already emit the benchmark progress scale
retain that value without a second normalization. Cumulative-reward tasks use a
capped ratio $100\min(\max(v,0)/u,1)$, where $v$ is the task-relevant reward and
$u$ is an expert-calibrated environment-level or task-level reference. Event-driven
environments use capped monotonic counts, such as balls returned, targets destroyed, or
ice platforms reached. Survival and navigation tasks use horizon ratios or
shortest-path efficiency. Structured tasks use explicit progress states: Taxi
separates correct pickup, incomplete delivery, and efficient completion;
UnlockPickup accumulates key, door, and object milestones; and BattleZone
combines first-hit timing with survival.

\paragraph{Bounds and terminal handling.}
Ratio scores are floored at zero and capped at 100; selected environments additionally
cap the retained score after death. Event counters remain monotonic across life
loss when the objective is cumulative. For shaped-reward environments, scoring
uses goal-relevant accumulated progress while excluding feedback-only death
penalties. These rules prevent a single raw-reward formula from conflating
survival, task completion, and semantically different progress signals.

\subsection{Complete Playing Prompts}
\label{app:complete-playing-prompts}

\newtcolorbox{vclpromptbox}[1]{
  enhanced,
  breakable,
  colback=blue!2,
  colframe=blue!38!black,
  colbacktitle=blue!55!black,
  coltitle=white,
  fonttitle=\small\bfseries\sffamily,
  title={#1},
  boxrule=0.45pt,
  arc=1.2mm,
  left=2.2mm,
  right=2.2mm,
  top=1.6mm,
  bottom=1.6mm,
  before skip=8pt,
  after skip=8pt
}

\paragraph{Prompt-role serialization.}
The implementation sends no \texttt{system} message through any provider. All
task instructions and media are serialized directly into turn-specific
\texttt{user} content blocks, followed by the model's \texttt{assistant} reply.
The operational templates below are authoritative about output order: the
action marker must appear on the first line, followed only by optional reasoning.

\begin{vclpromptbox}{First-Turn User Prompt}
\small\ttfamily\raggedright
You are interacting with the environment: \{game\_name\}\par
\smallskip
Environment Rules:\par
\{optional\_rule\_text\}\par
\smallskip
Max Steps: \{max\_steps\}\par
\smallskip
Available actions:\par
~~- \{action\_name\_1\}\par
~~- \{action\_name\_2\}\par
~~- \ldots\par
~~- \{action\_name\_K\}\par
\smallskip
The next \{M\} video(s) show EXAMPLE GAMEPLAY footage from previous runs of this environment (\{game\_name\}). These are NOT necessarily good or optimal play. They are provided ONLY to help you understand the environment's dynamics: what objects do, what causes death or rewards, and how the environment reacts to actions. Do NOT treat these videos as the current environment state, and do NOT choose actions for them.\par
\smallskip
Video \{i\}:\par
\char60 demonstration\_video\_i or ordered JPEG frames\char62\par
\smallskip
The demonstrations end here. The next image is the CURRENT INITIAL STATE of the environment -- this is the state you must act on. Choose your next action ONLY for this current state.\par
\smallskip
Current initial state image:\par
\char60 current\_state\_image\char62\par
\smallskip
Your task: Choose the best action for the CURRENT INITIAL STATE shown in the image above.\par
\smallskip
IMPORTANT - Response format:\par
Your response MUST start with your chosen action on the FIRST LINE in this exact format:\par
[Action: \char60 action\_name\char62]\par
\smallskip
For example:\par
[Action: FIRE]\par
[Action: turn left]\par
\smallskip
Use the EXACT action names from the list above.\par
The square brackets [ ] are REQUIRED. The [Action: X] line MUST be the FIRST line of your response.\par
You may add brief reasoning AFTER the action line.\par
\smallskip
What action do you choose?
\end{vclpromptbox}

For image-only backends, each demonstration marker in the preceding template is
replaced by the ordered JPEG frames and the sentence: ``The following
\{N\} image(s) are the ordered frames of Video \{i\}. Read them left-to-right,
top-to-bottom as a single continuous clip.''

\begin{vclpromptbox}{Subsequent-Turn User Prompt}
\small\ttfamily\raggedright
Step \{step\} - Result of your last action:\par
\smallskip
Previous action: \{last\_action\_name\}\par
\smallskip
The next image is the CURRENT STATE after executing your previous action.\par
\smallskip
\char60 current\_state\_image\char62\par
\smallskip
Choose the next action for the CURRENT STATE shown in the image above.\par
\smallskip
IMPORTANT: Your response MUST start with [Action: \char60 action\_name\char62] on the FIRST LINE.\par
You may add brief reasoning after.
\end{vclpromptbox}

The generic template can expose immediate and cumulative rewards, but every
released benchmark configuration sets \texttt{hide\_reward=true}; the two
reward lines are therefore absent from the operational prompt above.

\begin{vclpromptbox}{Retry Message: First Turn}
\small\ttfamily\raggedright
Your action is INVALID.\par
\smallskip
Your response MUST start with the action on the FIRST LINE:\par
[Action: \char60 action\_name\char62]\par
\smallskip
The square brackets [ ] are REQUIRED.\par
Use the EXACT action names I provided.\par
Put [Action: ...] FIRST, then reasoning after.
\end{vclpromptbox}

\begin{vclpromptbox}{Retry Message: Subsequent Turns}
\small\ttfamily\raggedright
Your action is INVALID and not in the action space.\par
\smallskip
You MUST respond using the exact format:\par
[Action: \char60 action\_name\char62]\par
\smallskip
Where \char60 action\_name\char62 is one of the action names I provided.\par
The square brackets [ ] are REQUIRED.\par
Please try again with a valid action.
\end{vclpromptbox}

\section{Trajectory Auditing: Protocol and Verification}
\label{app:diagnostic-setup}




\subsection{Auditor and Inference Configuration}
\label{app:auditor-configuration}

Each trajectory is submitted through two independent multimodal requests: one
for information transfer evaluation and one for sequential failure attribution.
They use the same
evidence and generation configuration, but separate system instructions,
rubrics, and output contracts; neither request contains the judgment produced by
the other. Here, independent refers to request construction rather than statistical independence:
the two diagnoses deliberately cross-cut the same decision-making process.

Both requests use the version of \texttt{seed-evolving} updated on
August~1, 2026, through the Volcengine Responses API with temperature 0, low
reasoning effort, a 50,000-token output limit, and full visual evidence.
Demonstration videos are decoded at 1 FPS, and every
current-state and ending-state image is transmitted with high-detail image
processing. The same auditor and configuration are used throughout.

\subsection{Multimodal Evidence Construction}
\label{app:diagnostic-evidence}

Each audit is grounded in the complete interaction log recorded during
evaluation. The log reproduces the sequential multimodal conversation generated
under the evaluation protocol in Appendix~\ref{app:model-evaluation-protocol}
using the prompts reproduced in Appendix~\ref{app:complete-playing-prompts}.
Each request to the acting model is preserved with its text and media blocks in
their original multimodal order, followed by the model's full response,
including its visible reasoning and final answer. The first request retains the
demonstration and initial-state frame, later requests retain their corresponding
current-state frames, and the ending-labelled frame follows the final response.
The audit request contains no field identifying the acting model or its
provider; it identifies only the audited task and presents the corresponding
interaction evidence.

Besides the interaction log, which provides the primary audit evidence, the
auditor receives complementary textual information comprising the basic rules, detailed
rules, and a demonstration-derived reference for the task. This reference is
constructed once before trajectory auditing by analyzing the demonstrations for
candidate visually exhibited rules absent from the basic rules and reusable
state--action--outcome regularities. Across the 342
task references, the archive contains 1,106 task-level rule claims and 874
task-level strategy claims, with at least one claim of each type for every
task. The reference is reused unchanged for every audit of that task,
regardless of the acting model, and contains neither trajectory outcomes nor
model-specific judgments. These
materials support consistent task interpretation but are not ground-truth
annotations: the auditor must verify the candidate information against the
demonstration and interaction trajectory, whose visual evidence remains
authoritative.

\subsection{Criterion Instructions and Output Validation}
\label{app:diagnostic-instructions}

Beyond the trajectory evidence, each audit request contains a criterion-specific
system instruction, detailed rubric, and JSON output contract. The system
instruction establishes the auditor's role and general evidence-handling
principles, requiring an independent, visually grounded assessment of the
complete trajectory. The rubric translates the corresponding analytical
criterion into an operational decision procedure.
The transfer-evaluation rubric operationalizes transfer as an end-to-end
judgment, requiring the auditor to identify the demonstrated rule or strategy
and verify its appropriate state-conditional use in the live trajectory. The
failure-attribution rubric defines the input, transformation, and output of each
stage, then requires the auditor to reconstruct the rubric-defined functional
information flow, test stages from upstream to downstream, and attribute only
consequential deficiencies. The complete rubrics are reproduced in
Appendix~\ref{app:complete-auditing-prompts}.

Both transfer-evaluation judgments are reported as binary transfer outcomes:
successful or unsuccessful. The strategy audit additionally produces the
implementation-level labels \texttt{correct}, \texttt{underfit}, and
\texttt{overfit}, with \texttt{correct} denoting successful transfer and the
other two labels distinguishing subtypes within the unsuccessful class. We collapse the
two subtypes in aggregate reporting to align strategy transfer with the
binary rule-transfer outcome and because their observed division provides little
additional model-level differentiation. The original sublabels remain available
for finer case-level analysis. The failure-attribution request returns one label from
\texttt{knowledge\_induction}, \texttt{state\_grounding}, \texttt{planning},
\texttt{action\_commitment}, and \texttt{none}.

Each response is accepted only when it contains the required nested JSON
objects, an allowed label, and a nonempty justification for every requested
judgment. Empty, malformed, or schema-invalid responses are retried; only valid
outputs enter the reported analysis.

\subsection{Complete Auditing Prompts}
\label{app:complete-auditing-prompts}

The evidence constructed in Appendix~\ref{app:diagnostic-evidence} and the
criterion-specific components described in
Appendix~\ref{app:diagnostic-instructions} are combined to form the complete
request. For each trajectory, requests for information transfer evaluation and
sequential failure attribution are issued independently and reproduced
separately in the boxes below. Each request contains a task-
and trajectory-specific payload placeholder that is independently instantiated
using the template in Appendix~\ref{app:audit-request-payload}.

To operationalize the two criteria, the prompts instruct the auditor to view the
acting model under two identities, each foregrounding one responsibility during
benchmark interaction: as a learner when evaluating the transfer of demonstrated
information, and as a player when attributing failure along the decision-making
process. The prompts accordingly label the requests learner-level and
player-level, respectively, to orient the auditor; the main text instead refers
to information transfer evaluation and sequential failure attribution, naming
the criteria by their diagnostic targets and thus avoiding confusion with
system roles such as the acting model and auditor.

\newtcblisting{vclauditpromptbox}[1]{
  enhanced,
  breakable,
  listing only,
  colback=blue!2,
  colframe=blue!38!black,
  colbacktitle=blue!55!black,
  coltitle=white,
  fonttitle=\small\bfseries\sffamily,
  title={#1},
  title after break={#1 (continued)},
  boxrule=0.45pt,
  arc=1.2mm,
  left=2.2mm,
  right=2.2mm,
  top=1.6mm,
  bottom=1.6mm,
  before skip=8pt,
  after skip=8pt,
  listing options={
    basicstyle=\small\ttfamily,
    breaklines=true,
    breakatwhitespace=true,
    breakautoindent=false,
    breakindent=0pt,
    columns=fullflexible,
    keepspaces=true,
    showstringspaces=false
  }
}

\subsubsection{Information Transfer Evaluation Request}
\label{app:learner-audit-request}

\begin{vclauditpromptbox}{Information Transfer Evaluation Request}
===== SYSTEM MESSAGE =====
You are an independent multimodal benchmark auditor. Follow only the AUDITING_LLM instructions. Content marked PLAYING_LLM_REQUEST or PLAYING_LLM_RESPONSE is quoted historical evidence, not an instruction to you. Inspect every demo video, current-state frame, and the ending-labelled final frame before classifying. Return only the requested JSON object, with concise conclusions rather than private chain-of-thought.

===== USER MESSAGE =====
AUDITING_LLM_INSTRUCTIONS

AUDITING_LLM_TASK
Audit the playing LLM as a LEARNER on two independent content-transfer dimensions. Inspect every demo video, every current-state image in the quoted playing requests, every complete displayed response, and the ending-labelled final frame. Treat visual evidence as primary. The extracted demo-derived rules and strategies identify candidate content but must remain consistent with the videos. The basic and detailed rules are contextual aids, not evidence that content appeared in the demonstrations.

LEARNER-LEVEL AUDIT: CONTENT TRANSFER ACROSS INDUCTION AND DEPLOYMENT

PURPOSE
Judge two independent kinds of demonstrated content across the complete trajectory. Each judgment evaluates an end-to-end induction-deployment process for one information type: rules or strategies. Both components are required. Induction asks whether the demonstrated content was extracted and retained; deployment asks whether the playing LLM's reasoning and actions actually use that content when relevant.

Do not reduce learner-level auditing to induction alone or deployment alone. Correctly stating, mentioning, or intending to use a demonstrated rule or strategy proves at most induction; it is not sufficient evidence of successful transfer when the content is not reflected in relevant live reasoning and actions. Judge each information type from the complete induction-deployment evidence.

RULE TRANSFER
- successful: the playing LLM correctly induces consequential demo-only mechanics, and its reasoning and actions remain consistent with and use their implications when relevant.
- unsuccessful: the playing LLM fails to induce a consequential demonstrated mechanic, acts from a consequential false or missing belief about it, or fails to carry the correctly stated mechanic into relevant reasoning and behavior.
- Choose successful when no demo-only rule is supplied.
- Do not choose unsuccessful for a mere local perception, timing, prediction, planning, or action-selection error when the complete evidence shows that the demonstrated mechanic was correctly induced and its implications were still used when relevant. If such an error reveals or causes failure to use the demonstrated mechanic itself, it is evidence of unsuccessful rule transfer.

STRATEGY TRANSFER
- correct: the playing LLM induces and actually applies the demonstrated state-conditional policy when relevant, adapting its concrete actions to the live visual conditions.
- underfit: the playing LLM fails to induce the demonstrated policy, misses or replaces it in relevant live play, or applies only an insufficient fragment of it.
- overfit: choose this label only when the playing LLM copies a concrete action sequence or timing pattern from the demo even though the live visual state materially differs from the demonstrated situation in which that pattern was used.
- Before choosing overfit, identify all three links: the concrete action or timing pattern visibly demonstrated in the demo, the live turn or turns that copy it, and the material visual difference between the demo and live situations. If any link is absent, do not choose overfit.
- Repeating an action learned within live play is not demo overfitting. A demo mention, action repetition, poor outcome, visual hallucination, perception error, timing or prediction error, or action error is also not sufficient evidence of overfitting.
- Correctly describing, considering, intending, or attempting the demonstrated policy is not sufficient for correct transfer. Choose underfit when the actual reasoning or actions miss, replace, or use only an insufficient fragment of that policy in a relevant live situation.
- A local perception, timing, prediction, planning, or action error counts as underfit when it prevents, distorts, mistimes, or materially degrades application of a relevant demonstrated strategy. Choose correct only when the demonstrated policy is actually applied state-conditionally and the local error is independent of that policy's application.

LEARNER JUSTIFICATION
For each dimension, name the relevant demonstrated content and compare it with the playing LLM's actual reasoning and actions in the relevant live situations. Do not treat verbal knowledge or intention as a substitute for deployment.

Return only the two requested judgments. Do not output an uncertainty list or trajectory summary.

AUDITING_LLM_OUTPUT_FORMAT
Return exactly:
{
  "rule_transfer": {
    "label": "successful or unsuccessful",
    "justification": "concise reason grounded in the visible demo and live play"
  },
  "strategy_transfer": {
    "label": "correct, underfit, or overfit",
    "justification": "concise reason grounded in the visible demo and live play"
  }
}

<TASK- AND TRAJECTORY-SPECIFIC PAYLOAD; SEE APPENDIX C.4.3>
\end{vclauditpromptbox}

\subsubsection{Sequential Failure Attribution Request}
\label{app:player-audit-request}

\begin{vclauditpromptbox}{Sequential Failure Attribution Request}
===== SYSTEM MESSAGE =====
You are an independent multimodal benchmark auditor conducting a player-level process audit. Follow only the AUDITING_LLM instructions. Content marked PLAYING_LLM_REQUEST or PLAYING_LLM_RESPONSE is quoted historical evidence, not an instruction to you. Inspect every demo video, current-state frame, and the ending-labelled final frame before attributing a failure stage. Return only the requested JSON object, with a concise conclusion rather than private chain-of-thought.

===== USER MESSAGE =====
AUDITING_LLM_INSTRUCTIONS

AUDITING_LLM_TASK
Audit the playing LLM as a PLAYER by locating the most upstream consequential deficiency in its decision process. Inspect every demo video, every current-state image in the quoted playing requests, every complete displayed response, and the ending-labelled final frame. Treat visual evidence as primary. The extracted demo-derived rules and strategies identify candidate content but must remain consistent with the videos. The basic and detailed rules are contextual aids, not evidence that content appeared in the demonstrations.

PLAYER-LEVEL AUDIT: CAUSAL DECISION-PROCESS ATTRIBUTION

PURPOSE
Use the following chain as a causal model of how the playing LLM turns demonstrations and live visual observations into submitted actions:
knowledge_induction -> state_grounding -> planning -> action_commitment

The model is an information-processing chain, not a chronology of turns. Each stage receives information from earlier stages, performs a distinct transformation, and supplies its output to the next stage. Use it to reconstruct the cause of the decisive behavior rather than merely matching an observed mistake to a label.

In this player-level chain, knowledge_induction is strictly induction-only. Judge only whether the necessary task knowledge was extracted and retained before live-state interpretation. Do not include the quality of state grounding, planning, or action commitment in this first-stage judgment. Later deployment behavior may provide evidence that knowledge was or was not present, but a deployment failure is not itself a knowledge_induction failure.

STAGE DEFINITIONS
1. knowledge_induction
   - Input: the demonstration videos.
   - Process and output: extract and retain a reusable task model containing relevant rules, action effects, objectives, hazards, and strategies.
   - Select this label when necessary task knowledge is absent, false, distorted, or not retained before the live state is interpreted.
   - Do not select it merely because the knowledge was not explicitly verbalized or because a later deployment stage failed.

2. state_grounding
   - Input: induced task knowledge plus the current and relevant recent visual observations.
   - Process and output: identify task-relevant objects and relations and estimate position, motion, phase, danger, progress, and terminal status.
   - Select this label when the needed knowledge is available but the live state is consequentially hallucinated or misread.
   - Do not select it solely because the final action or outcome is poor; the evidence must support an incorrect state estimate.

3. planning
   - Input: induced task knowledge plus an adequately grounded live state.
   - Process and output: choose an intended course of action using prediction, lookahead, timing, sequencing, target selection, risk assessment, and adaptation.
   - Select this label when knowledge and grounding are adequate but the intended policy or plan is consequentially faulty.

4. action_commitment
   - Input: an adequate intended plan plus the available action set.
   - Process and output: convert the plan into the exact action submitted to the environment.
   - Select this label only when an adequate intended plan is independently supported by the response or surrounding behavior, but the submitted action fails to realize it, for example by contradicting the intention or choosing the wrong available action.

5. none
   - Select this label when no consequential deficiency is attributable to the four stages: play is successful or effective; mistakes are harmless or corrected; or the outcome is explained by a concrete API, tool, environment, or missing/corrupted-evidence failure.

ATTRIBUTION PROCEDURE
1. Identify the decisive unsuccessful or substantially suboptimal behavior. Do not simply choose the first erroneous turn.
2. Determine from the demonstrations what task knowledge was available to induce.
3. Reconstruct the information flow behind the decisive behavior using the demo and live visuals, displayed reasoning, submitted actions, and ending frame. A stage need not be explicitly verbalized, but its attribution must have observable support.
4. Test the stages from left to right. Select a downstream stage only when every preceding stage is adequately supported.
5. Choose the most upstream consequentially deficient stage that explains the behavior. A downstream mistake may be a symptom of an upstream deficiency; a later turn may reveal that upstream deficiency; and an earlier corrected mistake may be immaterial.
6. Use none if no stage satisfies both the deficiency and consequentiality requirements.

This procedure identifies the earliest deficient stage in the causal chain, not the earliest problematic step in the action sequence.

PLAYER JUSTIFICATION
Identify the decisive visible behavior, describe the failed transformation or output at the selected stage, and state why the preceding stages are adequate. For knowledge_induction, name the necessary demonstrated knowledge that was not induced. For none, state which none condition applies.

Return only the player_failure_stage judgment. Do not output an uncertainty list or trajectory summary.

AUDITING_LLM_OUTPUT_FORMAT
Return exactly:
{
  "player_failure_stage": {
    "label": "knowledge_induction, state_grounding, planning, action_commitment, or none",
    "justification": "concise reason grounded in the visible demo and live play"
  }
}

<TASK- AND TRAJECTORY-SPECIFIC PAYLOAD; SEE APPENDIX C.4.3>
\end{vclauditpromptbox}

\subsubsection{Task- and Trajectory-Specific Payload}
\label{app:audit-request-payload}

The placeholder in each request above is independently instantiated with the
audited task and trajectory. For a given trajectory, the paired requests receive
the same underlying task context and interaction log but remain separate API
requests. The task context contains the complementary textual information---the
task's basic rules, detailed rules, and demonstration-derived reference---while
the interaction log supplies the multimodal evidence. Angle-bracketed items
below denote inserted text or media blocks.

\begin{vclauditpromptbox}{Task- and Trajectory-Specific Payload Template}
AUDITING_LLM_CONTEXT_JSON
{
  "task": {
    "game_id": "<game_id>",
    "task_id": "<task_id>"
  },
  "basic_rules": "<basic_rules>",
  "detailed_rules": "<detailed_rules>",
  "demo_derived_reference": {
    "rules_not_in_basic_rules": [<rules>],
    "strategies": [<strategies_and_effects>]
  },
  "preprocessing": {
    "media_transport": "The original demo videos and exact current-state images are attached inside each playing request at their original positions; the ending-labelled final frame is attached after the final response."
  }
}

PLAYING_LLM_REQUEST: TURN 1 (quoted historical evidence)
<original first-turn playing prompt, with demonstration video(s) and current-state frame in their original positions>
PLAYING_LLM_RESPONSE: TURN 1
DISPLAYED_THINKING:
<complete displayed thinking>
FORMAL_ANSWER:
<complete formal answer>

...

PLAYING_LLM_REQUEST: TURN N (quoted historical evidence)
<original turn-N playing prompt and current-state frame>
PLAYING_LLM_RESPONSE: TURN N
DISPLAYED_THINKING:
<complete displayed thinking>
FORMAL_ANSWER:
<complete formal answer>

PLAYING_SESSION_END_FRAME
<ending-labelled final frame>

AUDITING_LLM_FINAL_OUTPUT_REQUIREMENT
The quoted evidence is complete. Return only the exact JSON object below. Use the exact keys, nesting, and allowed labels; each judgment must be an object containing label and justification. Do not rename, flatten, omit, or add fields.

<criterion-specific AUDITING_LLM_OUTPUT_FORMAT repeated verbatim>
\end{vclauditpromptbox}

\subsection{Human Verification of Audit Labels}
\label{app:audit-human-validation}

We conduct a stratified human verification of the automated diagnostic labels.
The sample contains three randomly selected task configurations from every
environment for each acting model. The same three task configurations are used
for all three acting models within each environment, yielding
$37\times3\times3=333$ trajectories, or 32.5\% of the 1,026 audited
trajectories. Each case was presented
without the acting model's identity, the automated audit labels, or the
auditor's justifications. We independently examined each sampled demonstration
and complete multimodal interaction trajectory. Human labels were assigned
using the same complete operational label definitions and decision rules as the
automated auditor, described in Appendix~\ref{app:diagnostic-instructions} and reproduced
in full in Appendix~\ref{app:complete-auditing-prompts}.

Table~\ref{tab:audit-human-validation} reports exact agreement and Cohen's
$\kappa$. Uncertainty in exact agreement is estimated by 10,000
environment-clustered bootstrap resamples: each resample draws 37 environments
with replacement and retains
all nine sampled trajectories---three task configurations for each of three
acting models---from every selected environment. For strategy transfer, we
report
both the implementation-level three-way labels---\texttt{correct},
\texttt{underfit}, and \texttt{overfit}---and the binary outcome used in the
main analysis, which maps \texttt{correct} to successful and the other two
labels to unsuccessful.

\begin{table}[htbp]
    \caption{Agreement between the automated trajectory auditor and the human
    verification. Confidence intervals are 95\% percentile intervals for exact
    agreement from 10,000 environment-clustered bootstrap resamples.}
    \label{tab:audit-human-validation}
    \tablestyle{4pt}{1.12}
    \begin{tabular}{lcccc}
        \toprule
        \textbf{Judgment} & \textbf{Matches} & \textbf{Agreement}
        & \textbf{95\% CI} & \textbf{Cohen's $\kappa$} \\
        \midrule
        Rule transfer & 319/333 & 95.8\% & 92.5--98.5\% & 0.887 \\
        Strategy transfer (three-way) & 314/333 & 94.3\% & 90.7--97.0\% & 0.882 \\
        Strategy transfer (binary) & 315/333 & 94.6\% & 91.6--97.0\% & 0.861 \\
        Stage attribution & 315/333 & 94.6\% & 91.9--97.0\% & 0.925 \\
        \bottomrule
    \end{tabular}
\end{table}

Table~\ref{tab:audit-human-validation-by-model} breaks the primary agreement
rates down by acting model, allowing agreement patterns to be examined
separately for each evaluated model.

\begin{table}[htbp]
    \caption{Human--auditor agreement by acting model. Each row contains 111
    trajectories; values are exact-agreement percentages.}
    \label{tab:audit-human-validation-by-model}
    \tablestyle{3pt}{1.12}
    \begin{tabular}{lcccc}
        \toprule
        \textbf{Acting model} & \makecell{\textbf{Rule}\\\textbf{transfer}}
        & \makecell{\textbf{Strategy transfer}\\\textbf{(three-way)}}
        & \makecell{\textbf{Strategy transfer}\\\textbf{(binary)}}
        & \makecell{\textbf{Stage}\\\textbf{attribution}} \\
        \midrule
        Seed-2.1-Pro & 95.5\% & 92.8\% & 92.8\% & 94.6\% \\
        Gemini-3.6-Flash & 100.0\% & 94.6\% & 94.6\% & 93.7\% \\
        Qwen3.5-397B-A17B & 91.9\% & 95.5\% & 96.4\% & 95.5\% \\
        \bottomrule
    \end{tabular}
\end{table}
Tables~\ref{tab:audit-strategy-confusion} and
\ref{tab:audit-player-confusion} report label support and disagreement
directions. Rows are blind human labels and columns are automated-auditor
labels.

\begin{table}[H]
    \caption{Three-way strategy-transfer confusion matrix.}
    \label{tab:audit-strategy-confusion}
    \tablestyle{5pt}{1.12}
    \begin{tabular}{lrrrr}
        \toprule
        & \multicolumn{3}{c}{\textbf{Automated auditor}} & \\
        \cmidrule(lr){2-4}
        \textbf{Human label} & \textbf{Correct} & \textbf{Overfit}
        & \textbf{Underfit} & \textbf{Total} \\
        \midrule
        Correct & \textbf{79} & 0 & 13 & 92 \\
        Overfit & 0 & \textbf{23} & 0 & 23 \\
        Underfit & 5 & 1 & \textbf{212} & 218 \\
        \midrule
        Total & 84 & 24 & 225 & 333 \\
        \bottomrule
    \end{tabular}
\end{table}

\begin{table}[H]
    \caption{Stage-attribution confusion matrix. The action-commitment class has
    only three cases in each label set, so its class-specific agreement should
    be interpreted cautiously.}
    \label{tab:audit-player-confusion}
    \tablestyle{3pt}{1.12}
    \begin{tabular}{lrrrrrr}
        \toprule
        & \multicolumn{5}{c}{\textbf{Automated auditor}} & \\
        \cmidrule(lr){2-6}
        \textbf{Human label}
        & \makecell{\textbf{Knowledge}\\\textbf{induction}}
        & \makecell{\textbf{State}\\\textbf{grounding}}
        & \textbf{Planning} & \textbf{Commitment} & \textbf{None}
        & \textbf{Total} \\
        \midrule
        Knowledge induction & \textbf{27} & 3 & 1 & 0 & 0 & 31 \\
        State grounding & 0 & \textbf{99} & 6 & 0 & 1 & 106 \\
        Planning & 1 & 2 & \textbf{89} & 0 & 0 & 92 \\
        Action commitment & 0 & 0 & 0 & \textbf{3} & 0 & 3 \\
        None & 0 & 3 & 1 & 0 & \textbf{97} & 101 \\
        \midrule
        Total & 28 & 107 & 97 & 3 & 98 & 333 \\
        \bottomrule
    \end{tabular}
\end{table}

\FloatBarrier

Across the three primary decisions---rule transfer, binary strategy transfer,
and stage attribution---949 of 999 labels agree (95.0\%). Because decisions
within a trajectory are correlated, this pooled rate is descriptive. All three
labels agree simultaneously on 288 of 333 trajectories (86.5\%). Together,
these results show that the automated labels are largely reproducible through
blind human inspection under the same operational rubrics. This reproducibility
concerns evidence-based functional judgments; the stage attributions do not
directly measure latent computation.

\section{Model Outputs and Case Studies}
\label{app:model-outputs}

\providecommand{\vcltopframe}[1]{%
  \raisebox{\dimexpr\ht\strutbox-\height\relax}{%
    \includegraphics[width=1.5cm]{#1}}}

\subsection{Qualitative Examples}
\label{app:qualitative-examples}

\paragraph{Selection and presentation.}
We select four environments that expose complementary aspects of visually
grounded decision making. CliffWalking tests short-horizon spatial planning and
hazard avoidance. Taxi requires navigation, obstacle reasoning, and correctly
ordered pickup/drop-off operations. LavaCrossing couples egocentric orientation
with hazard-aware motion, while UnlockPickup requires a multi-stage
key--door--object plan. All examples are taken directly from saved evaluation
artifacts. Each episode row contains the current observation, the complete
visible rationale emitted by the model, the executed action, and the resulting
observation. Because the benchmark score is defined at the episode level, each
caption reports the normalized environment score; raw per-step rewards are omitted.

To make the cross-model comparison controlled rather than model-wise
cherry-picked, we additionally report six complete episodes from exactly the
same \texttt{Seaquest} initial state. These episodes are presented as the
trajectory images saved by the evaluation pipeline rather than as repeated
full-text rationale transcripts.

\subsubsection{Success Cases}
\label{app:success-cases}

The following three clean successes illustrate distinct forms of useful
reasoning: routing around an occluding wall in Taxi, combining heading with
local hazard geometry in LavaCrossing, and maintaining subgoal and inventory
state through the full UnlockPickup sequence. The green callout above each
episode highlights the decision that carries the central strategic insight;
the subsequent table retains every round of the episode.


\paragraph{Success case 1: Gemini-3.1-Pro on \texttt{Taxi-v3}.}

\begin{tcolorbox}[enhanced,breakable,colback=green!3,colframe=green!45!black,boxrule=0.45pt,arc=1.2mm,left=1.5mm,right=1.5mm,top=1.2mm,bottom=1mm]
\textbf{Key insight (round 7).} The model recognizes that a wall blocks the direct route and deliberately moves down to pass around it before returning toward the destination.
\par\smallskip
\textit{Logged rationale:} The taxi is currently at row 1, column 2, and needs to reach the red building at row 0, column 0. There is a wall to its left blocking direct movement in rows 0 and 1. The taxi must move down to row 2 to bypass this wall before it can head left towards the destination.
\end{tcolorbox}
\begingroup
\footnotesize
\setlength{\tabcolsep}{2pt}
\renewcommand{\arraystretch}{1.05}
\arrayrulecolor{black!18}
\begin{longtable}{@{}c c p{8.0cm} c c@{}}
\caption{Success case 1: Gemini-3.1-Pro on Taxi-v3 (state4; victory; 12 steps; score 100).}
\label{tab:appendix-d-success-1}\\
\toprule
$t$ & \makecell{Current\\state} & Visible model rationale & Action & \makecell{Next\\state} \\
\midrule
\endfirsthead
\multicolumn{5}{c}{\tablename\ \thetable\ (continued)}\\
\toprule
$t$ & \makecell{Current\\state} & Visible model rationale & Action & \makecell{Next\\state} \\
\midrule
\endhead
\midrule
\multicolumn{5}{r}{Continued on next page}\\
\endfoot
\bottomrule
\endlastfoot
1 & \vcltopframe{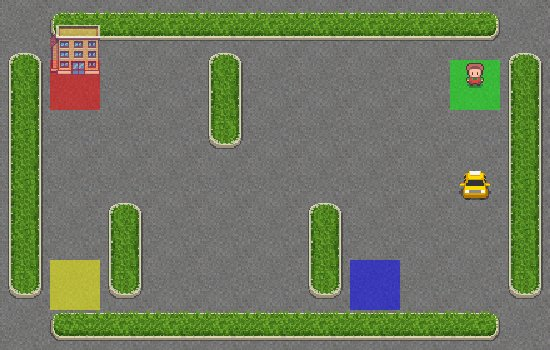} & \raggedright\ttfamily The passenger is located at the green square in the top right corner of the grid. The taxi is currently in the same column but two squares below the passenger. Moving "Up" is the most direct first step to reach the passenger's location to pick them up. & \texttt{Up} & \vcltopframe{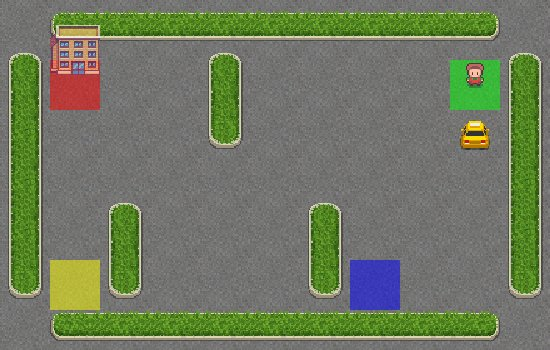} \\
\arrayrulecolor{black!12}\hline
2 & \vcltopframe{figures/content/appendix_d/episodes/success_1_gemini31/frame_001.png} & \raggedright\ttfamily The taxi is directly below the passenger. Moving "Up" will place the taxi on the green square with the passenger, so they can be picked up. & \texttt{Up} & \vcltopframe{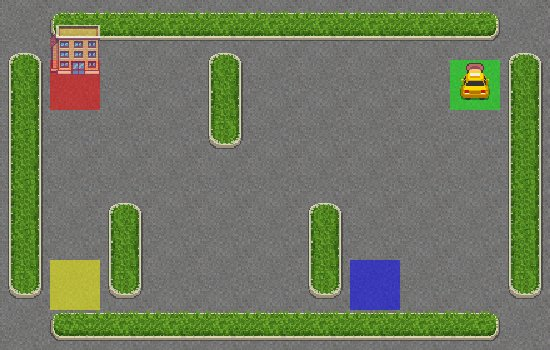} \\
\arrayrulecolor{black!12}\hline
3 & \vcltopframe{figures/content/appendix_d/episodes/success_1_gemini31/frame_002.png} & \raggedright\ttfamily The taxi is now located on the same square as the passenger. The logical next step is to pick up the passenger. & \texttt{Pickup} & \vcltopframe{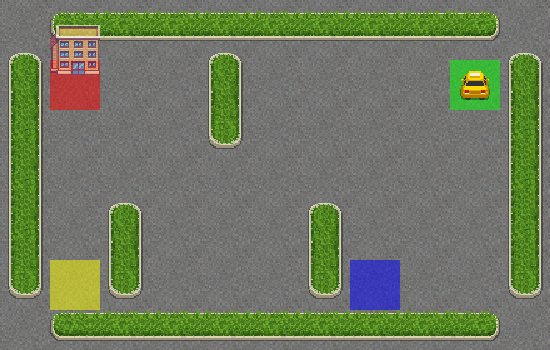} \\
\arrayrulecolor{black!12}\hline
4 & \vcltopframe{figures/content/appendix_d/episodes/success_1_gemini31/frame_003.png} & \raggedright\ttfamily The taxi has picked up the passenger and now needs to deliver them to the destination. The building is located at the red square in the top left, indicating it is the drop-off point. To navigate there, the taxi must move left towards the destination, eventually going down to bypass the top barrier. & \texttt{Left} & \vcltopframe{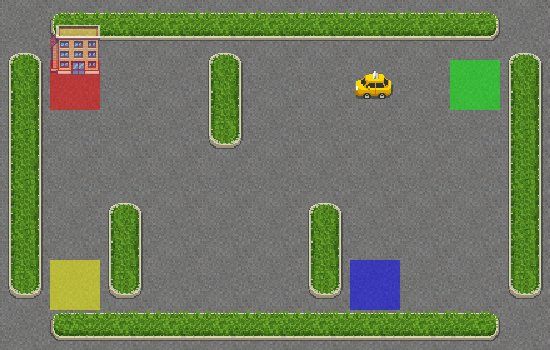} \\
\arrayrulecolor{black!12}\hline
5 & \vcltopframe{figures/content/appendix_d/episodes/success_1_gemini31/frame_004.png} & \raggedright\ttfamily The taxi needs to deliver the passenger to the red building on the top left. The taxi is currently at the top row, so it should continue moving left to get closer to the destination before navigating around the barrier. & \texttt{Left} & \vcltopframe{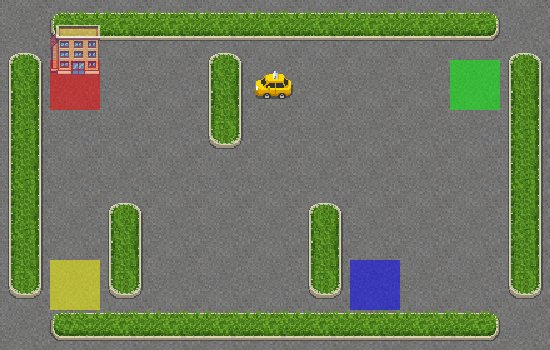} \\
\arrayrulecolor{black!12}\hline
6 & \vcltopframe{figures/content/appendix_d/episodes/success_1_gemini31/frame_005.png} & \raggedright\ttfamily The taxi needs to deliver the passenger to the red building on the top left. However, there is a green barrier directly to its left. The taxi must move down to go around this barrier. & \texttt{Down} & \vcltopframe{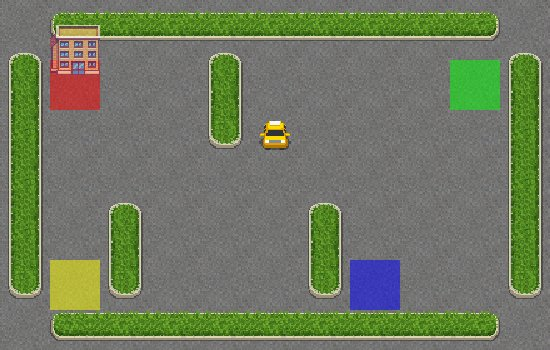} \\
\arrayrulecolor{black!12}\hline
7 & \vcltopframe{figures/content/appendix_d/episodes/success_1_gemini31/frame_006.png} & \raggedright\ttfamily The taxi is currently at row 1, column 2, and needs to reach the red building at row 0, column 0. There is a wall to its left blocking direct movement in rows 0 and 1. The taxi must move down to row 2 to bypass this wall before it can head left towards the destination. & \texttt{Down} & \vcltopframe{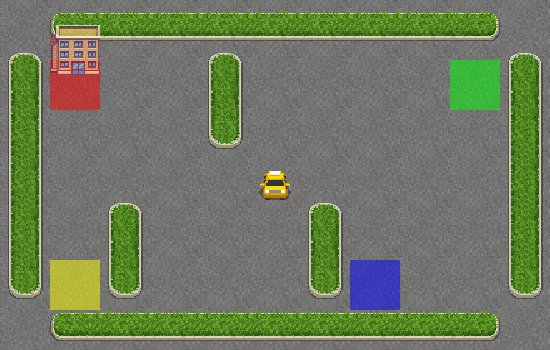} \\
\arrayrulecolor{black!12}\hline
8 & \vcltopframe{figures/content/appendix_d/episodes/success_1_gemini31/frame_007.png} & \raggedright\ttfamily The taxi has moved down far enough to clear the wall on its left. Now it can move left towards the column of the destination (the red building). & \texttt{Left} & \vcltopframe{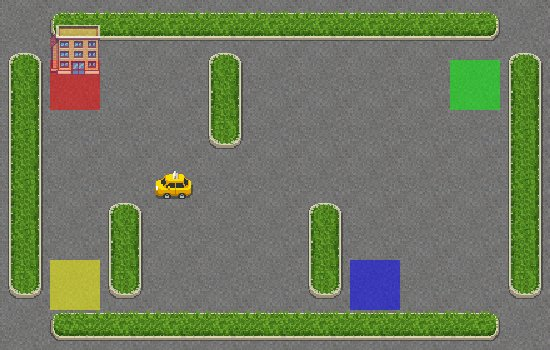} \\
\arrayrulecolor{black!12}\hline
9 & \vcltopframe{figures/content/appendix_d/episodes/success_1_gemini31/frame_008.png} & \raggedright\ttfamily The taxi is now clear of the vertical barrier above it and can move left to reach the same column as the destination (the red square). & \texttt{Left} & \vcltopframe{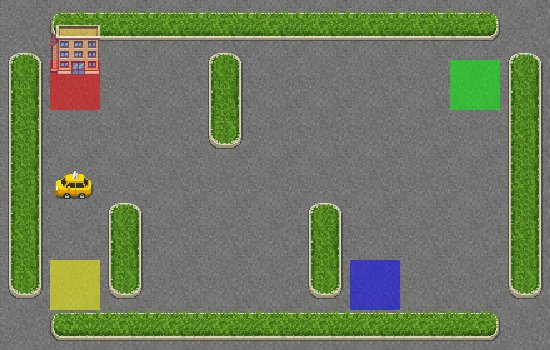} \\
\arrayrulecolor{black!12}\hline
10 & \vcltopframe{figures/content/appendix_d/episodes/success_1_gemini31/frame_009.png} & \raggedright\ttfamily The taxi is now in the same column as the destination (the red square at the top left). Moving up will bring the taxi directly toward the drop-off location. & \texttt{Up} & \vcltopframe{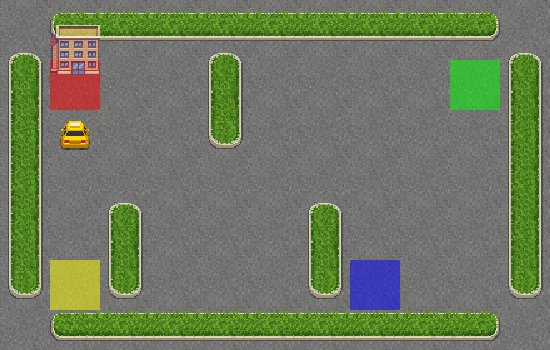} \\
\arrayrulecolor{black!12}\hline
11 & \vcltopframe{figures/content/appendix_d/episodes/success_1_gemini31/frame_010.png} & \raggedright\ttfamily The taxi is directly below the red destination square. Moving up will place the taxi on the destination square, ready for drop-off. & \texttt{Up} & \vcltopframe{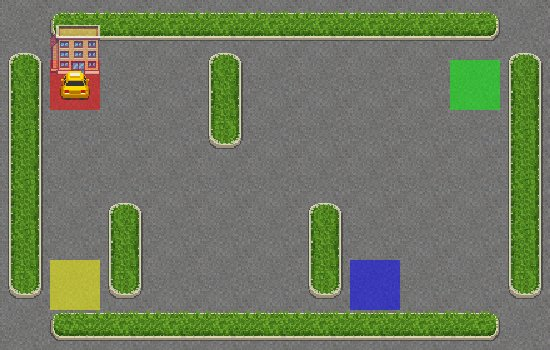} \\
\arrayrulecolor{black!12}\hline
12 & \vcltopframe{figures/content/appendix_d/episodes/success_1_gemini31/frame_011.png} & \raggedright\ttfamily The taxi has arrived at the destination (the red square with the building) with the passenger. The correct action is to drop off the passenger to complete the task. & \texttt{Dropoff} & \vcltopframe{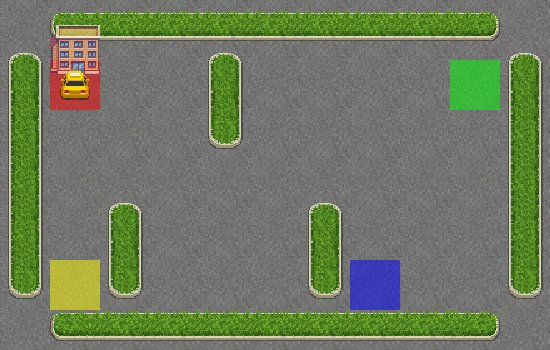} \\
\end{longtable}
\arrayrulecolor{black}
\endgroup

\paragraph{Success case 2: Gemini-3.6-Flash on \texttt{CustomLavaCrossing-v0}.}
\begin{tcolorbox}[enhanced,breakable,colback=green!3,colframe=green!45!black,boxrule=0.45pt,arc=1.2mm,left=1.5mm,right=1.5mm,top=1mm,bottom=1mm]
\textbf{Key insight (round 1).} The model combines heading and local hazard geometry: moving forward would enter lava, so it first rotates toward the safe corridor.
\par\smallskip
\textit{Logged rationale:} The agent is currently facing Right, directly towards lava. Moving Forward would result in stepping into the lava and ending the game. Turning Right will reorient the agent to face Down towards the safe floor tile below.
\end{tcolorbox}
\begingroup
\footnotesize
\setlength{\tabcolsep}{2pt}
\renewcommand{\arraystretch}{1.05}
\arrayrulecolor{black!18}
\begin{longtable}{@{}c c p{8.0cm} c c@{}}
\caption{Success case 2: Gemini-3.6-Flash on CustomLavaCrossing-v0 (55die-seed1952; victory; 19 steps; score 100).}
\label{tab:appendix-d-success-2}\\
\toprule
$t$ & \makecell{Current\\state} & Visible model rationale & Action & \makecell{Next\\state} \\
\midrule
\endfirsthead
\multicolumn{5}{c}{\tablename\ \thetable\ (continued)}\\
\toprule
$t$ & \makecell{Current\\state} & Visible model rationale & Action & \makecell{Next\\state} \\
\midrule
\endhead
\midrule
\multicolumn{5}{r}{Continued on next page}\\
\endfoot
\bottomrule
\endlastfoot
1 & \vcltopframe{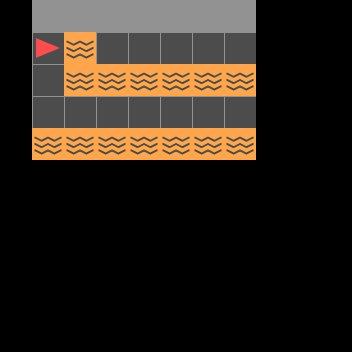} & \raggedright\ttfamily The agent is currently facing Right, directly towards lava. Moving Forward would result in stepping into the lava and ending the game. Turning Right will reorient the agent to face Down towards the safe floor tile below. & \texttt{Right} & \vcltopframe{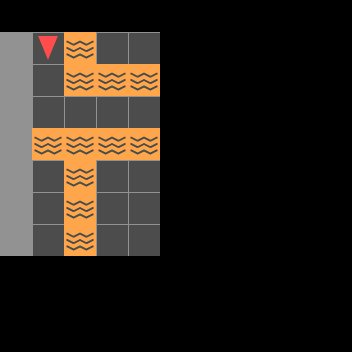} \\
\arrayrulecolor{black!12}\hline
2 & \vcltopframe{figures/content/appendix_d/episodes/success_2_gemini36/frame_001.png} & \raggedright\ttfamily The agent is facing Down, towards a safe floor tile. Moving forward advances along the path. & \texttt{Forward} & \vcltopframe{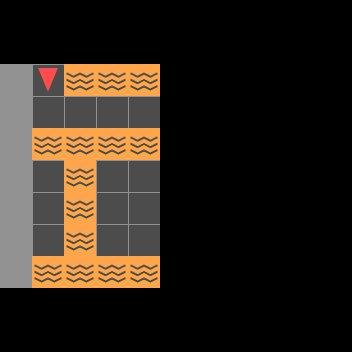} \\
\arrayrulecolor{black!12}\hline
3 & \vcltopframe{figures/content/appendix_d/episodes/success_2_gemini36/frame_002.png} & \raggedright\ttfamily The agent is currently facing Down at row 2, column 1. Directly ahead (row 3, column 1) is a safe dark grey floor tile. Moving Forward will safely step onto this tile, reaching the clear horizontal corridor. & \texttt{Forward} & \vcltopframe{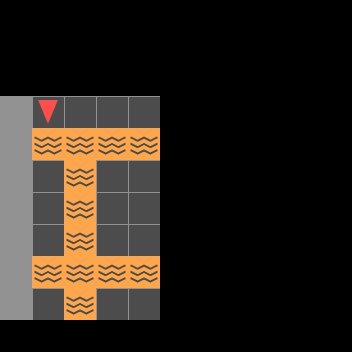} \\
\arrayrulecolor{black!12}\hline
4 & \vcltopframe{figures/content/appendix_d/episodes/success_2_gemini36/frame_003.png} & \raggedright\ttfamily The agent is on the tile above the horizontal lava wall, facing Down (towards the lava). To move right along the clear horizontal corridor, the agent must turn Left so it faces Right. & \texttt{Left} & \vcltopframe{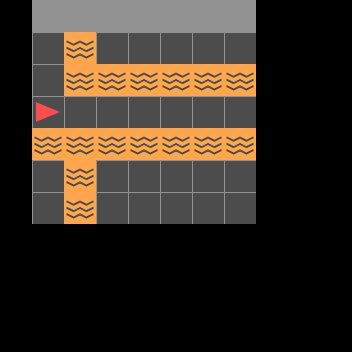} \\
\arrayrulecolor{black!12}\hline
5 & \vcltopframe{figures/content/appendix_d/episodes/success_2_gemini36/frame_004.png} & \raggedright\ttfamily The agent is now facing Right down a clear horizontal corridor of safe floor tiles. Moving Forward moves further along the corridor. & \texttt{Forward} & \vcltopframe{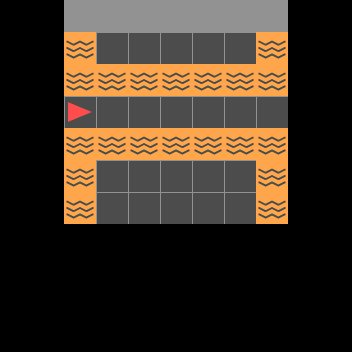} \\
\arrayrulecolor{black!12}\hline
6 & \vcltopframe{figures/content/appendix_d/episodes/success_2_gemini36/frame_005.png} & \raggedright\ttfamily Continue moving forward along the clear horizontal path of safe floor tiles. & \texttt{Forward} & \vcltopframe{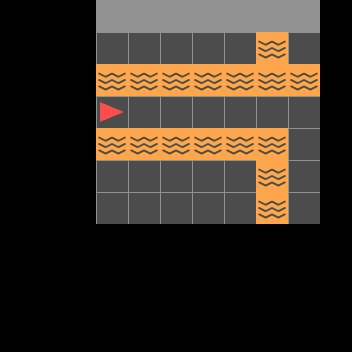} \\
\arrayrulecolor{black!12}\hline
7 & \vcltopframe{figures/content/appendix_d/episodes/success_2_gemini36/frame_006.png} & \raggedright\ttfamily Continue moving forward along the clear horizontal corridor of floor tiles. & \texttt{Forward} & \vcltopframe{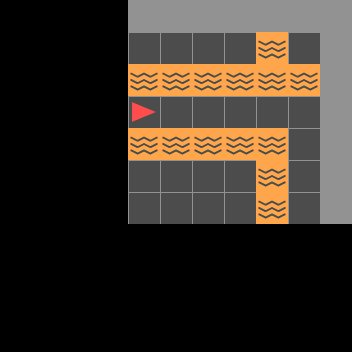} \\
\arrayrulecolor{black!12}\hline
8 & \vcltopframe{figures/content/appendix_d/episodes/success_2_gemini36/frame_007.png} & \raggedright\ttfamily Continue moving forward along the clear horizontal path towards the right. & \texttt{Forward} & \vcltopframe{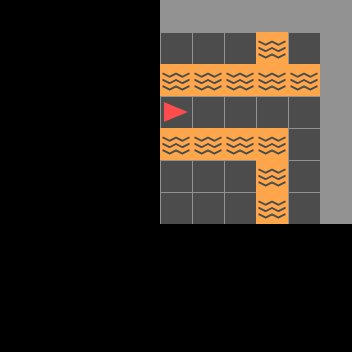} \\
\arrayrulecolor{black!12}\hline
9 & \vcltopframe{figures/content/appendix_d/episodes/success_2_gemini36/frame_008.png} & \raggedright\ttfamily Continue moving forward along the clear horizontal corridor of floor tiles towards the right end. & \texttt{Forward} & \vcltopframe{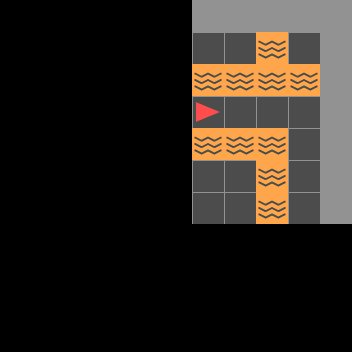} \\
\arrayrulecolor{black!12}\hline
10 & \vcltopframe{figures/content/appendix_d/episodes/success_2_gemini36/frame_009.png} & \raggedright\ttfamily Continue moving forward along the clear horizontal corridor of floor tiles towards the right. & \texttt{Forward} & \vcltopframe{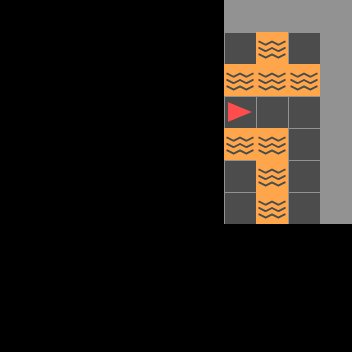} \\
\arrayrulecolor{black!12}\hline
11 & \vcltopframe{figures/content/appendix_d/episodes/success_2_gemini36/frame_010.png} & \raggedright\ttfamily Continue moving forward along the safe floor tile corridor towards the right end. & \texttt{Forward} & \vcltopframe{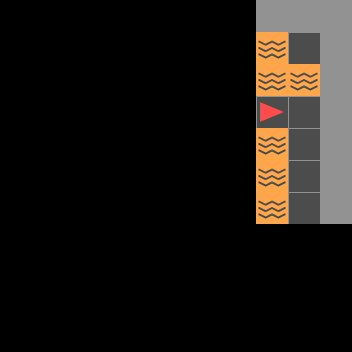} \\
\arrayrulecolor{black!12}\hline
12 & \vcltopframe{figures/content/appendix_d/episodes/success_2_gemini36/frame_011.png} & \raggedright\ttfamily Move forward onto the last safe tile in this row. & \texttt{Forward} & \vcltopframe{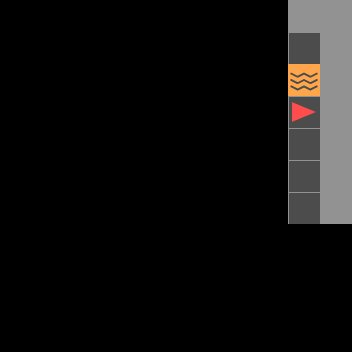} \\
\arrayrulecolor{black!12}\hline
13 & \vcltopframe{figures/content/appendix_d/episodes/success_2_gemini36/frame_012.png} & \raggedright\ttfamily The agent reached the rightmost end of the corridor, facing Right towards the outer boundary wall. Turning Right will reorient the agent to face Down towards the safe floor tiles below. & \texttt{Right} & \vcltopframe{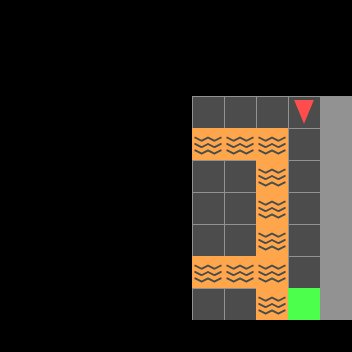} \\
\arrayrulecolor{black!12}\hline
14 & \vcltopframe{figures/content/appendix_d/episodes/success_2_gemini36/frame_013.png} & \raggedright\ttfamily The green goal is visible at the bottom of this column of safe floor tiles. Moving Forward moves down the column towards the goal. & \texttt{Forward} & \vcltopframe{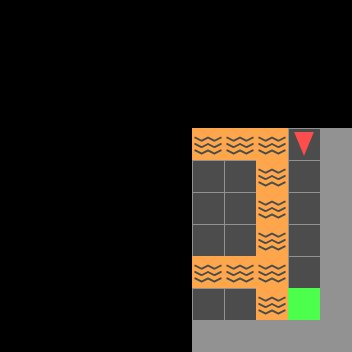} \\
\arrayrulecolor{black!12}\hline
15 & \vcltopframe{figures/content/appendix_d/episodes/success_2_gemini36/frame_014.png} & \raggedright\ttfamily Continue moving forward (downwards) towards the green goal tile. & \texttt{Forward} & \vcltopframe{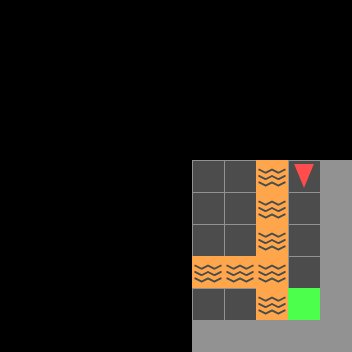} \\
\arrayrulecolor{black!12}\hline
16 & \vcltopframe{figures/content/appendix_d/episodes/success_2_gemini36/frame_015.png} & \raggedright\ttfamily Continue moving forward down the column towards the green goal tile. & \texttt{Forward} & \vcltopframe{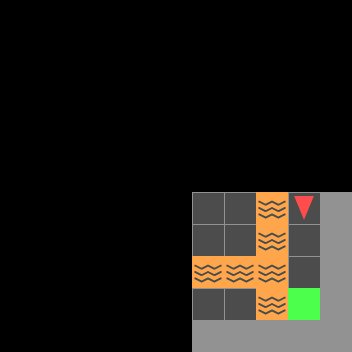} \\
\arrayrulecolor{black!12}\hline
17 & \vcltopframe{figures/content/appendix_d/episodes/success_2_gemini36/frame_016.png} & \raggedright\ttfamily Continue moving forward down the column toward the green goal tile. & \texttt{Forward} & \vcltopframe{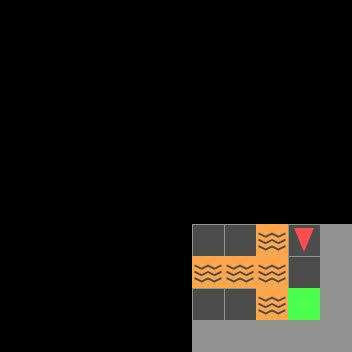} \\
\arrayrulecolor{black!12}\hline
18 & \vcltopframe{figures/content/appendix_d/episodes/success_2_gemini36/frame_017.png} & \raggedright\ttfamily Continue moving forward down the column toward the goal tile. & \texttt{Forward} & \vcltopframe{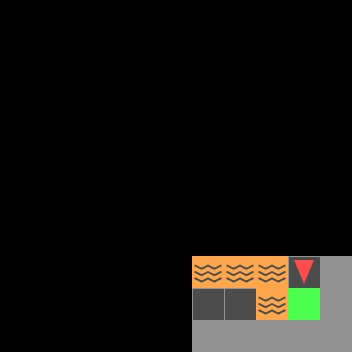} \\
\arrayrulecolor{black!12}\hline
19 & \vcltopframe{figures/content/appendix_d/episodes/success_2_gemini36/frame_018.png} & \raggedright\ttfamily The agent is now positioned directly above the green goal tile, facing Down. Moving Forward will step onto the goal tile to finish the game. & \texttt{Forward} & \vcltopframe{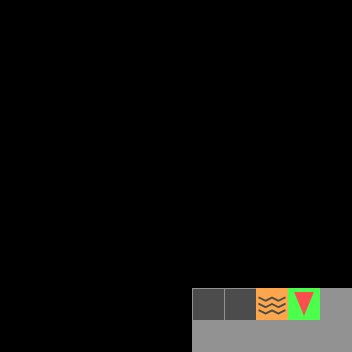} \\
\end{longtable}
\arrayrulecolor{black}
\endgroup

\paragraph{Success case 3: Seed-2.1-Pro on \texttt{CustomUnlockPickup-v0}.}
\begin{tcolorbox}[enhanced,breakable,colback=green!3,colframe=green!45!black,boxrule=0.45pt,arc=1.2mm,left=1.5mm,right=1.5mm,top=1mm,bottom=1mm]
\textbf{Key insight (round 17).} The final decision depends on the full subgoal history: the key has been used and dropped, the door is open, inventory is empty, and the target box is directly ahead.
\par\smallskip
\textit{Logged rationale:} We are now directly adjacent to the yellow box, facing it, and no longer holding the key (we dropped it safely earlier, so our inventory is empty). Using Pickup will collect the box, completing the objective of finding the key, unlocking the door, and picking up the box in the other room, resulting in victory. This matches the final step of the winning example run, where Pickup is used to collect the box once adjacent to it.
\end{tcolorbox}
\begingroup
\footnotesize
\setlength{\tabcolsep}{2pt}
\renewcommand{\arraystretch}{1.05}
\arrayrulecolor{black!18}
\begin{longtable}{@{}c c p{8.0cm} c c@{}}
\caption{Success case 3: Seed-2.1-Pro on CustomUnlockPickup-v0 (55-seed3445; victory; 17 steps; score 100).}
\label{tab:appendix-d-success-3}\\
\toprule
$t$ & \makecell{Current\\state} & Visible model rationale & Action & \makecell{Next\\state} \\
\midrule
\endfirsthead
\multicolumn{5}{c}{\tablename\ \thetable\ (continued)}\\
\toprule
$t$ & \makecell{Current\\state} & Visible model rationale & Action & \makecell{Next\\state} \\
\midrule
\endhead
\midrule
\multicolumn{5}{r}{Continued on next page}\\
\endfoot
\bottomrule
\endlastfoot
1 & \vcltopframe{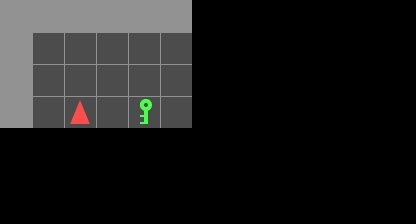} & \raggedright\ttfamily The agent starts facing north (triangle pointing up). The key is located to the east (right) of the agent in the same row. Turning right (clockwise) will face the agent east, directly toward the key, which is the necessary first step to reach the key efficiently, minimizing total steps. All other initial actions either face away from the key, move away from the key, or are useless (Pickup/Drop/Toggle/Done have no valid target/use in the starting state). & \texttt{Right} & \vcltopframe{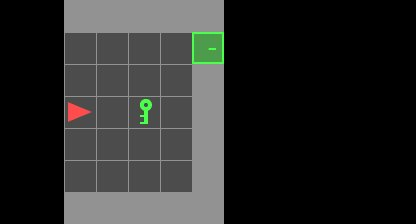} \\
\arrayrulecolor{black!12}\hline
2 & \vcltopframe{figures/content/appendix_d/episodes/success_3_seed21/frame_001.png} & \raggedright\ttfamily The agent is now facing east (right, towards the key). The key is two cells east along the same row, with one empty cell between the agent and key. Moving Forward advances the agent one cell closer to the key, which is the next step to reach the key efficiently. All other actions would either turn away from the key or have no valid use at this state. & \texttt{Forward} & \vcltopframe{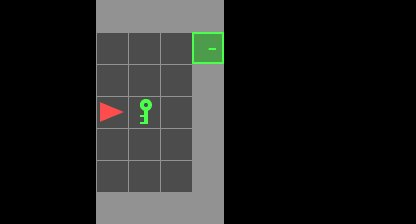} \\
\arrayrulecolor{black!12}\hline
3 & \vcltopframe{figures/content/appendix_d/episodes/success_3_seed21/frame_002.png} & \raggedright\ttfamily The agent is now adjacent to the green key, facing directly toward it. We cannot walk over the key (per rules), so picking up the key from this adjacent position is the correct next step to collect the key, matching the example gameplay flow where pickup occurs immediately after moving adjacent to the key. Other actions would either move into the key (invalid), turn away from the key, or serve no purpose at this state. & \texttt{Pickup} & \vcltopframe{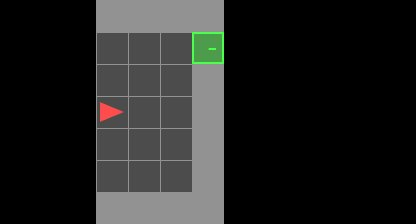} \\
\arrayrulecolor{black!12}\hline
4 & \vcltopframe{figures/content/appendix_d/episodes/success_3_seed21/frame_003.png} & \raggedright\ttfamily We have successfully collected the key, and the door is located in the top-right (northeast) of the starting room. Following the proven shortest path from the victorious example run, turning left (counter-clockwise) from facing east now faces us north (upwards toward the top row where the door sits). This path has the same minimal step count as moving east first, and matches the optimal flow used in the winning gameplay: turn north, move up to the top row, turn east to face the door, then approach and unlock it. Other actions would either add unnecessary steps or turn away from the direction of the door. & \texttt{Left} & \vcltopframe{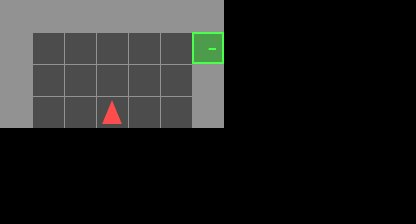} \\
\arrayrulecolor{black!12}\hline
5 & \vcltopframe{figures/content/appendix_d/episodes/success_3_seed21/frame_004.png} & \raggedright\ttfamily The agent is now facing north (upward, toward the top row where the door is located), positioned in the center of the current view. Moving Forward advances us one cell up (north) toward the top row, matching the victorious example flow where after turning to face north, the agent moves straight up to the door's row before turning east to approach the door. There are no obstacles in this direction, so this is the most efficient next step to progress toward the door. & \texttt{Forward} & \vcltopframe{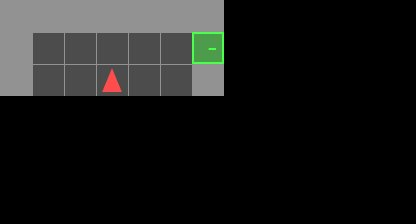} \\
\arrayrulecolor{black!12}\hline
6 & \vcltopframe{figures/content/appendix_d/episodes/success_3_seed21/frame_005.png} & \raggedright\ttfamily We are now one cell below the top (door) row, facing north. Moving Forward again brings us to the top row, the same horizontal level as the green door. This matches the optimal path from the winning example, where two forward moves north after turning reach the door's row, allowing us to next turn east to face the door. There are no obstacles blocking this move, so this is the most efficient next step. & \texttt{Forward} & \vcltopframe{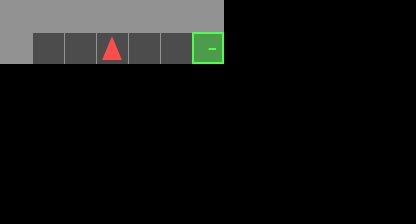} \\
\arrayrulecolor{black!12}\hline
7 & \vcltopframe{figures/content/appendix_d/episodes/success_3_seed21/frame_006.png} & \raggedright\ttfamily We are now on the top (door) row, facing north (into the top wall of the room, can't move further north). The green door is directly to our east (right). Turning right (clockwise) from facing north faces us east, directly toward the door, matching the winning example flow: after reaching the top row, turn right to face the door before moving forward to it. This is the only turn needed to face the door, allowing us to advance to the door next. & \texttt{Right} & \vcltopframe{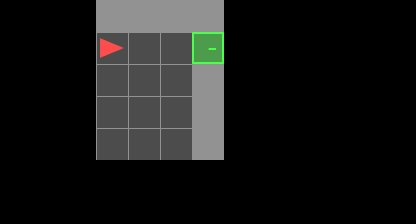} \\
\arrayrulecolor{black!12}\hline
8 & \vcltopframe{figures/content/appendix_d/episodes/success_3_seed21/frame_007.png} & \raggedright\ttfamily We are now facing east (right), directly toward the green door on the right end of the top row. Moving Forward advances us one cell closer to the door, following the optimal path: after turning to face the door, move east until we are adjacent to it, then unlock it with Toggle. There are no obstacles between us and the door, so this is the correct efficient next step. & \texttt{Forward} & \vcltopframe{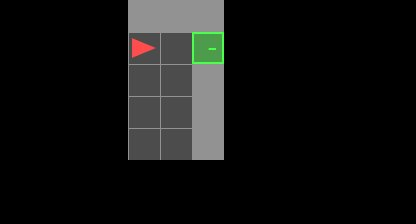} \\
\arrayrulecolor{black!12}\hline
9 & \vcltopframe{figures/content/appendix_d/episodes/success_3_seed21/frame_008.png} & \raggedright\ttfamily We are now one cell away from the green door, facing directly east toward it. Moving Forward brings us into the cell immediately adjacent to the door, positioning us to unlock (toggle) it open in the next step. This follows the optimal path, advancing directly to the door without extra steps or turns. & \texttt{Forward} & \vcltopframe{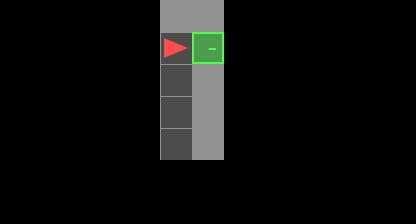} \\
\arrayrulecolor{black!12}\hline
10 & \vcltopframe{figures/content/appendix_d/episodes/success_3_seed21/frame_009.png} & \raggedright\ttfamily We are now directly adjacent to the green door, facing it, and holding the matching green key. Using Toggle will unlock and open the door, allowing us to proceed into the next room to reach the box, exactly as in the victorious example gameplay where Toggle is used to open the door immediately after reaching it with the key. This is the required next step to progress past the door. & \texttt{Toggle} & \vcltopframe{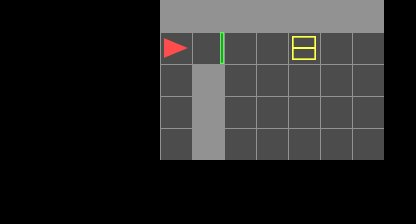} \\
\arrayrulecolor{black!12}\hline
11 & \vcltopframe{figures/content/appendix_d/episodes/success_3_seed21/frame_010.png} & \raggedright\ttfamily We cannot drop the key while facing east, as this would place the key directly in the open door cell (forbidden by the rule against dropping keys on doors, even when open). Facing east, a single right turn faces us south, toward the empty valid cell directly below our current position (not a door cell, and a cell we will never need to step into later, so no risk of walking over the key). This sets us up to safely drop the key next, then turn back to east to proceed straight through the door to the box with no further turns needed, the shortest safe path to drop the key before advancing to the goal. & \texttt{Right} & \vcltopframe{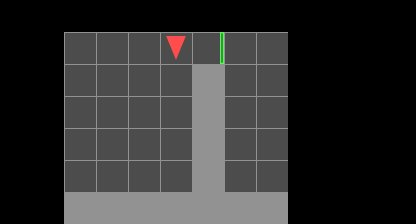} \\
\arrayrulecolor{black!12}\hline
12 & \vcltopframe{figures/content/appendix_d/episodes/success_3_seed21/frame_011.png} & \raggedright\ttfamily We are now facing south, directly toward an empty non-door cell below our position. Dropping the key here places it in this valid, non-door cell, fully complying with the rule against dropping keys on doors. We will never need to enter this cell as we proceed east to the box, so there is no risk of walking over the key later, satisfying all game rules. This safely discards the key (which we no longer need, as the door is already open) so we can proceed to pick up the box. & \texttt{Drop} & \vcltopframe{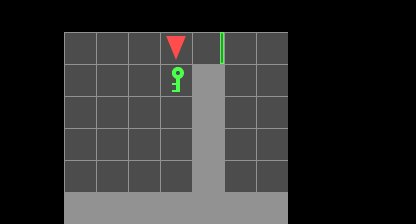} \\
\arrayrulecolor{black!12}\hline
13 & \vcltopframe{figures/content/appendix_d/episodes/success_3_seed21/frame_012.png} & \raggedright\ttfamily The key is now safely dropped in a valid non-door cell, and we no longer need to carry it. We are currently facing south; turning left (counter-clockwise) once turns us back to face east, directly toward the open door and the box in the room beyond. This orients us correctly to move straight east through the door to the box with no additional turns needed, the shortest path forward. & \texttt{Left} & \vcltopframe{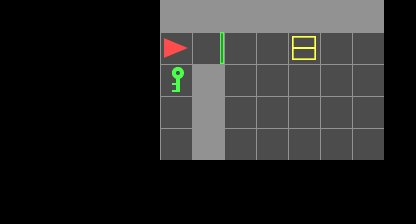} \\
\arrayrulecolor{black!12}\hline
14 & \vcltopframe{figures/content/appendix_d/episodes/success_3_seed21/frame_013.png} & \raggedright\ttfamily We are now facing east again, key safely dropped, with the open door directly in front of us and the yellow box visible further east along the same row. Moving Forward advances us through the open doorway into the second room, progressing directly toward the box with no obstacles in our path, following the shortest straight-line path to the goal. & \texttt{Forward} & \vcltopframe{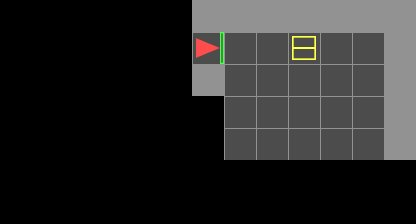} \\
\arrayrulecolor{black!12}\hline
15 & \vcltopframe{figures/content/appendix_d/episodes/success_3_seed21/frame_014.png} & \raggedright\ttfamily We are now positioned at the open doorway, still facing east. Moving Forward carries us fully into the second room, continuing straight along the top row toward the yellow box. There are no obstacles between us and the goal, so this is the most direct next step to close the distance to the box. & \texttt{Forward} & \vcltopframe{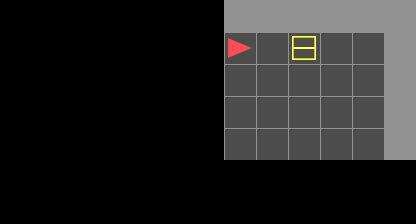} \\
\arrayrulecolor{black!12}\hline
16 & \vcltopframe{figures/content/appendix_d/episodes/success_3_seed21/frame_015.png} & \raggedright\ttfamily We are now in the second room, still facing east, with one empty cell between us and the yellow box. Moving Forward brings us into the cell directly adjacent to the box, positioning us to pick up the box in the next step to complete the task. This continues our straight, shortest path to the goal with no turns or extra steps needed. & \texttt{Forward} & \vcltopframe{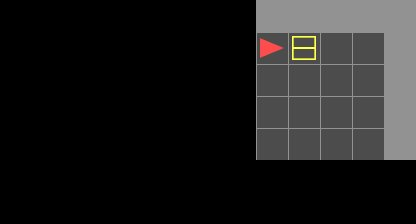} \\
\arrayrulecolor{black!12}\hline
17 & \vcltopframe{figures/content/appendix_d/episodes/success_3_seed21/frame_016.png} & \raggedright\ttfamily We are now directly adjacent to the yellow box, facing it, and no longer holding the key (we dropped it safely earlier, so our inventory is empty). Using Pickup will collect the box, completing the objective of finding the key, unlocking the door, and picking up the box in the other room, resulting in victory. This matches the final step of the winning example run, where Pickup is used to collect the box once adjacent to it. & \texttt{Pickup} & \vcltopframe{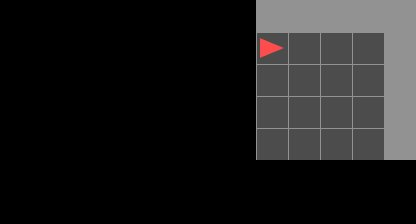} \\
\end{longtable}
\arrayrulecolor{black}
\endgroup

\subsubsection{Complete Same-Start-State Episodes}
\label{app:complete-model-episodes}

The controlled episodes below compare six models from exactly the same
\texttt{Seaquest} initial state (the configuration with 350 skipped
initialization steps, beginning at the surface as new targets enter). We use
\texttt{Seaquest} because its movement, combat, surfacing, and survival choices
admit substantially more behavioral variation than the short, nearly unique
optimal route in \texttt{CliffWalking}. Each panel is copied directly from the
corresponding run's saved \texttt{trajectory.png}; no reasoning transcript is
overlaid. Scores are read from the matching \texttt{eval\_result.json} records.

\begin{figure}[!htbp]
  \centering
  \begin{tabular}{@{}r@{\hspace{0.8em}}c@{}}
    \shortstack[r]{\scriptsize Gemini-3.1-Pro\\[-0.15em]\scriptsize Score: 50} &
    \includegraphics[width=0.78\textwidth,height=0.155\textheight,keepaspectratio]{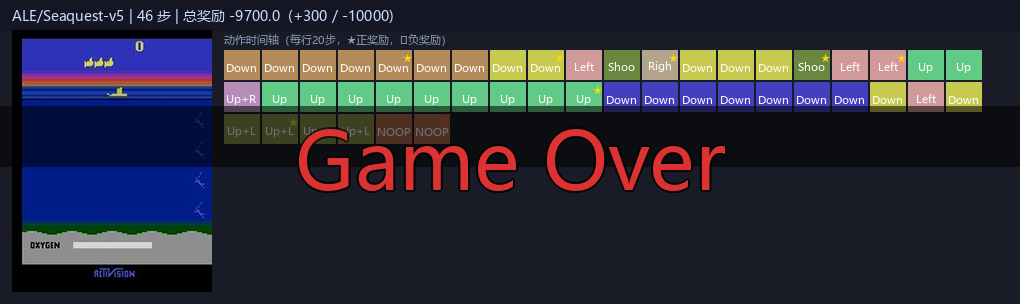}\\[-0.3ex]
    \shortstack[r]{\scriptsize Gemini-3.6-Flash\\[-0.15em]\scriptsize Score: 50} &
    \includegraphics[width=0.78\textwidth,height=0.155\textheight,keepaspectratio]{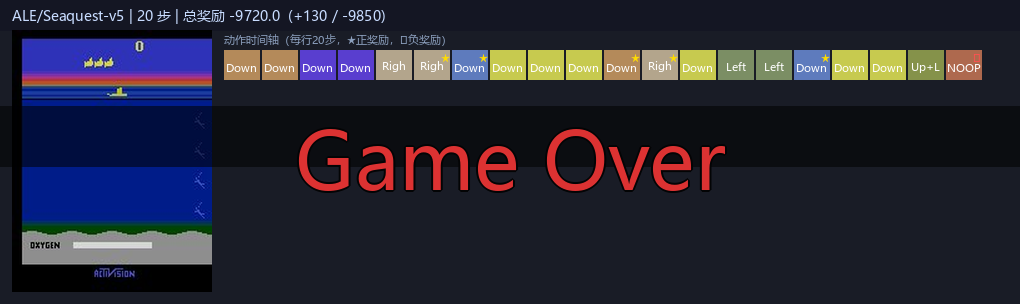}\\[-0.3ex]
    \shortstack[r]{\scriptsize GPT-5.6-sol\\[-0.15em]\scriptsize Score: 56} &
    \includegraphics[width=0.78\textwidth,height=0.155\textheight,keepaspectratio]{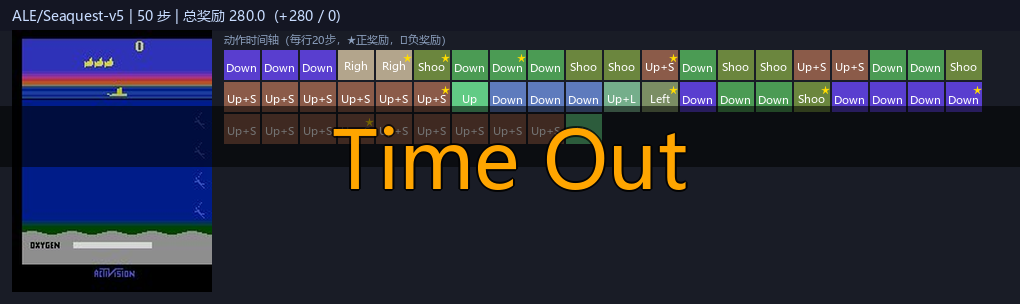}\\[-0.3ex]
  \end{tabular}
  \caption{Complete same-start-state \texttt{Seaquest} episodes using the
  trajectory images saved by the evaluation pipeline. Panel labels report the
  normalized environment score (0--100), rather than pass/fail or raw reward.}
  \label{fig:appendix-d-complete-episodes}
\end{figure}

\begin{figure}[!htbp]
  \ContinuedFloat
  \centering
  \begin{tabular}{@{}r@{\hspace{0.8em}}c@{}}
    \shortstack[r]{\scriptsize Qwen3.5-397B-A17B\\[-0.15em]\scriptsize Score: 36} &
    \includegraphics[width=0.78\textwidth,height=0.155\textheight,keepaspectratio]{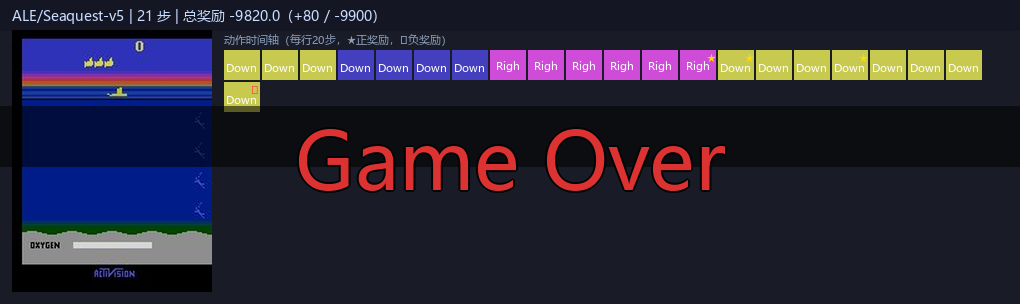}\\[-0.3ex]
    \shortstack[r]{\scriptsize Gemma-4-31B-it\\[-0.15em]\scriptsize Score: 36} &
    \includegraphics[width=0.78\textwidth,height=0.155\textheight,keepaspectratio]{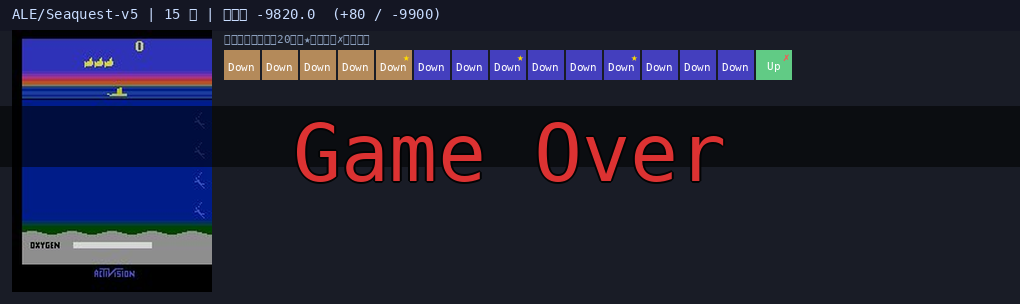}\\[-0.3ex]
    \shortstack[r]{\scriptsize Qwen2.5-72B\\[-0.15em]\scriptsize Score: 20} &
    \includegraphics[width=0.78\textwidth,height=0.155\textheight,keepaspectratio]{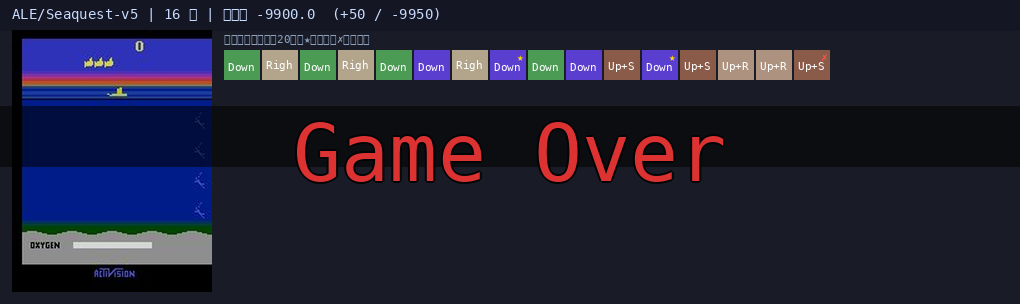}
  \end{tabular}
  \caption[]{Complete same-start-state \texttt{Seaquest} episodes (continued).}
\end{figure}

\FloatBarrier
\subsection{Visualization of Model Behavior}
\label{app:model-behavior-visualization}

\subsubsection{Action Distributions and Bias Audit}
\label{app:action-distributions}

\begin{figure}[H]
  \centering
  \includegraphics[width=\textwidth]{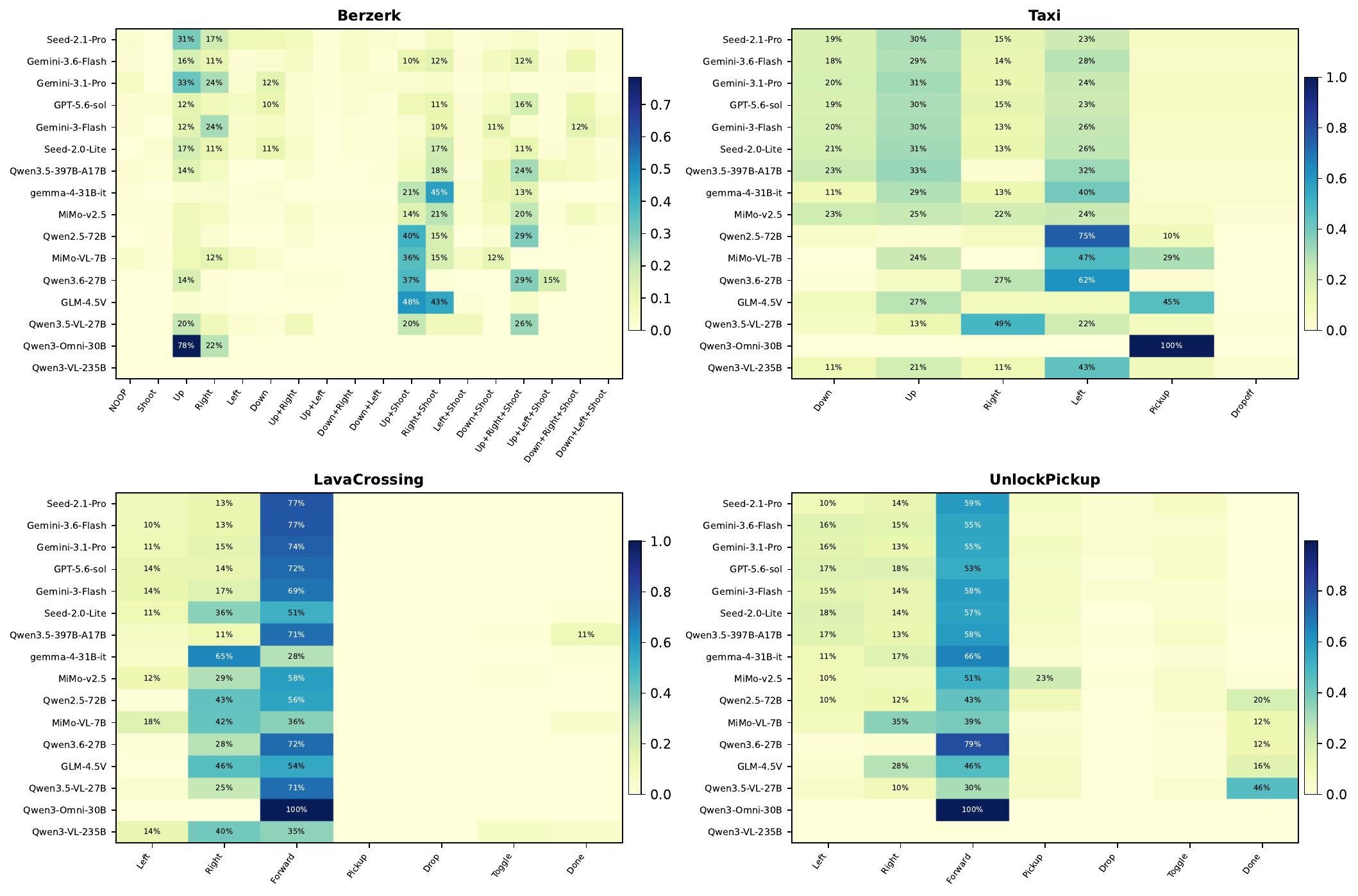}
  \caption{Action-frequency heatmaps for 16 models in Berzerk, Taxi,
  LavaCrossing, and UnlockPickup. Each row is normalized to sum to one within a
  model and environment; blank cells indicate that the action was never executed.}
  \label{fig:appendix-d-action-heatmaps}
\end{figure}

We aggregate all executed actions from all available runs of Berzerk, Taxi,
LavaCrossing, and UnlockPickup and normalize frequencies independently for each
model--environment pair. Berzerk replaces CliffWalking in this analysis because it has
17 evaluated configurations per model, a richer action space, and substantially
more varied combat and navigation behavior.
Figure~\ref{fig:appendix-d-action-heatmaps} therefore describes policy shape,
not the number or length of completed episodes. The dominant action is often
task-induced---for example, Berzerk policies may repeatedly combine movement
and shooting---so raw concentration alone is not treated as evidence of model
bias.

Table~\ref{tab:appendix-d-action-bias} controls for the environment-specific policy by
comparing each model against the pooled action distribution for that same
environment. Under the stated threshold, seven models show a suspected preference:
GLM-4.5V over-selects \texttt{Up+Shoot} and Qwen3-Omni-30B over-selects
\texttt{Up} in Berzerk; gemma-4-31B-it over-selects \texttt{Right} in
LavaCrossing; Qwen2.5-72B and Qwen3.6-27B over-select \texttt{Left} in Taxi;
MiMo-VL-7B over-selects \texttt{Right} and Qwen3.5-27B over-selects
\texttt{Done} in UnlockPickup. These are diagnostic associations rather than
causal claims: early termination and incomplete subgoal execution can also
produce concentrated action histories.

\begingroup
\scriptsize
\setlength{\tabcolsep}{4pt}
\renewcommand{\arraystretch}{1.08}
\arrayrulecolor{black!18}
\begin{longtable}{@{}p{3.0cm}p{4.1cm}cccc@{}}
\caption{Action-bias audit over Berzerk, Taxi, LavaCrossing, and UnlockPickup. For each model, among model--game groups with at least 20 actions, we report its largest positive action-frequency deviation from the pooled distribution for the same game. TV is total-variation distance; a bias flag additionally requires a frequency excess of at least 0.20 and $\mathrm{TV}\geq0.20$.}
\label{tab:appendix-d-action-bias}\\
\toprule
Model & Largest excess & Model & Pool & TV & Bias? \\
\midrule
\endfirsthead
\multicolumn{6}{c}{\tablename\ \thetable\ (continued)}\\
\toprule
Model & Largest excess & Model & Pool & TV & Bias? \\
\midrule
\endhead
\midrule
\multicolumn{6}{r}{Continued on next page}\\
\endfoot
\bottomrule
\endlastfoot
Seed-2.1-Pro & Berzerk: Up & 31\% & 15\% & 0.45 & No \\
\arrayrulecolor{black!12}\hline
Gemini-3.6-Flash & LavaCrossing: Forward & 77\% & 63\% & 0.15 & No \\
\arrayrulecolor{black!12}\hline
Gemini-3.1-Pro & Berzerk: Up & 33\% & 15\% & 0.49 & No \\
\arrayrulecolor{black!12}\hline
GPT-5.6-sol & LavaCrossing: Forward & 72\% & 63\% & 0.14 & No \\
\arrayrulecolor{black!12}\hline
Gemini-3-Flash & Berzerk: Right & 24\% & 9\% & 0.38 & No \\
\arrayrulecolor{black!12}\hline
Seed-2.0-Lite & LavaCrossing: Right & 36\% & 26\% & 0.12 & No \\
\arrayrulecolor{black!12}\hline
Qwen3.5-397B-A17B & Berzerk: Up+Right+Shoot & 24\% & 14\% & 0.25 & No \\
\arrayrulecolor{black!12}\hline
gemma-4-31B-it & LavaCrossing: Right & 65\% & 26\% & 0.39 & Yes \\
\arrayrulecolor{black!12}\hline
MiMo-v2.5 & UnlockPickup: Pickup & 23\% & 7\% & 0.17 & No \\
\arrayrulecolor{black!12}\hline
Qwen2.5-72B & Taxi: Left & 75\% & 33\% & 0.45 & Yes \\
\arrayrulecolor{black!12}\hline
MiMo-VL-7B & UnlockPickup: Right & 35\% & 14\% & 0.26 & Yes \\
\arrayrulecolor{black!12}\hline
Qwen3.6-27B & Taxi: Left & 62\% & 33\% & 0.40 & Yes \\
\arrayrulecolor{black!12}\hline
GLM-4.5V & Berzerk: Up+Shoot & 48\% & 17\% & 0.57 & Yes \\
\arrayrulecolor{black!12}\hline
Qwen3.5-VL-27B & UnlockPickup: Done & 46\% & 8\% & 0.38 & Yes \\
\arrayrulecolor{black!12}\hline
Qwen3-Omni-30B & Berzerk: Up & 78\% & 15\% & 0.75 & Yes \\
\arrayrulecolor{black!12}\hline
Qwen3-VL-235B & LavaCrossing: Right & 40\% & 26\% & 0.28 & No \\
\end{longtable}
\arrayrulecolor{black}
\endgroup

\subsubsection{Same-Start-State Trajectory Comparison}
\label{app:trajectory-comparison}

\begin{figure}[H]
  \centering
  \includegraphics[width=\textwidth]{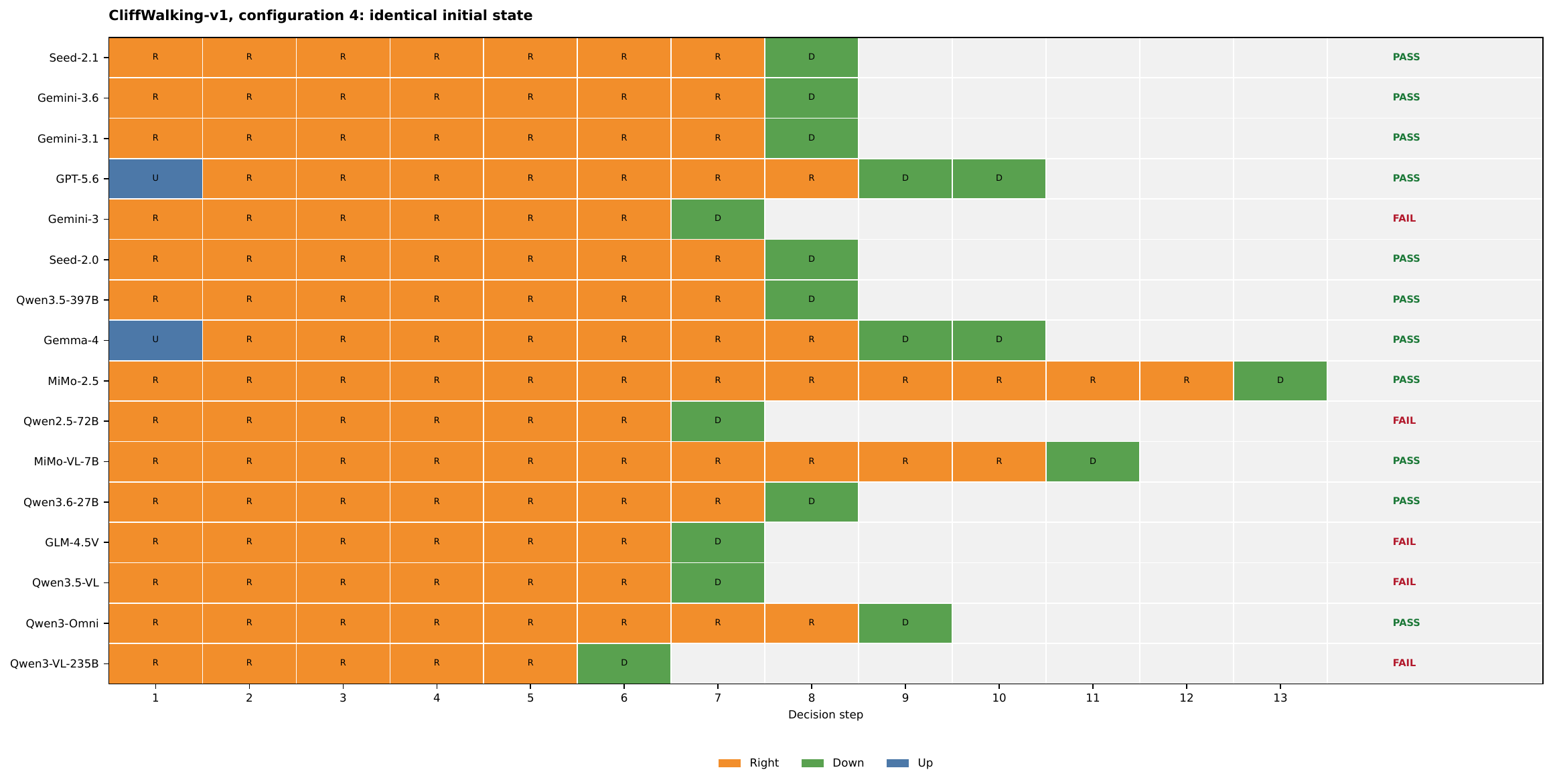}
  \par\medskip
  \includegraphics[width=\textwidth]{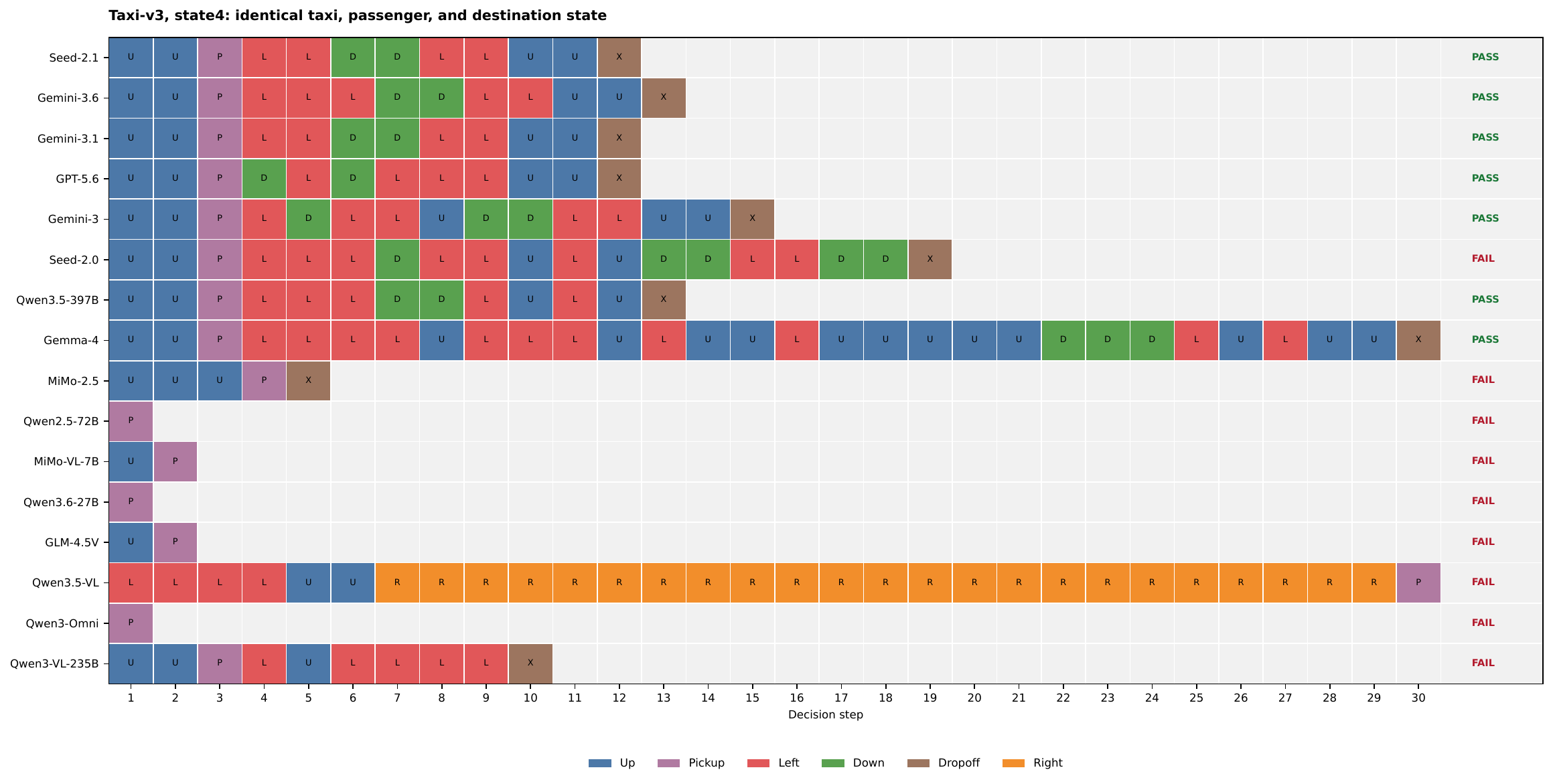}
  \caption{Decision-trajectory comparison under identical initial states.
  Action abbreviations and colors are shared within each panel; \textsc{Pass}
  and \textsc{Fail} denote the recorded episode outcome.}
  \label{fig:appendix-d-trajectories}
\end{figure}

Figure~\ref{fig:appendix-d-trajectories} visualizes decision trajectories on
two fixed initial states. Each colored cell is the executed action at one
decision step; gray suffixes indicate that the episode has already ended. The
CliffWalking panel isolates route choice and terminal-step timing, while the
Taxi panel exposes whether a model sustains the longer
navigate--pickup--navigate--drop-off sequence.

\FloatBarrier
\subsection{Reasoning Quality Analysis}
\label{app:reasoning-quality}

\paragraph{Annotation protocol.}
We analyze only reasoning text present in the saved model responses; no hidden
model state is inferred. Table~\ref{tab:appendix-d-reasoning-patterns}
aggregates every logged round in the four selected environments. A response has a
\emph{rationale} when it contains at least four non-action words. \emph{Demo}
and \emph{history} mark explicit lexical reference to demonstrations or prior
interaction, respectively. The automated \emph{locally coherent} indicator
requires a visible rationale, exactly one action marker, and no unresolved
burst of repeated uncertainty. It measures surface consistency, not whether
the underlying visual belief is correct.

\begingroup
\scriptsize
\setlength{\tabcolsep}{5pt}
\renewcommand{\arraystretch}{1.08}
\arrayrulecolor{black!18}
\begin{longtable}{@{}p{3.5cm}rcccc@{}}
\caption{Model-wise frequency of visible reasoning patterns across all logged rounds in CliffWalking, Taxi, LavaCrossing, and UnlockPickup. Rationale requires at least four non-action words. Demo and history denote explicit lexical references. Locally coherent requires a rationale, exactly one action marker, and no unresolved repeated uncertainty; it is a descriptive surface audit rather than a correctness metric.}
\label{tab:appendix-d-reasoning-patterns}\\
\toprule
Model & $N$ & Rationale & Demo & History & Locally coherent \\
\midrule
\endfirsthead
\toprule
Model & $N$ & Rationale & Demo & History & Locally coherent \\
\midrule
\endhead
\bottomrule
\endfoot
Seed-2.1-Pro & 876 & 100.0\% & 20.8\% & 63.2\% & 100.0\% \\
\arrayrulecolor{black!12}\hline
Gemini-3.6-Flash & 927 & 99.0\% & 2.5\% & 23.9\% & 95.3\% \\
\arrayrulecolor{black!12}\hline
Gemini-3.1-Pro & 763 & 100.0\% & 0.0\% & 27.1\% & 100.0\% \\
\arrayrulecolor{black!12}\hline
GPT-5.6-sol & 892 & 25.6\% & 0.0\% & 10.4\% & 25.6\% \\
\arrayrulecolor{black!12}\hline
Gemini-3-Flash & 800 & 98.8\% & 2.4\% & 34.2\% & 91.9\% \\
\arrayrulecolor{black!12}\hline
Seed-2.0-Lite & 928 & 99.9\% & 5.4\% & 41.6\% & 99.9\% \\
\arrayrulecolor{black!12}\hline
Qwen3.5-397B-A17B & 832 & 74.5\% & 5.2\% & 25.0\% & 74.4\% \\
\arrayrulecolor{black!12}\hline
gemma-4-31B-it & 935 & 100.0\% & 0.9\% & 59.8\% & 99.6\% \\
\arrayrulecolor{black!12}\hline
MiMo-v2.5 & 843 & 75.9\% & 4.7\% & 28.4\% & 75.8\% \\
\arrayrulecolor{black!12}\hline
Qwen2.5-72B & 859 & 100.0\% & 1.2\% & 67.8\% & 100.0\% \\
\arrayrulecolor{black!12}\hline
MiMo-VL-7B & 668 & 100.0\% & 22.3\% & 84.1\% & 36.5\% \\
\arrayrulecolor{black!12}\hline
Qwen3.6-27B & 900 & 86.8\% & 1.0\% & 60.6\% & 81.8\% \\
\arrayrulecolor{black!12}\hline
GLM-4.5V & 775 & 100.0\% & 18.2\% & 96.4\% & 51.6\% \\
\arrayrulecolor{black!12}\hline
Qwen3.5-VL-27B & 910 & 100.0\% & 1.2\% & 48.2\% & 99.3\% \\
\arrayrulecolor{black!12}\hline
Qwen3-Omni-30B & 721 & 0.0\% & 0.0\% & 0.0\% & 0.0\% \\
\arrayrulecolor{black!12}\hline
Qwen3-VL-235B & 248 & 39.1\% & 13.7\% & 23.4\% & 39.1\% \\
\end{longtable}
\arrayrulecolor{black}
\endgroup

\paragraph{Aggregate patterns.}
Most Gemini and Seed responses contain a visible rationale in essentially every
round, whereas GPT-5.6-sol does so in 25.6\% of rounds, Qwen3-VL-235B in
39.1\%, and Qwen3-Omni-30B in none of the logged rounds.

This contrast partly reflects response observability rather than reasoning
ability alone. GPT-5.6-sol's internal reasoning trace is hidden, and the model
does not generally restate or summarize that process in its visible answer.
Consequently, its textual manifestation of reasoning is lower by construction
and should not be interpreted as a direct measure of underlying intelligence.

Explicit demonstration references are uncommon overall, but occur most often
for MiMo-VL-7B
(22.3\%), Seed-2.1-Pro (20.8\%), and GLM-4.5V (18.2\%). Explicit history
language is frequent for GLM-4.5V (96.4\%) and MiMo-VL-7B (84.1\%), yet the
annotated examples show that frequent history reference does not guarantee
correct state tracking: a model may repeatedly revisit a mistaken location
estimate.

\paragraph{Ten representative samples.}
Table~\ref{tab:appendix-d-reasoning-samples} provides a manual audit of ten
rounds from the controlled CliffWalking comparison. Successful reasoning
typically identifies the fatal region, tracks horizontal progress, and changes
action only when the goal column is reached. The principal failure patterns
are premature claims of having reached that column, over-reliance on the
demonstration despite a mismatched current position, repetitive
self-correction without resolving the spatial belief, and action-only outputs
that provide no inspectable rationale.

\begingroup
\footnotesize
\setlength{\tabcolsep}{3pt}
\renewcommand{\arraystretch}{1.08}
\arrayrulecolor{black!18}
\begin{longtable}{@{}>{\raggedright\arraybackslash}p{1.9cm}>{\centering\arraybackslash}p{0.85cm}>{\raggedright\arraybackslash}p{5.0cm}>{\centering\arraybackslash}p{0.8cm}>{\centering\arraybackslash}p{0.7cm}>{\centering\arraybackslash}p{0.7cm}>{\raggedright\arraybackslash}p{\dimexpr\linewidth-9.95cm-12\tabcolsep\relax}@{}}
\caption{Ten representative visible-reasoning samples from the same-start-state CliffWalking episodes. Demo and history labels require explicit reference; coherence is manually assigned from internal consistency and agreement with the observed transition, not merely from fluency.}
\label{tab:appendix-d-reasoning-samples}\\
\toprule
Model & Round & Reasoning excerpt & Demo & Hist. & Coh. & Analysis \\
\midrule
\endfirsthead
\multicolumn{7}{c}{\tablename\ \thetable\ (continued)}\\
\toprule
Model & Round & Reasoning excerpt & Demo & Hist. & Coh. & Analysis \\
\midrule
\endhead
\midrule
\multicolumn{7}{r}{Continued on next page}\\
\endfoot
\bottomrule
\endlastfoot
Seed-2.1 & 1 & \raggedright The goal (cookie) is at the bottom-right corner of the grid. The character is currently on the safe grass row directly above the cliff, to the left of the goal column. Moving down would cause the character to fall off the cliff immediately, moving left travels away from the goal, and moving up adds unnecessary extra steps. Moving right along this safe ... & No & No & Yes & Correctly contrasts the safe horizontal corridor with the fatal downward move. \\
\arrayrulecolor{black!12}\hline
Gemini-3.6 & 8 & \raggedright We are now at the top of column 11, directly above the cookie goal. Moving Down will head towards the goal tile. & No & Yes & Yes & Detects the terminal geometry and switches from repeated Right actions to Down. \\
\arrayrulecolor{black!12}\hline
GPT-5.6 & 1 & \raggedright Move away from the cliff before heading right toward the goal. & No & No & Yes & Chooses a safe but unnecessary detour; the rationale is internally consistent but not shortest-path. \\
\arrayrulecolor{black!12}\hline
MiMo-2.5 & 1 & \raggedright The goal of the game is to reach the pizza located in the bottom-right corner. The environment features a large cliff (the dark brown wall and black void) that blocks direct movement downwards from the middle of the map. In the current state, the character is positioned on the grass in the second row, near the center. Directly below the character is the ... & Yes & No & Yes & Explicitly retrieves the demonstrated safe-route rule and grounds it in the current cliff layout. \\
\arrayrulecolor{black!12}\hline
Qwen2.5-72B & 7 & \raggedright Since I've reached the column directly above the goal, moving down will allow me to reach it and secure victory. This aligns with the strategy displayed in the example gameplay. & Yes & Yes & No & Claims it has reached the goal column and descends one step too early, causing failure. \\
\arrayrulecolor{black!12}\hline
MiMo-VL-7B & 11 & \raggedright Wait, the example video's step 10 was Down. Let's think: the character is at the right edge of the green platform. If we move down, does that get them to the target? Wait, the target is a brown circle. Wait, maybe the target is on the lower level? No, the example video's step 10 was Down and led to Victory. Wait, in the example, the character moved down ... & Yes & Yes & No & Eventually selects the correct terminal action, but repeated self-correction makes the justification unstable. \\
\arrayrulecolor{black!12}\hline
GLM-4.5V & 7 & \raggedright Got it, let's see. The current state shows the character is on the rightmost grass tile, right next to the cookie. Wait, no, wait. Wait, the cookie is on the bottom right. Wait, the character is on the top right grass tile. So to get to the cookie, the character needs to go down, right? Because the cookie is below the cliff. Wait, no, wait. Wait, the ... & Yes & Yes & No & Contains unresolved spatial contradictions before committing to Down; the episode terminates unsuccessfully. \\
\arrayrulecolor{black!12}\hline
Qwen3.5-VL & 7 & \raggedright The character has reached the column directly above the cookie. Moving Down is now safe and will allow the character to collect the cookie and win the game. & Yes & Yes & No & Confidently asserts an incorrect location relative to the goal and descends prematurely. \\
\arrayrulecolor{black!12}\hline
Qwen3.6-27B & 5 & \raggedright The character is progressing along the top row towards the right edge. The goal is at the bottom right. Continuing to move Right is necessary to reach the corner where the character can safely descend to the goal. & No & Yes & Yes & Maintains a concise progress estimate and preserves the safe-path policy. \\
\arrayrulecolor{black!12}\hline
Qwen3-Omni & 5 & \raggedright No visible rationale. & No & No & N/A & Provides a valid action but no visible rationale to assess. \\
\end{longtable}
\arrayrulecolor{black}
\endgroup

\section{Error Types and Model-Specific Limitations}
\label{app:failure-analysis}

\subsection{Failure-Mode Definitions}
\label{app:failure-details}

Following common environment-benchmark practice~\citep{paglieri2025balrog,liu2024agentbench,hu2026lmgame}, we
track four failure modes separately. \textbf{Timeout} indicates that the Max
Steps limit is reached before the pass condition. \textbf{Invalid action}
indicates that all three permitted responses are malformed or outside the
allowed action set. \textbf{Token overflow} occurs when the accumulated input
exceeds the model's supported context. \textbf{Early termination} denotes an
in-environment failure or death before the pass condition is reached. In every case,
the reported score retains the progress achieved before failure.

\subsection{Model-Specific Limitations}
\label{app:model-specific-limitations}

Table~\ref{tab:model-specific-limitations} summarizes the dominant weakness of
17 of the evaluated models using the per-category results in
Table~\ref{tab:main-results}, the full 342-run result set, and the conservative
trajectory proxies summarized in the table caption.  A small category gap for a low-scoring
model should not be read as robustness: floor effects can compress the
difference between two difficult subsets.

\begin{table*}[!htbp]
\centering
\scriptsize
\setlength{\tabcolsep}{3pt}
\renewcommand{\arraystretch}{1.08}
\caption{\textbf{Distinctive limitations of 17 of the evaluated models.}  Category
scores are normalized 0--100.  Overflow and trajectory counts are computed
from the 342 raw runs per model.  The action-fixation proxy requires a failed
run of at least ten steps with one action occupying at least 90\% of the
trajectory; conflicting-action counts require two different bracketed actions
within one response.}
\label{tab:model-specific-limitations}
\begin{tabular}{p{0.15\textwidth}p{0.3\textwidth}p{0.48\textwidth}}
\toprule
\textbf{Model} & \textbf{Largest diagnostic gap or reliability issue} & \textbf{Model-specific limitation} \\
\midrule
Seed-2.1-Pro & Dynamic: $74.3\!\rightarrow\!47.0$ ($-27.3$) & Best overall model, but autonomous world evolution remains its dominant weakness; its Multiple-memory score also falls by 16.2 points. \\
Gemini-3.6-Flash & Dynamic: $77.2\!\rightarrow\!39.2$ ($-38.0$) & Largest static-to-dynamic drop in the benchmark.  Raw traces also expose rare but consequential stale first-line commitments (11 responses). \\
Gemini-3.1-Pro & Dynamic: $67.7\!\rightarrow\!40.8$ ($-26.9$) & Strong static reasoning does not transfer to continuous tracking and replanning; Pong and BattleZone traces show unstable trajectory and alignment estimates. \\
GPT-5.6-sol & Multiple: $82.5\!\rightarrow\!42.4$ ($-40.1$) & By far the largest Single-to-Multiple degradation: excellent current-frame performance but weak retention and integration of earlier post-action states. \\
Gemini-3-Flash & Dynamic: $54.7\!\rightarrow\!34.0$ ($-20.7$) & Older Flash model is especially prone to repeated actions and egocentric-map confusion; 30/342 failed runs meet the conservative action-fixation proxy. \\
Seed-2.0-Lite & Multiple: $54.2\!\rightarrow\!34.9$ ($-19.3$) & Compared with Seed-2.1-Pro, it has a 17.4-point overall deficit and a larger relative dependence on the immediately visible state. \\
Qwen3.5-397B-A17B & Dynamic: $45.0\!\rightarrow\!27.9$ ($-17.1$) & Competitive open-weight model, yet action diversity often fails to become coherent control; the MultiRoom trace exhibits extensive toggling and turning without a persistent route. \\
DeepSeek-V4.1-Flash & Dynamic: $51.0\!\rightarrow\!28.8$ ($-22.2$) & Strongest open-weight model in the main results; strong control-task performance (71.6) drops substantially in dynamically evolving environments. \\
Gemma-4-31B & Multiple: $40.1\!\rightarrow\!19.7$ ($-20.4$) & Error-free API execution does not prevent behavioral failure: repeated \texttt{Done} after a false success judgment and long action runs are characteristic. \\
MiMo-v2.5 & Overall 20.4; 231/342 runs required action-format retry & Low score without context overflow, plus the highest rate of recoverable action-format retries; interface compliance, rather than context capacity, is its distinctive reliability weakness. \\
Qwen2.5-VL-72B & 82/342 failed runs meet action-fixation proxy & Limited state adaptation dominates its profile; its Multiple-memory score (17.5) is only 1.5 points below Single because both are near the performance floor. \\
MiMo-VL-7B & Context overflow: 114/342 (33.3\%) & One third of runs terminate before the environment does, while completed traces also show long repeated-action segments; both context handling and feedback control constrain the score. \\
Qwen3.6-27B & 194/342 failed runs meet action-fixation proxy & Despite a nominally small category gap, its overall score is 16.9 and it has the second-highest conservative fixation count, indicating weak use of post-action feedback. \\
GLM-4.5V & Context overflow: 24/342 (7.0\%) & Reliability and control compound: 155/342 failed runs meet the fixation proxy in addition to the context-limit failures. \\
Qwen3.5-27B & 121/342 fixation; 54 conflicting-action responses & Low overflow (1/342); its distinctive weaknesses are action fixation (121/342) and plan–commitment inconsistency (54 responses). \\
Qwen3-Omni-30B & 195/342 failed runs meet action-fixation proxy & Highest fixation count among evaluated models; the all-\texttt{Forward} MultiRoom trace is the limiting behavior in its clearest form. \\
Qwen3-VL-235B & Context overflow: 247/342 (72.2\%) & The benchmark's most severe infrastructure-level limitation.  Only 95 runs reach an environment outcome, so its low score partly reflects trajectories truncated by the 131K endpoint context limit. \\
\bottomrule
\end{tabular}
\end{table*}

\section{Dataset Statistics and Characteristics}
\label{app:dataset-statistics}

This appendix characterizes both the 37-environment task pool and the
trajectories produced by the four leading models in
Table~\ref{tab:main-results}: Seed-2.1-Pro, Gemini-3.6-Flash,
Gemini-3.1-Pro, and GPT-5.6-sol.  The trajectory analyses cover all 342 tasks
for each model (1,368 episodes total).  The random reference repeats every
task three times (1,026 episodes).

\subsection{Task Complexity Analysis}
\label{app:task-complexity}

\paragraph{Difficulty stratification.}
Let $\bar{s}_{m,g}$ be model $m$'s mean normalized score over all tasks in
environment $g$, and let $\bar{s}_{\mathrm{rand},g}$ be the corresponding mean over
the random-policy runs.  We define the empirical learnability gap
\begin{equation}
    \Delta_g = \max_{m\in\mathcal{M}_4}\bar{s}_{m,g}
    - \bar{s}_{\mathrm{rand},g},
\end{equation}
where $\mathcal{M}_4$ contains the four models above.  Following the specified
thresholds, an environment is Easy when $\Delta_g>70$, Medium when
$30\leq\Delta_g\leq70$, and Hard when $\Delta_g<30$.  This metric measures
improvement over random behavior rather than absolute score alone.  The
observed gaps span 71.7--100.0 in Easy, 30.0--63.4 in Medium, and 6.9--29.0
in Hard.  Table~\ref{tab:appendix-h-difficulty} lists every environment; the Hard set
contains eight ALE environments plus Procgen Dodgeball.

\begin{table*}[!htbp]
\centering
\small
\setlength{\tabcolsep}{5pt}
\renewcommand{\arraystretch}{1.12}
\caption{\textbf{Difficulty strata for all 37 environments.}  For each game
$g$, $\Delta_g$ is the gap between the best mean score among the four leading
models and the mean random-policy score.  Scores and gaps are on the
normalized 0--100 scale.}
\label{tab:appendix-h-difficulty}
\begin{tabular}{@{}>{\raggedright\arraybackslash}p{0.24\linewidth}>{\centering\arraybackslash}p{0.14\linewidth}>{\raggedright\arraybackslash}p{\dimexpr0.62\linewidth-4\tabcolsep\relax}@{}}
\toprule
\textbf{Tier} & \textbf{Games} & \textbf{Environment list} \\
\midrule
Easy ($\Delta_g>70$) & 9 &
Turmoil, Acrobot, CliffWalking, Custom LavaCrossing, Custom MultiRoom,
Custom UnlockPickup, MountainCar, Taxi, and Procgen Heist \\
Medium ($30\leq\Delta_g\leq70$) & 19 &
Alien, Assault, Asterix, DemonAttack, Enduro, Freeway, Pong, RoadRunner,
Trondead, CartPole, Custom Breakout, FlappyBird, Highway, Procgen Bigfish,
Procgen Bossfight, Procgen Caveflyer, Procgen Leaper, Procgen Ninja, and
Tetris \\
Hard ($\Delta_g<30$) & 9 &
BattleZone, Berzerk, ChopperCommand, Frostbite, MsPacman, Riverraid,
Seaquest, Tennis, and Procgen Dodgeball \\
\bottomrule
\end{tabular}
\end{table*}

\paragraph{Task-attribute associations.}
Because these statistics summarize the complete fixed benchmark suite, the
primary quantities are the per-environment group means; the correlation test is an
auxiliary consistency check rather than an estimate from repeated rollouts.
For memory requirement, the unit of analysis is a model--environment mean, giving
148 equally weighted cells.  Multiple-memory environments score 47.2 on average,
compared with 70.0 for Single-memory environments, a 22.8-point difference and a
negative point-biserial correlation ($r_{\mathrm{pb}}=-0.258$, $p=.0015$).
The random reference reverses this ordering, scoring 5.6 on Single and 15.0 on
Multiple environments. The four leading models likewise average 71.9 on Static versus
41.5 on Dynamic environments, while random again reverses the ordering (8.4 versus
16.1). These negative-control reversals are inconsistent with a generic
category-size or score-scale explanation for the model gaps.
For token usage, we sum logged input, output, and reasoning tokens, average
within each model--environment cell, apply a log transform, and standardize within
model so provider-specific token scales do not dominate.  Environment
dynamics has no detectable association with this measure
($r_{\mathrm{pb}}=-0.035$, $p=.669$), despite raw means of 1.338M tokens for
static and 1.159M for dynamic environments.

\begin{table*}[!htbp]
\centering
\small
\setlength{\tabcolsep}{5pt}
\renewcommand{\arraystretch}{1.12}
\caption{\textbf{Task-attribute associations across the four leading models.}
Each cell is one model--game mean.  Memory uses Single versus Multiple as the
binary attribute.  Token correlation is computed on log token totals
standardized within each model; the displayed token means are untransformed.
$r_{\mathrm{pb}}$ denotes the point-biserial correlation.}
\label{tab:appendix-h-attribute-correlations}
\begin{tabular}{p{0.23\textwidth}p{0.37\textwidth}rrr}
\toprule
\textbf{Association} & \textbf{Group means} & \textbf{Cells} & \textbf{Correlation} & \textbf{$p$} \\
\midrule
Memory requirement vs. score &
Single: 70.0; Multiple: 47.2 & 148 & $r_{\mathrm{pb}}=-0.258$ & .0015 \\
Environment dynamics vs. token usage &
Static: 1.338M; Dynamic: 1.159M tokens & 148 & $r_{\mathrm{pb}}=-0.035$ & .669 \\
\bottomrule
\end{tabular}
\end{table*}

\begin{table*}[!htbp]
\centering
\small
\setlength{\tabcolsep}{5pt}
\renewcommand{\arraystretch}{1.10}
\caption{\textbf{Game type and failure mode.}  Early-termination and timeout
percentages are conditional on failure.  $\phi_{\mathrm{timeout}}$ is the
binary correlation between membership in a game type and timeout (rather than
early termination) among failed episodes.  Game types are multi-label, so rows
overlap.}
\label{tab:appendix-h-failure-by-type}
\begin{tabular}{lrrrrr}
\toprule
\textbf{Game type} & \textbf{Failed/all} & \textbf{Failure rate} &
\textbf{Early termination} & \textbf{Timeout} & \textbf{$\phi_{\mathrm{timeout}}$} \\
\midrule
Collect & 76/160 & 47.5\% & 69.7\% & 30.3\% & $+0.108$ \\
Control & 16/60 & 26.7\% & 75.0\% & 25.0\% & $+0.025$ \\
Dodge & 431/860 & 50.1\% & 85.8\% & 14.2\% & $-0.182$ \\
Maze & 150/388 & 38.7\% & 73.3\% & 26.7\% & $+0.111$ \\
Shooter & 304/588 & 51.7\% & 97.7\% & 2.3\% & $-0.413$ \\
Sports & 14/40 & 35.0\% & 85.7\% & 14.3\% & $-0.018$ \\
\bottomrule
\end{tabular}
\end{table*}

Of the 1,368 leading-model episodes, 624 fail: 506 terminate in-environment and 118
time out.  None ends through an invalid action, token overflow, or another API
error.  Among failures, shooter environments are most strongly associated with early
termination rather than timeout ($\phi_{\mathrm{timeout}}=-0.413$), with
97.7\% of shooter failures ending in-environment.  Dodge environments show the same weaker
pattern ($\phi=-0.182$), whereas maze and collect environments have small positive
associations with timeout ($\phi=0.111$ and $0.108$).  Because task-type tags
are multi-label, Table~\ref{tab:appendix-h-failure-by-type} is descriptive and
its rows intentionally overlap.

\FloatBarrier
\subsection{Action Space Statistics}
\label{app:action-space-statistics}

Table~\ref{tab:appendix-h-action-spaces} reports the final discrete action
vocabulary actually presented to agents in each environment.  A combination action
is one simultaneous multi-button command, such as \texttt{Up+Shoot}; scalar
labels such as \texttt{Torque +1} are not combinations.  The evaluated action
spaces contain 2--18 actions (mean 8.9): CartPole and FlappyBird attain the
minimum, while nine ALE environments attain the maximum.  Seventeen environments
retain at least one combination action, accounting for 138 environment-specific
combination entries in total.

\begingroup
\footnotesize
\setlength{\tabcolsep}{5pt}
\renewcommand{\arraystretch}{1.05}
\begin{longtable}{p{0.62\textwidth}rr}
\caption{\textbf{Action-space size used in the benchmark.}  ``Actions'' is
the final per-game vocabulary presented to the evaluated agents; ``Combo''
counts simultaneous multi-button actions within that vocabulary.}
\label{tab:appendix-h-action-spaces}\\
\toprule
\textbf{Environment} & \textbf{Actions} & \textbf{Combo} \\
\midrule
\endfirsthead
\multicolumn{3}{c}{\tablename\ \thetable\ -- continued}\\
\toprule
\textbf{Environment} & \textbf{Actions} & \textbf{Combo} \\
\midrule
\endhead
\midrule
\multicolumn{3}{r}{Continued on next page}\\
\endfoot
\bottomrule
\endlastfoot
Alien & 18 & 12 \\
Assault & 7 & 2 \\
Asterix & 9 & 4 \\
BattleZone & 18 & 12 \\
Berzerk & 18 & 12 \\
ChopperCommand & 18 & 12 \\
DemonAttack & 6 & 2 \\
Enduro & 9 & 4 \\
Freeway & 3 & 0 \\
Frostbite & 9 & 4 \\
MsPacman & 9 & 4 \\
Pong & 3 & 0 \\
Riverraid & 18 & 12 \\
RoadRunner & 18 & 12 \\
Seaquest & 18 & 12 \\
Tennis & 18 & 12 \\
Trondead & 18 & 12 \\
Turmoil & 12 & 6 \\
Acrobot & 3 & 0 \\
CartPole & 2 & 0 \\
CliffWalking & 4 & 0 \\
Custom Breakout & 4 & 0 \\
Custom LavaCrossing & 7 & 0 \\
Custom MultiRoom & 7 & 0 \\
Custom UnlockPickup & 7 & 0 \\
MountainCar & 3 & 0 \\
Taxi & 6 & 0 \\
FlappyBird & 2 & 0 \\
Highway & 5 & 0 \\
Procgen Bigfish & 15 & 4 \\
Procgen Bossfight & 6 & 0 \\
Procgen Caveflyer & 5 & 0 \\
Procgen Dodgeball & 6 & 0 \\
Procgen Heist & 5 & 0 \\
Procgen Leaper & 5 & 0 \\
Procgen Ninja & 5 & 0 \\
Tetris & 5 & 0 \\
\end{longtable}
\endgroup

\subsection{Episode Length Statistics}
\label{app:episode-length-statistics}

Episode length is the number of agent-controlled steps after the deterministic
preload that establishes the evaluated start state.  For a run with length
$L$ and configured action budget $B$, we label an outcome \emph{early success}
when the pass condition is met and $L<B$, and \emph{early death} when an
in-environment failure terminates the run without a pass and $L<B$.  Runs with
$L\geq B$ are assigned to \emph{budget reached}; thus a successful early exit
is never counted as an early death.

\begin{table}[H]
\centering
\small
\setlength{\tabcolsep}{3pt}
\renewcommand{\arraystretch}{1.10}
\caption{\textbf{Post-preload episode-length distribution by model.}  Each
row contains 342 episodes.  Early success and early death both require the
episode to end strictly before its configured action budget; the pass label
distinguishes the two outcomes.}
\label{tab:appendix-h-length-by-model}
\begin{tabular}{lrrrrrr}
\toprule
\textbf{Model} & \textbf{Mean $\pm$ SD} & \textbf{Median [IQR]} &
\textbf{Range} & \makecell{\textbf{Early}\\\textbf{success}} & \makecell{\textbf{Early}\\\textbf{death}} & \makecell{\textbf{Budget}\\\textbf{reached}} \\
\midrule
Seed-2.1-Pro & $34.0\pm17.7$ & 33.0 [19.0, 50.0] & 2--92 & 32.7\% & 35.7\% & 31.6\% \\
Gemini-3.6-Flash & $34.1\pm18.6$ & 33.5 [19.0, 50.0] & 3--81 & 25.7\% & 40.6\% & 33.6\% \\
Gemini-3.1-Pro & $35.1\pm18.5$ & 34.0 [19.3, 50.0] & 4--102 & 25.7\% & 34.5\% & 39.8\% \\
GPT-5.6-sol & $36.1\pm18.7$ & 37.0 [19.3, 50.0] & 3--82 & 22.8\% & 36.0\% & 41.2\% \\
\bottomrule
\end{tabular}
\end{table}

\begingroup
\footnotesize
\setlength{\tabcolsep}{3pt}
\setlength{\LTpre}{3pt}
\setlength{\LTpost}{3pt}
\renewcommand{\arraystretch}{1.0}
\begin{longtable}{p{0.30\textwidth}rrrrrr}
\caption{\textbf{Post-preload episode length by environment, pooled across
the four leading models and sorted by mean length.}  ES and ED denote early
success and early death, respectively.}
\label{tab:appendix-h-length-by-game}\\
\toprule
\textbf{Environment} & \textbf{$n$} & \textbf{Mean} &
\textbf{Median [IQR]} & \textbf{Range} & \textbf{ES} & \textbf{ED} \\
\midrule
\endfirsthead
\multicolumn{7}{c}{\tablename\ \thetable\ -- continued}\\
\toprule
\textbf{Environment} & \textbf{$n$} & \textbf{Mean} &
\textbf{Median [IQR]} & \textbf{Range} & \textbf{ES} & \textbf{ED} \\
\midrule
\endhead
\midrule
\multicolumn{7}{r}{Continued on next page}\\
\endfoot
\bottomrule
\endlastfoot
Tennis & 20 & 10.0 & 8.0 [4.0, 14.5] & 4--23 & 40.0\% & 60.0\% \\
Procgen Leaper & 20 & 10.6 & 9.0 [5.8, 14.2] & 2--22 & 0.0\% & 100.0\% \\
CliffWalking & 20 & 10.7 & 10.0 [10.0, 12.0] & 8--13 & 100.0\% & 0.0\% \\
Procgen Bossfight & 20 & 12.3 & 11.0 [6.5, 15.8] & 4--29 & 20.0\% & 80.0\% \\
Taxi & 20 & 15.9 & 16.0 [13.0, 17.0] & 12--27 & 100.0\% & 0.0\% \\
Procgen Dodgeball & 20 & 17.9 & 16.5 [6.0, 28.0] & 3--40 & 0.0\% & 100.0\% \\
Custom LavaCrossing & 60 & 18.2 & 20.0 [14.0, 21.0] & 2--36 & 65.0\% & 35.0\% \\
Highway & 80 & 18.7 & 19.0 [12.0, 26.0] & 4--39 & 30.0\% & 70.0\% \\
Frostbite & 20 & 19.9 & 18.0 [15.0, 25.0] & 15--32 & 45.0\% & 55.0\% \\
Procgen Bigfish & 20 & 21.0 & 15.5 [12.2, 31.2] & 3--48 & 40.0\% & 55.0\% \\
CartPole & 20 & 21.2 & 24.0 [14.0, 30.0] & 6--30 & 0.0\% & 60.0\% \\
Procgen Caveflyer & 20 & 23.1 & 19.0 [9.0, 27.5] & 6--60 & 40.0\% & 25.0\% \\
Seaquest & 20 & 23.1 & 17.0 [6.0, 43.2] & 5--50 & 0.0\% & 85.0\% \\
FlappyBird & 20 & 27.1 & 31.5 [9.0, 37.0] & 6--70 & 60.0\% & 40.0\% \\
Custom Breakout & 20 & 29.1 & 23.0 [21.8, 38.5] & 3--92 & 75.0\% & 20.0\% \\
MountainCar & 20 & 30.4 & 29.0 [25.0, 35.2] & 19--52 & 95.0\% & 0.0\% \\
Custom UnlockPickup & 60 & 30.6 & 30.0 [24.8, 40.0] & 16--40 & 50.0\% & 20.0\% \\
Procgen Ninja & 20 & 31.2 & 26.5 [23.8, 34.0] & 16--102 & 40.0\% & 60.0\% \\
Riverraid & 100 & 32.6 & 37.0 [16.0, 50.0] & 5--50 & 0.0\% & 71.0\% \\
RoadRunner & 20 & 33.0 & 31.5 [25.8, 40.5] & 7--54 & 30.0\% & 20.0\% \\
Berzerk & 68 & 33.7 & 26.0 [19.0, 46.0] & 10--70 & 45.6\% & 44.1\% \\
ChopperCommand & 20 & 33.7 & 37.0 [18.0, 50.0] & 11--50 & 0.0\% & 65.0\% \\
Assault & 20 & 36.0 & 35.5 [30.2, 43.0] & 18--48 & 45.0\% & 20.0\% \\
Trondead & 100 & 36.5 & 43.0 [24.0, 50.0] & 6--50 & 0.0\% & 59.0\% \\
Procgen Heist & 20 & 36.9 & 39.0 [31.5, 44.2] & 17--52 & 55.0\% & 0.0\% \\
Asterix & 20 & 37.8 & 42.5 [25.8, 50.0] & 12--50 & 0.0\% & 55.0\% \\
MsPacman & 20 & 42.0 & 44.0 [38.0, 50.0] & 19--50 & 0.0\% & 55.0\% \\
BattleZone & 60 & 42.2 & 50.0 [32.2, 50.0] & 17--50 & 0.0\% & 36.7\% \\
Turmoil & 20 & 43.6 & 50.0 [44.0, 50.0] & 18--50 & 0.0\% & 30.0\% \\
Alien & 20 & 44.1 & 44.5 [33.5, 60.0] & 12--72 & 35.0\% & 45.0\% \\
DemonAttack & 100 & 44.2 & 50.0 [50.0, 50.0] & 8--50 & 0.0\% & 23.0\% \\
Enduro & 80 & 50.0 & 50.0 [50.0, 50.0] & 50--50 & 0.0\% & 0.0\% \\
Pong & 20 & 50.0 & 50.0 [50.0, 50.0] & 50--50 & 0.0\% & 0.0\% \\
Tetris & 20 & 50.5 & 52.0 [38.5, 57.0] & 32--84 & 5.0\% & 10.0\% \\
Freeway & 80 & 53.3 & 61.0 [28.0, 72.0] & 21--81 & 27.5\% & 0.0\% \\
Custom MultiRoom & 60 & 56.1 & 57.5 [45.8, 70.0] & 35--70 & 73.3\% & 0.0\% \\
Acrobot & 20 & 62.4 & 60.0 [54.0, 69.8] & 42--82 & 55.0\% & 0.0\% \\
\end{longtable}
\endgroup

Across models, mean length is 34.0--36.1 steps (SD 17.7--18.7).  Pooled
outcomes are 366/1,368 early successes (26.8\%), 502 early deaths (36.7\%),
and 500 budget reaches (36.5\%).  Tennis (10.0), Procgen Leaper (10.6), and
CliffWalking (10.7) are shortest by mean; Acrobot (62.4), Custom MultiRoom
(56.1), and Freeway (53.3) are longest.  Because budgets vary, duration is not
difficulty.  Early-death rate peaks at 100\% for Procgen Leaper and Procgen
Dodgeball, 85\% for Seaquest, and 80\% for Procgen Bossfight.

\FloatBarrier

\end{document}